\documentclass[10pt,conference,letterpaper]{IEEEtran}
\usepackage{times,amsmath,epsfig}
\usepackage{graphicx}
\usepackage{balance}
\usepackage{flushend}
\usepackage[tight,footnotesize]{subfigure}
\usepackage[hyphens]{url}
\usepackage[ruled,vlined,linesnumbered]{algorithm2e}
\SetKwRepeat{Do}{do}{while}
\usepackage{algpseudocode}
\usepackage{amsmath}
\usepackage{amssymb}
\usepackage{wrapfig}
\usepackage[font=bf,skip=3pt]{caption}
\usepackage{cite}
\usepackage{xcolor}

\title{Learning to Route with Sparse Trajectory Sets\\---Extended Version}
\author{%
{Chenjuan Guo, Bin Yang, Jilin Hu, Christian S. Jensen }\\%
\fontsize{10}{10}\selectfont\itshape
Department of Computer Science, Aalborg University, Denmark\\
\fontsize{9}{9}\selectfont\ttfamily\upshape
\{cguo, byang, hujilin, csj\}@cs.aau.dk\\
}
\begin{document}
\maketitle
\begin{abstract}
Motivated by the increasing availability of vehicle trajectory data, we propose \emph{learn-to-route}, a comprehensive trajectory-based routing solution.
Specifically, we first construct a graph-like structure from trajectories as the routing infrastructure. Second, we enable trajectory-based routing given an arbitrary (source, destination) pair.

In the first step, given a road network and a collection of trajectories, we propose a trajectory-based clustering method that identifies regions in a road network.
If a pair of regions are connected by trajectories, we maintain the paths used by these trajectories and learn a routing preference for travel between the regions.
As trajectories are skewed and sparse, %and although the introduction of regions serves to consolidate the sparse data,
many region pairs are not connected by trajectories. We thus transfer routing preferences from region pairs with sufficient trajectories to such region pairs and then use the transferred preferences to identify paths between the regions.
In the second step, we exploit the above graph-like structure to achieve a comprehensive trajectory-based routing solution.
Empirical studies with two substantial trajectory data sets offer insight into the proposed solution, indicating that it is practical.
A comparison with a leading routing service offers evidence that the paper's proposal is able to enhance routing quality.

This is an extended version of ``Learning to Route with Sparse Trajectory Sets''~\cite{icde2018}, to appear in IEEE ICDE 2018.

\end{abstract}

% NOTE keywords are not used for conference papers so do not populate them
% \begin{keywords}
% keyword-1, keyword-2, keyword-3
% \end{keywords}
%
\section{Introduction}

Vehicular transportation is an important aspect of the daily lives of many people and is essential to many businesses as well as society as a whole~\cite{DBLP:journals/sigmod/GuoJ014,DBLP:journals/tc/Ding0C016}.
As a part of the continued digitization of societal processes, more and more data is becoming available in the form
of trajectories that capture the movements of vehicles~\cite{RiskAware,DBLP:journals/tits/Ding0GL15}. This data offers a foundation for improving vehicular transportation, including vehicle routing.
%Such trajectories offer insight into the behavior of drivers, e.g., their choice of paths, in addition to many other aspects of a transportation infrastructure, e.g., waiting times at intersections, free-flow speeds, and congestion.

Traditional routing is {\it cost-centric} and aims at returning paths with minimal costs, e.g., distance, travel time, or fuel consumption. The cost of a path is computed from edge costs in \emph{edge-based cost modeling}~\cite{DBLP:conf/icde/Guo0AJT15, DBLP:conf/edbt/HuaP10, tkde2017, DBLP:conf/icde/YangGJKS14, icde20182, DBLP:journals/pvldb/0002GJ13} or sub-path costs  in \emph{path-based cost modeling}~\cite{DBLP:journals/pvldb/DaiYGJH16, PACE, DBLP:conf/gis/AljubayrinYJZ16, DBLP:journals/geoinformatica/HuYJM17}.
%To account for dynamic traffic conditions, the costs are often time-varying~\cite{DBLP:conf/edbt/DingYQ08, DBLP:conf/icde/YangGJKS14} and sometimes uncertain~\cite{pvldb17pathcost, DBLP:conf/edbt/HuaP10}.
%
%including both \emph{edge-based routing}~\cite{dijkstra1959note, DBLP:conf/edbt/HuaP10, DBLP:conf/icde/YangGJKS14} and \emph{path-based routing}~\cite{pvldb17pathcost, DBLP:conf/gis/AljubayrinYJZ16}.
%
%Cost-centric routing aims at identifying paths with minimal costs, where the cost of a path is computed from costs of edges in edge-based routing or of sub-paths in path-based routing.
%
In such routing, trajectory data is often used for annotating the edges or sub-paths with travel costs such as travel times; and routing services employ shortest path algorithms, e.g., Dijkstra's algorithm or contraction hierarchies~\cite{DBLP:conf/wea/GeisbergerSSD08}, to return fastest, or simply shortest, paths.
%
%Trajectory data records the movement of vehicles. It is well recognized that historical trajectories reflect traffic conditions~\cite{DBLP:journals/tkde/YuanZXS13, DBLP:journals/vldb/0002GMJ15}. Further, local drivers who generated the trajectories are considered as an intelligent group who have good knowledge of driving conditions in a city~\cite{DBLP:conf/icde/ChenSZ11, DBLP:conf/sigmod/LuoT0N13}.
%
However, an existing study~\cite{DBLP:conf/mdm/CeikuteJ13} suggests that local drivers who drive passenger vehicles follow paths that differ substantially from the paths computed using cost-centric routing and are often neither fastest nor shortest. Our paper also focuses on trajectory data that was generated from passenger vehicles.

We study a very different routing approach that relies on the availability of trajectories from local drivers.
Assuming that local drivers implicitly take into account a multitude of factors, such as traffic conditions, turns, travel time, fuel consumption, road types, and traffic lights, when making routing decisions and thus know best which paths are preferable, we propose a methodology that utilizes paths found in historical trajectories to construct new paths between arbitrary $($source, destination$)$ pairs. %Since we do not model explicit costs of paths,
We call this \emph{trajectory-based} routing.

If historical trajectories show that many drivers traveling from a source $s$ to a destination $d$ follow a particular path, it is straightforward to recommend that path to drivers asking for directions from $s$ to $d$.
The big challenge now is how to benefit from historical trajectories when no historical trajectories capture
paths from $s$ to $d$. This is important because any set of historical trajectories is \emph{sparse} in the sense that it is unlikely to provide paths for all $s$'s and $d$'s.
For example, the road network of Denmark, a small country, contains some 1.6 million edges. Thus, if all edges are candidate $s$'s and $d$'s, a minimum of 2.6 trillion $(s, d)$ pairs are needed. Given that the distribution of trajectories in a road network is skewed, an enormous set of trajectories (e.g., trillions for Denmark and quadrillions for Germany) would be needed before routing could be done by simply looking up paths of past trajectories for any $(s, d)$ pair.

\begin{figure}[ht!]
\centering
  \includegraphics[width=0.95\columnwidth]{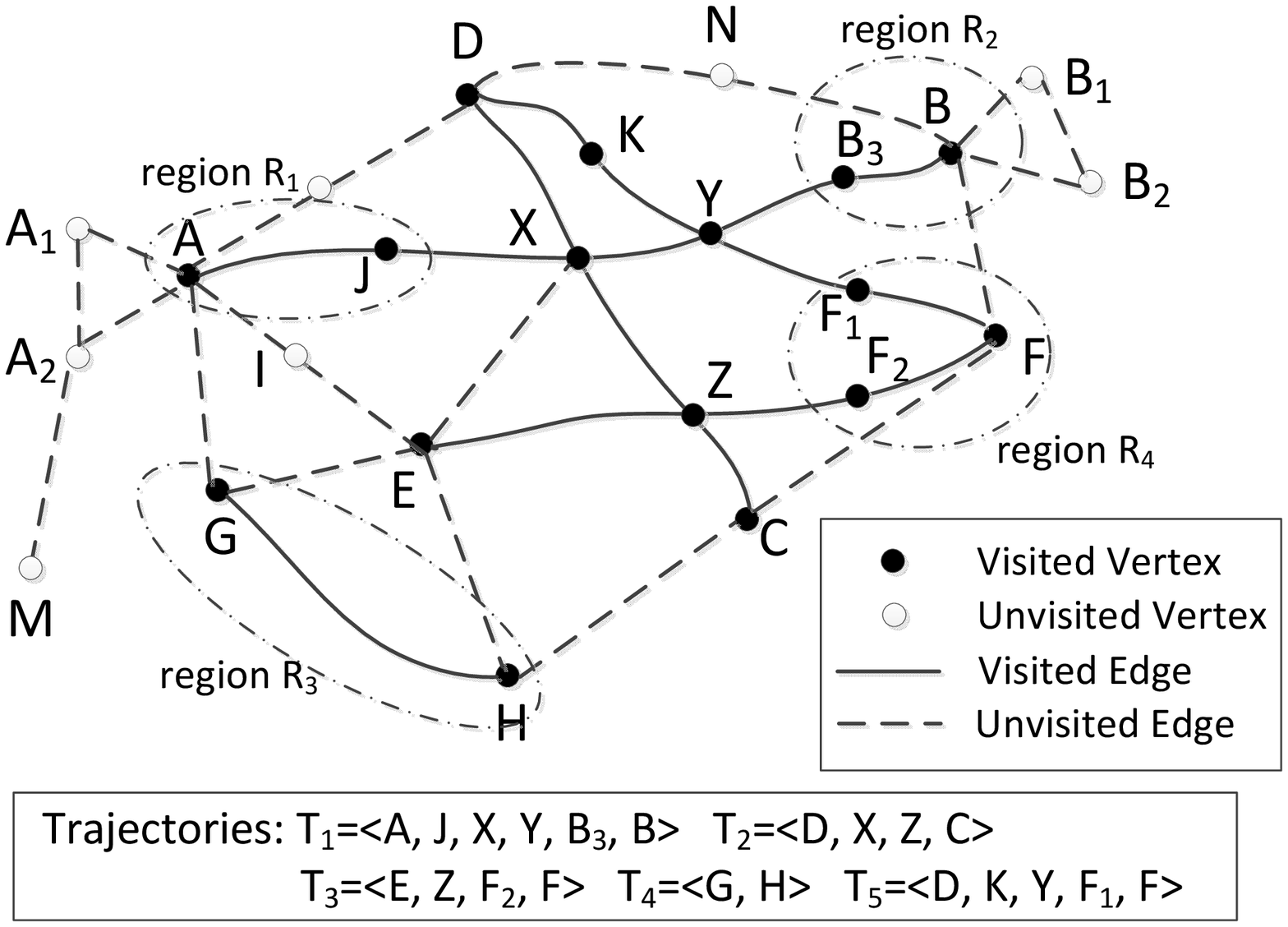}
\caption{Motivating Example}
\label{fig:intro}
%%\vspace{-10pt}
\end{figure}

%Figure~\ref{fig:intro} exemplifies the main idea of the paper's proposal.
Figure~\ref{fig:intro} exemplifies the problem setting.
The solid edges and filled vertices are covered by a set of five trajectories, while the dashed edges and unfilled vertices are not covered by any trajectories. For example, trajectory $T_1$ visited $A$ and then $J$, $X$, $Y$, and $B_3$ before reaching $B$. If routing from $A$ to $B$ is requested, the path $A\rightarrow$ $J \rightarrow$ $X \rightarrow$  $Y \rightarrow$ $ B_3 \rightarrow$  $B$, as captured by trajectory $T_1$, can be recommended directly. The challenge is to enable routing for $(s, d)$ pairs that are not connected by trajectories, e.g., $(A_1$, $B_2)$ and $(H$, $F)$.

%%\vspace{-1pt}

To enable trajectory-based routing with massive, but still sparse, sets of historical trajectories, we propose means that are able to generalize the cases where historical trajectories can be utilized for routing. This includes \emph{three} steps.
In the first step, we cluster vertices into regions and thus map a road network graph into a \emph{region graph}. %
Trajectories that originally connect vertices in the road network graph now connect regions in the region graph. %
%If a pair of regions are connected by trajectories, we maintain the paths used by these trajectories.
This arrangement generalizes the cases where trajectories can be used for routing from being between specific vertex pairs to being between region pairs. As regions include multiple vertices, this arrangement contributes to solving the data sparseness problem.
%
%However, in the region graph, some region pairs are still not connected by any trajectories.

For example, in Figure~\ref{fig:intro}, $A$ and $J$ are clustered into region $R_1$, and $B_3$ and $B$ are clustered into region $R_2$. Now, although no trajectories connect $A_1$ and $B_2$, $T_1$ connects regions $R_1$ and $R_2$ that are close to $A_1$ and $B_2$. Thus, the path of $T_1$ can be used for recommending a path from $A_1$ to $B_2$. For instance, a user may go from $A_1$ to $A$, then follow the path used by $T_1$ to reach $B$, and then go to $B_2$. This enables trajectory-based routing between regions connected by trajectories.
However, in the region graph, some region pairs are still not connected by any trajectories, e.g., regions $R_3$ and $R_4$ in Figure~\ref{fig:intro}.

In the second step, we learn routing preferences from available historical trajectories that connect some region pairs and then transfer these preferences to similar region pairs that are not connected by trajectories. Based on the transferred preferences, we identify paths for the non-covered region pairs.
Note that the routing preferences are learned for different region pairs, not for different individual drivers.
%% not covered by trajectories.
%
Assume that $(R_1$, $R_2)$ is similar to $(R_3$, $R_4)$, e.g., because both are from a residential area to a business district. Next, we extract a routing preference from the trajectories connecting $R_1$ and $R_2$ that explains the choice of paths from $R_1$ to $R_2$. We transfer this routing preference to driving from $R_3$ to $R_4$ and then identify paths connecting $R_3$ and $R_4$, upon which trajectory-based routing from $H$ to $F$ is possible.

In the third step, we provide a unified routing solution, called \emph{learn-to-route (L2R)}, which performs path finding on the region graph, thus enabling routing between arbitrary $(s, d)$ pairs in the original road network graph.

%

%The proposed methodology generalizes historical trajectories into routing preferences that can be applied to recommend paths between sources and destinations that are not covered by historical trajectories.
%We first cluster homogeneous vertices into \emph{regions}. Then, a \emph{region graph} with these regions as vertices is formed. Trajectories that originally connected vertices now connect regions. This arrangement increases the cases where trajectories can be used for routing, which helps dealing with data sparseness.

%The next step is to enable routing for region pairs not connected by trajectories, e.g., from region $R_3$ to region $R_4$. To do so, we first identify connected region pairs that are similar to the unconnected region pairs. Assume that $(R_1$, $R_2)$ is similar to $(R_3$, $R_4)$, e.g., because both are from a residential area to a business district. Next, we extract a driving preference from the trajectories connecting $R_1$ and $R_2$ that explains the choice of paths from $R_1$ to $R_2$. We transfer this routing preference to driving from $R_3$ to $R_4$ and then identify paths connecting $R_3$ and $R_4$, upon which trajectory-based routing from $H$ to $F$ is possible.

%Finally, we provide a unified routing solution, called \emph{learn-to-route (L2R)}, that combines the above techniques to enable routing on the region graph.
%
To the best of our knowledge, this is the first solution that learns routing preferences from historical trajectories and transfers the learned preferences to the part of a road network that is not covered by trajectories, thus supporting comprehensive trajectory-based routing for arbitrary $(s, d)$ pairs.

The paper makes four contributions. First, it presents a trajectory-based road network clustering algorithm that produces the data foun\-dation---the region graph. Second, it presents a general routing preference model, including an algorithm that extracts preferences from historical trajectories and an algorithm that transfers preference to similar region pairs. Third, it presents a unified routing algorithm for the region graph. Fourth, it reports on an empirical evaluation that offers insight into the proposed solution, indicating that it is capable of efficiently computing paths that match those of local drivers better than do traditional routing services.

%\noindent
Paper Outline:
%The remainder of the paper is organized as follows.
Section 2 covers related work. Section 3 covers preliminaries. % and solution overview.
Section 4 presents Step 1, region graph generation. Section 5 presents Step 2, preference learning and transfer. Section 6 presents Step 3, unified routing. Section~7 reports on empirical evaluations. Section 8 concludes.% and covers research directions.

\section{Related Work}
\label{sec:relatedwork}

%\textbf{Path recommending based on trajectories:}
We first review studies on employing historical trajectories for \textbf{path recommendation}, considering three cases.

\emph{Case 1:} Given a source and a destination, complete trajectories exist that connect the source to the destination. For example, given $A$ and $B$ in Figure~\ref{fig:intro}, trajectory $T_1$ went from $A$ to $B$. Then, the path of trajectory $T_1$ is recommended.
%
%In case different paths are used by different trajectories, each path is assigned a popularity value, and the path with the highest popularity is recommended.
When multiple paths exist, the path with the highest popularity is recommended, where the popularity can be defined using different strategies~\cite{DBLP:conf/icde/ChenSZ11,DBLP:conf/mdm/CeikuteJ15,DBLP:conf/sigmod/LuoT0N13}. % have been proposed to assign a popularity values.
%Existing studies take into account
%the numbers of traversals~\cite{DBLP:conf/icde/ChenSZ11}, distinct users associated with a path~\cite{DBLP:conf/mdm/CeikuteJ15} and also take temporal aspects into account~\cite{DBLP:conf/sigmod/LuoT0N13}.
% the total number of traversals~\cite{DBLP:conf/icde/ChenSZ11} or the total number of distinct drivers~\cite{DBLP:conf/mdm/CeikuteJ15}. Some studies also distinguish popularity values during different time periods~\cite{DBLP:conf/sigmod/LuoT0N13}.
%
This is the simplest case, which is also considered in our proposal.

\emph{Case 2:} Given a source and a destination, no complete trajectories exist that connect the source to the destination, but trajectories exist that can be \emph{spliced} such that the spliced trajectories connect the source to the destination.
In Figure~\ref{fig:intro}, given $A$ and $F$, sub-paths $A\rightarrow X$ from $T_1$, $X\rightarrow Z$ from $T_2$, and $Z \rightarrow F$ from $T_3$ can be spliced to form a path from $A$ to $F$. Alternatively, $T_1$ and $T_5$ can also be spliced to enable a different path from $A$ to $F$.
To determine which spliced path is ``best'', absorbing Markov chains~\cite{DBLP:conf/icde/ChenSZ11} and hidden Markov models~\cite{DBLP:conf/icde/DaiYGD15} are employed to the probabilities that different spliced paths may occur based on historical trajectories. %  are evaluated based on these models.
The spliced path with the highest probability is chosen.
%
%However, these works splice complete trajectories into pieces, and thus they cannot preserve the information of drivers' intelligence and traversed paths that are original available in the complete paths. In contrast, our solution provides a comprehensive way of utilizing complete trajectories and meanwhile it still considers to minimize the travel distance.
%
In contrast, we learn routing preference vectors from trajectories and apply the preference vectors to identify best paths.

%\emph{Case 3:} When neither complete trajectories nor spliced trajectories are able to connect a source to a destination, existing methods~\cite{DBLP:conf/icde/ChenSZ11,DBLP:conf/mdm/CeikuteJ15, DBLP:conf/sigmod/LuoT0N13, DBLP:conf/icde/DaiYGD15} no longer work. The region graph used in our solution is able to generalize the use of historical trajectories in Case 3. Further, the mechanism of learning and transferring routing preferences captured by past trajectories can also be applied to enable path recommendation in Case 3.

\emph{Case 3:} Neither complete nor spliced trajectories are able to connect a source to a destination. In the example, consider, e.g., $A_1$ to $B_2$, $H$ to $F$, and $M$ to $N$.
Here, existing methods~\cite{DBLP:conf/icde/ChenSZ11,DBLP:conf/mdm/CeikuteJ15, DBLP:conf/sigmod/LuoT0N13, DBLP:conf/icde/DaiYGD15} no longer work. %, e.g., $A_1$ to $B_2$, $H$ to $F$, and $M$ to $N$.
%
%Our proposal is able to generalize and transfer routing preferences such that the behavior captured by past trajectories can also be applied to enable path recommendation in Case 3.
%
In this paper, the use of the proposed region graph, together with the mechanism of learning and transferring routing preferences captured by past trajectories, makes it possible to extend the situations where historical trajectories can be utilized to cover also Case 3. %In addition, the mechanism of learning and transferring routing preferences captured by past trajectories is applied to enable path recommendation in Case 3.

Next, we review related work on \textbf{road network clustering}. Gonzalez {\it et al.}~\cite{DBLP:conf/vldb/GonzalezHLMS07} propose a graph partition method based on prior knowledge of the road network hierarchy with $l$ levels, which may vary from country to country. %In particular, the graph is partitioned at each of the $l$ levels, where some vertices belong to a single region and others belong to multiple regions.
%This method does not use information from trajectory data.
%
Wei {\it et al}.~\cite{DBLP:conf/kdd/WeiZP12} propose a grid-based method for constructing regions using trajectories, where two adjacent grid cells are merged if more than $\tau$ trajectories exist that passed through them. %
%First, a road network is divided into disjoint cells. Then, a cell is merged with its neighbor cells if the number of trajectories that passed through them is above a threshold $\tau$.
%
These studies rely heavily on ``appropriate'' parameters, e.g., $l$ and $\tau$. Tuning such parameters is non-trivial.
Based on recent advances in modularity based graph clustering, we propose a generic, parameter-free region generation method, where parameters such as $l$ and $\tau$ are not needed. 
Our proposal is also different from POI clustering~\cite{DBLP:journals/tkde/ShangZJYKLW15}. 
Finally, we consider \textbf{learning of routing preferences}~\cite{DBLP:conf/ssd/BalteanuJS13, DBLP:journals/vldb/0002GMJ15, DBLP:conf/aaai/LetchnerKH06, DBLP:conf/gis/DellingGGKTW15}.
%
%A few studies propose different methods for extracting routing preferences from historical trajectories.
Methods~\cite{DBLP:journals/vldb/0002GMJ15,DBLP:conf/ssd/BalteanuJS13} compare the paths used by trajectories to skyline paths~\cite{DBLP:conf/icde/YangGJKS14} to identify different users' dominating factors when choosing paths, e.g., travel time, fuel consumption, or distance. %Other methods~\cite{DBLP:conf/icde/DaiYGD15} use the ratios between different factors, e.g., travel time divided by fuel consumption, to model routing preferences.
TRIP~\cite{DBLP:conf/aaai/LetchnerKH06} uses the ratios between individual drivers' travel time and average travel time to model personalized travel times. A recent study from Microsoft presents an algorithm that learns driver-specific parameters for Bing Maps' ranking function for candidate paths based on individual drivers's past trajectories~\cite{DBLP:conf/gis/DellingGGKTW15}.
However, all existing methods  work only when trajectories are available. In contrast, our proposal is also able to transfer routing preferences to places without trajectories, where existing methods do not apply.

%In summary, the paper's proposal generalizes and extends existing proposals to enable more versatile trajectory-based routing.

%%\vspace{-5pt}
\section{Preliminaries}% and Solution Overview}

We cover the definitions of important concepts, introduce the problem, and present a solution overview.

%\begin{definition}
%\label{def:roadnetwork}
A \textbf{road network} is a weighted graph $\mathcal{G} = (\mathbb{V}, \mathbb{E}, \mathbb{W})$, where vertex set $\mathbb{V}$ consists of vertices representing road intersections, edge set $\mathbb{E} \subseteq \mathbb{V} \times \mathbb{V}$ consists of edges representing road segments, and $\mathbb{W}$ is a set of weight functions, where each function has signature $\mathbb{E} \rightarrow \mathbb{R}^+$. %mapping to the edge set.
For specificity, we maintain four functions in $\mathbb{W}$. Functions % in this paper, but other functions can also be used.
$w_{\mathit{DI}}(\cdot)$, $w_{\mathit{TT}}(\cdot)$, $w_{\mathit{FC}}(\cdot)$, and $w_{\mathit{RT}}(\cdot)$ return the distance (\emph{DI}), travel time (\emph{TT}), fuel consumption (\emph{FC}), and road type (\emph{RT}) of the argument edge, respectively.

%\begin{definition}
%\label{def:route}
% Bin: path now is a sequence of vertices not edges.
A \textbf{path} ${P}$$=\langle v_1, v_2, \ldots, v_{a}
\rangle$ is a sequence of vertices where two consecutive vertices are connected by an edge. %that connect distinct vertices, where $e_i\in
%\mathbb{E}$, $e_i.d=e_{i+1}.s$ for $1 \leqslant i < a$, and the vertices $e_1.s$, $e_2.s$, $\ldots$, $e_a.s$,
%and $e_a.d$ are distinct.
%
%unique edges $\langle e_1, \ldots, e_m\rangle$, where two consecutive edges must share a vertex. Specifically, given two consecutive edges $e_i = (v_p, v_q)$ and $e_{i+1} = (v_x, v_y)$, where $1 \leqslant i < m $, we have $v_q=v_x$; given two arbitrary edges, we have $e_i\neq e_j$ if $i\neq j$, where $1 \leqslant i, j \leqslant m$.
%
%The cardinality of path $\mathcal{P}$, denoted as $|\mathcal{P}|$, is
%the number of edges in the path.
%
%Path ${P}'=\langle f_1, f_2, \ldots, f_{x}\rangle$ is a \textit{sub-path} of ${P}$ if $|\mathcal{P}'| \leqslant |\mathcal{P}|$ and there exists an edge sequence in $\mathcal{P}$ such that $f_1=e_i$, $f_2=e_{i+1}$, $\ldots$, $f_x=e_{i+x-1}$. $\square$
%
%\IEEEQEDclosed
%$\square$
%\end{definition}

%\begin{definition}
%\label{def:trajectory}
A \textbf{trajectory} $\mathcal{T}$ is a time-ordered sequence of GPS records capturing the movement of an object, where %In this paper, we refer to a moving object as a vehicle.
a GPS record captures the location of the object at a time point.
The time gap between two consecutive GPS records in trajectories varies, from a few seconds (a.k.a., high-frequency trajectories) to tens of seconds or a few minutes (a.k.a., low-frequency trajectories). In the experiments, we test the proposed method on both a high-frequency and a low-frequency GPS data sets.
%, and the GPS records in a trajectory are ordered based on their time points. $\square$
%A GPS sample $g_i = (\mathit{location}, t)$ records the location of the moving object at time point $t$.
%The GPS samples in a trajectory are ordered based on their time points, i.e., given two GPS samples $g_i, g_j\in %\mathcal{T}$, $g_i.t < g_j.t$ if $1 \leqslant i < j \leqslant n$.
%$\square$
%\end{definition}
%
Map matching~\cite{DBLP:conf/gis/NewsonK09} is able to align a trajectory with the road-network path that the trajectory traversed. %It is more challenge to map match low frequency trajectories than high-frequency trajectories, due to large time gaps between two consecutive GPS records.
%
%In the experiments, we test the proposed method on both a high-frequency and a low-frequency GPS data sets.
%
%It is out of the scope
For example, the path used by trajectory $\mathcal{T}_1$ is ${P}_{\mathcal{T}_1}=\langle A, J, X, Y, B_3, B\rangle$.

%
%in the underlying road network, i.e.,  %A trajectory $\mathcal{T}$ is then can be represented as a sequence of cost records $M_\mathcal{T} = \langle l_1, \ldots, l_p\rangle$. Each record $l_i = (e_i, t_i, \mathbf{C}_i)$ indicates that edge $e_i\in E$ is traversed by the trajectory $\mathcal{T}$; the traversal on $e_i$ starts at time point $t_i$; cost vector $\mathbf{C}_i= \langle c_1, \ldots, c_q\rangle$ records different travel costs, e.g., distance and travel time, for traversing $e_i$.
%
%the path that the trajectory traverses.
%
%in $M_\mathcal{T}$, denoted as ${P}_\mathcal{T}=\langle e_1, e_2, \ldots, e_p \rangle$.
%\IEEEQEDclosed
%

\noindent
\emph{\textbf{Problem Setting.}} We study a new routing methodology---\emph{trajectory-based} routing. Specifically, we study how to best utilize the paths found in trajectories to enable routing for arbitrary source and destination $(s, d)$ pairs such that the identified paths are similar to the paths chosen by local drivers.

\noindent
\emph{\textbf{Spareness.}} The \emph{spareness} considered in the paper means that past trajectories cannot cover paths between all possible $(s, d)$ pairs, so simply looking up paths of past trajectories for a given $(s, d)$ pair does not work.
Although it may be possible that a substantial set of trajectories cover the roads in a road network, e.g., the 1.6 million edges in Denmark, it is almost impossible to cover all possible $(s, d)$ pairs with paths. Having just one path for each $(s, d)$ pair in Denmark calls for 2.6 trillion trajectories.
The key challenge is to conquer data sparseness by making it possible to benefit from historical trajectories for routing from $s$ to $d$ when no trajectories capture paths from $s$ to $d$.

%Given a road network, a set of historical GPS trajectories from local drivers, a source and destination pair $(s, d)$, we identify a path such that it is

%To this end, we study three problems.
%the \emph{learn-to-route} problem, which consists of three sub-problems.
%
%First, how to generalize the cases where historical trajectories can be reused.
%
%Given a road network and a set of historical GPS trajectories, (1) we cluster vertices into regions and build a region graph; (2) we learn routing preferences from the trajectories that connect some region pairs and transfer routing preferences to the region pairs where no trajectories are available; (3) we use the learned and transferred routing preferences to recommend paths for arbitrary $(s, d)$ pairs.

\begin{figure}[h]
\centering
  \includegraphics[width=0.8\columnwidth]{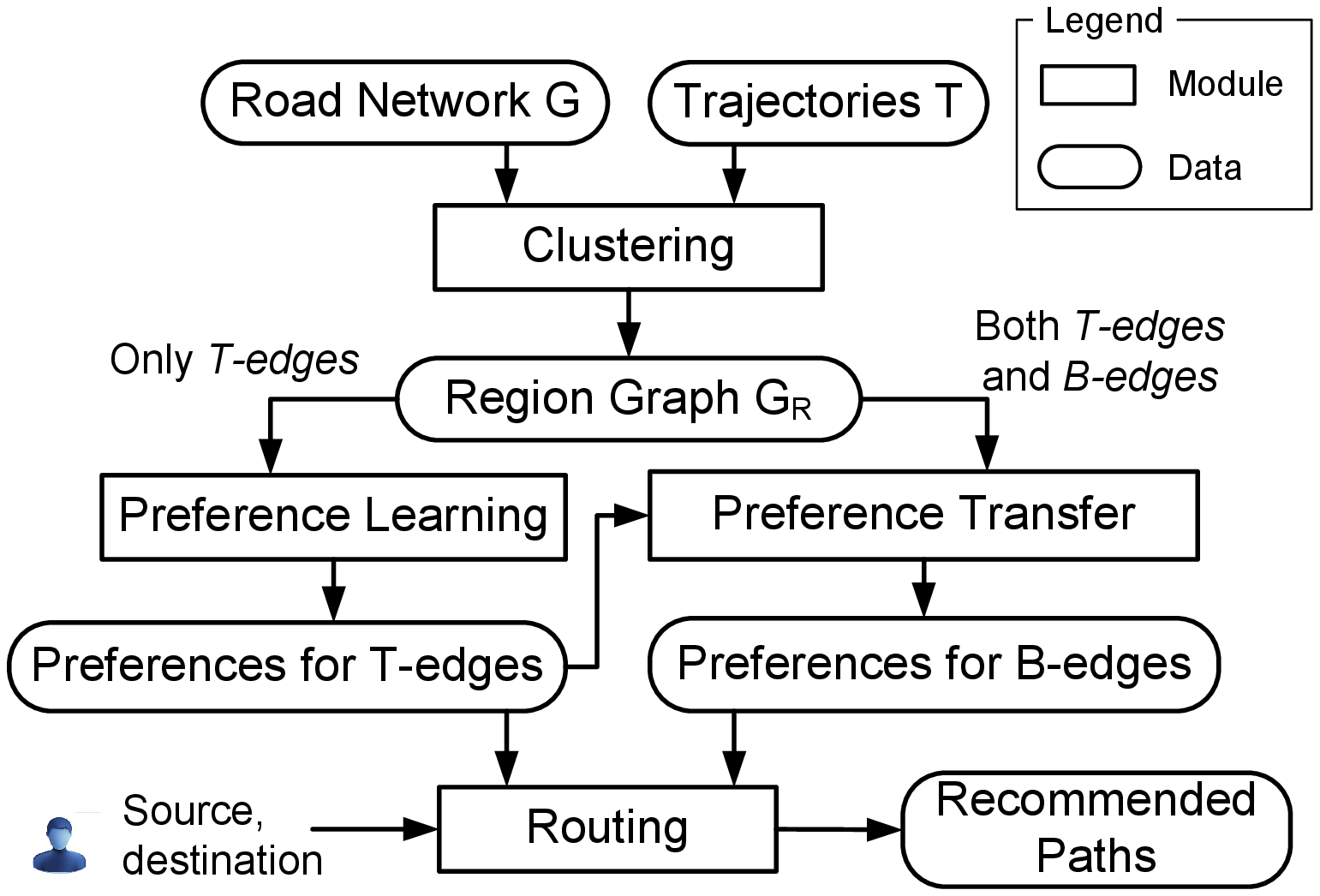}
\caption{Solution Overview}
\label{fig:overview}
%%\vspace{-15pt}
\end{figure}

%
%we recommend a route between a pair of arbitrary source and destination proposed by a user. In particular, the route is connected by a set of trajectory edges whose popularity are maximized and the number of complete trajectory paths is maximized.
\noindent
\emph{\textbf{Solution Overview.}} We propose a three-step procedure to conquer the data sparseness problem, as outlined in  Figure~\ref{fig:overview}. % offers a solution overview.

Given a road network $\mathcal{G}$ and a set of trajectories $\mathbb{T}$, the \emph{clustering module} employs modularity-based clustering to cluster vertices into regions, thus obtaining a region graph $\mathcal{G_R}$.
We partition the edges in a region graph into T-edges and B-edges, according to whether they are traversed or not traversed by trajectories, respectively.
For each T-edge, the \emph{preference learning module} learns a routing preference. The resulting preferences are fed into the \emph{preference transfer module} as training data, and the \emph{preference transfer module} transfers the preferences from T-edges to similar B-edges.
Based on the learned and transferred preferences, the \emph{routing module} recommends paths for user-specified $(s, d)$ pairs.

\noindent
\emph{\textbf{Scope of the paper.}} (1) To account for \emph{time-dependent} traffic conditions, we construct peak and off-peak region graphs using trajectories that occurred in peak and off-peak periods, respectively. These are constructed the same way, so we disregard the distinction in the presentation. Depending on the departure time, one of the two region graphs is chosen for routing. % depending on the leaving time. % and the other one for off-peak periods using trajectories that occurred in off-peak periods.
Modeling time-dependent traffic conditions at a finer granularity and building a dynamic region graph are interesting extensions that are left for future work.

\noindent
(2) \emph{L2R} utilizes trajectories from multiple drivers to recommend paths, and thus is not a personalized routing approach. In Section~\ref{sec_exp:compare}, we empirically compare \emph{L2R} with state-of-the-art personalized routing approaches. \emph{L2R} can also be adapted to support personalized routing by only using the trajectories from specific drivers, which we also leave as future work.

%The proposed solution does not consider the effect of traffic conditions, e.g., peak and off-peak periods, when connecting paths traversed by trajectories. Rather, the proposed solution provides a general approach for maximizing the profits of using drivers' intelligence concealed in the existing trajectories. Indeed, the traffic conditions are important factors to the routing effect, which is considered as future work on top of this general solution.
%
%3) The trajectory data considered in the paper are high frequency data, e.g., every 1 to 5 seconds per record. As the development of many trajectory tracking devices, high frequency data is ubiquitous. Thus, extra efforts on dealing with low frequency trajectory data, e.g., 1 minute per record, such as dealing with the uncertainty between two location records, are not consider here.

%\section{Generating Transfer Centers}
%\label{sec:tc}

\section{Building the Region Graph}
\label{sec:networkPartitioning}

%\section{Learning To Route}
%\label{sec:learningtoroute}

We propose a trajectory-based method for \emph{clustering} the vertices of a road network into \emph{regions} (Section~\ref{subsec:modularity}).
Then, we build a \emph{region graph} that connects pertinent regions (Section~\ref{subsec:connecting regions}).
The region graph extends the cases where trajectories can be used for recommending paths between an arbitrary pair of source and destination, thus providing a foundation for the final routing module.

\subsection{Clustering Vertices to Regions}
\label{subsec:modularity}

A {\bf region} is a set of homogenous vertices where the homogeneity is defined based on two properties that are used in urban planning~\cite{liang2013unraveling,forbes1999urban}: (i) the numbers of trajectories associated with the vertices in a region are similar~\cite{liang2013unraveling}; (ii) the edges connecting the vertices have the same road type~\cite{forbes1999urban}. The intuition is as follows. A region with vertices connected by edges of residential-road type may capture a residential area; and by taking into account the number of trajectories associated with the vertices, we can distinguish a residential area in the city from one in a suburb area because the former has more trajectories.

\begin{figure}[h]
\centering
  \includegraphics[width=0.75\columnwidth]{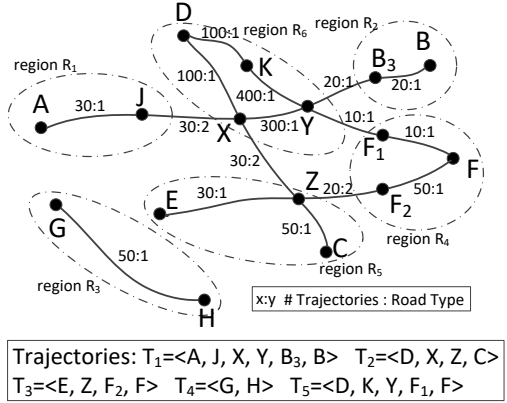}
\caption{An Example of Regions}
\label{fig:clusters}
%%\vspace{-10pt}
\end{figure}

Consider Figure~\ref{fig:clusters}, where the label $x$:$y$ on an edge indicates that $x$ trajectories occurred on the edge and that the road type of the edge is $y$. For example, 100 trajectories occurred on edge $(D, X)$, a type 1 road. According to the above two properties, vertices $D$, $K$, $X$, and $Y$ can be regarded as a region because they have more trajectories than the other vertices and are connected by road type 1 edges. Similarly, vertices $F$, $F_1$, and $F_2$ can be regarded as a region.

%which may be a part of city center. Similarly, Regions 1, 2, 4, and 5 are traversed by much fewer trajectories and may be different residential areas.

%\vspace{-1pt}

Based on the two properties, we propose a modularity-based method that clusters vertices connected by the same road types into regions.
The setting is a \emph{trajectory graph} that consists of vertices and edges that are traversed by trajectories. Figure~\ref{fig:clusters} shows the trajectory graph of the road network in Figure~\ref{fig:intro}. A trajectory graph may not be a connected graph.

Next, we define \emph{popularity} values for the edges and vertices in a trajectory graph.
The popularity $s_{ij}$ of edge $e=(v_i,v_j)$ is the number of trajectories that occurred on edge $e$.
The popularity $S_i$ of vertex $v_i$ is the sum of the trajectories that occurred on the edges that are incident to $v_i$, i.e., $S_i=\sum_{j} s_{ij}$.
Next, we define $S=\sum_{(v_i, v_j)\in \mathbb{E}} s_{ij}$ as the sum of the popularity values of all edges in the trajectory graph.

\emph{Modularity}, which is used widely in the network analysis literature~\cite{newman2004finding,newman2004analysis}, quantifies the quality of the clusters in a graph from a global perspective.
In our context, the modularity is high if the \emph{popularity} of edges inside clusters is high and the \emph{popularity} of edges between clusters is low, which is desired by property (i) of regions.
%
%A high modularity value is preferable if higher edge weights are expected inside clusters and lower edge weights are expected between clusters.

We define \emph{modularity gain}~\cite{newman2004finding,newman2004analysis,DBLP:conf/aaai/ShiokawaFO13} $\Delta Q_{v_iv_j}$ to quantify the benefit of merging vertices $v_i$ and $v_j$ into a cluster: % which is defined as
\[ \small
  \Delta Q_{v_iv_j} = \left\{
  \begin{array}{l l}
    \frac{s_{ij}}{S} - \frac{S_i \cdot S_j}{S^2} & ~~ \text{if } v_i, v_j \text{ are connected by an edge}; \\%
    0 & ~~ \text{otherwise}. \text{ }%\IEEEQEDclosed\\%
  \end{array} \right.
\]
It has been shown that if merging two vertices $v_i$ and $v_j$ gives a non-positive modularity gain, the two vertices should not be merged~\cite{DBLP:conf/aaai/ShiokawaFO13}. If the modularity gain is positive, vertices $v_i$ and $v_j$ are merged into an \emph{aggregate vertex} with a popularity that equals the sum of the popularity of the $v_i$ and $v_j$, i.e., $S_i+S_j$.

To take into account property (ii) of regions, i.e., the road type constraint,
%we define {\it road type} of an original, non-merged vertex as; we define {\it road type} of an aggregate vertex merged from vertices $v_i$ and $v_j$ as the road type of edge $(v_i, v_j)$.
we also associate a \emph{road type} attribute with an aggregate vertex that records the road type of edge $(v_i, v_j)$. % i.e., $w_{RT}(v_i, v_j)$.

%We proceed to propose a hierarchical clustering method that follows a bottom-up, agglomerative clustering strategy. %~\cite{han2011data}.
%%
%In the beginning, each vertex is treated as a cluster.
%%
%The method keeps merging clusters into larger clusters until no more clusters can be merged.
%%
%In particular, the method merges a vertex $v_k$ with the highest popularity, regardless of whether it is an aggregate or an ordinary vertex, with its adjacent vertices if the merging gives a positive modularity gain and only involves edges with the same road type.
%%
%If $v_k$ has no such adjacent vertices, $v_k$ forms a region.
%%
%Supplementary document~\cite{supp} offers algorithmic details.

We proceed to propose a hierarchical clustering method that follows a bottom-up, agglomerative clustering strategy~\cite{han2011data}.
In the beginning, each vertex is treated as a cluster.
The clustering method keeps merging clusters into larger clusters until no more clusters can be merged.

\textbf{Merging vertices.}
We call an original, non-merged vertex a \emph{simple vertex} and refer to a cluster that contains merged vertices as an \emph{aggregate vertex}. We differentiate the processes of merging adjacent simple and aggregate vertices.

\emph{Merging two simple vertices: } Given two adjacent simple vertices $v_i,v_j\in \mathbb{V}'$, if $\Delta Q_{v_iv_j}>0$, we merge the two vertices into an aggregate vertex $v_a$, whose popularity is the sum of the popularity values of both $v_i$ and $v_j$. In addition, we set the road type of the aggregate vertex $v_a.\mathit{RT}$ as the road type of edge $(v_i, v_j)$, i.e., $w_{\mathit{RT}}(v_i, v_j)$.

After merging $v_i$ and $v_j$, the topology of the graph is adjusted. First, $v_i$ and $v_j$ are removed from $\mathbb{V}'$, and the aggregate vertex $v_a$ is added to $\mathbb{V}'$.
Second, the edge $(v_i, v_j)$ is removed from $\mathbb{E}'$ and any edges that used to connect to $v_i$ and $v_j$ are now connected to $v_a$.
%In particular, for each edge $(v_m, v_i)$ that goes to $v_i$ is now changed to an edge $(v_m, v_a)$ that goes to $v_a$; and for each edge $(v_i, v_n)$ that goes from $v_i$ is now changed to an edge $(v_a, v_n)$ that goes from $v_a$. The popularity values of all the edges keep the same.
%%
%The same transformation also applies to vertex $v_j$.
We use function $\mathit{MergeSS(v_i, v_j)}$ to denote the procedure of merging two simple vertices $v_i$ and $v_j$.
%%
%\\=====Examples?? or example in figure 3 is enough?

\emph{Merging an aggregate vertex and a simple vertex: } We use function $\mathit{MergeAS(v_a, v_i)}$ to denote the procedure of merging an aggregate vertex $v_a$ and a simple vertex $v_i$. If the modularity gain $\Delta Q_{v_av_i}>0$ and if the road type $w_{\mathit{RT}}(v_a, v_i)$ is consistent with the road type of the aggregate vertex $v_a.\mathit{RT}$, the two vertices are merged in a way similar to merging two simple vertices. Otherwise, the two vertices cannot be merged, and edge $(v_a, v_i)$ is removed from edge set $\mathbb{E}'$.

\emph{Merging two aggregate vertices: } We use function $\mathit{MergeAA}(v_a$, $v_{a'})$ to denote the procedure of merging two aggregate vertices $v_a$ and $v_{a'}$. If $\Delta Q_{v_av_{a'}}>0$ and if the road type $v_a.\mathit{RT}$ is consistent with $v_{a'}.\mathit{RT}$, the two vertices are merged similarly to the process of merging two simple vertices. Otherwise, they are not merged, and edge $(v_a, v_{a'})$ is removed from edge set $\mathbb{E}'$.

\textbf{Clustering process.}
The clustering method always chooses a vertex $v_k$ with the highest popularity, regardless of whether it is an aggregate or a simple vertex, to merge with its adjacent vertices.
If $v_k$ has no adjacent vertex, $v_k$ forms a region.
If $v_k$ has adjacent vertices, we compose a vertex set $\mathit{VA}$ that consists of all the adjacent vertices, which are candidates for being merged with $v_k$. Next, we further filter the vertices in $\mathit{VA}$ to identify the vertices that are final candidates for being merged with $v_k$, which forms vertex set $\mathit{VB}$.

\emph{Checking qualification: }
Function $\mathit{CheckQ}(v_k$, $v_j)$ checks whether $v_k$ can be merged with an adjacent vertex $v_j$ in $\mathit{VA}$. It returns {\it true} if $v_k$ and $v_j$ can be merged.
We distinguish cases according to whether vertices $v_k$ and $v_j$ are simple or aggregate vertices. All cases should satisfy the condition $\Delta Q_{v_kv_j}>0$. In addition, we state the additional road type related conditions that vertices $v_k$ and $v_j$ must satisfy to be merged in Table~\ref{tab:mergeCondition}.

%%\vspace{5pt}
\begin{table}[h]
\small
\center
\begin{tabular}{|c|c|c|}
  \hline
     &   $v_k$: Simple&   $v_k$: Aggre.\\  \hline \hline
  $v_j$: Simple &       $\emptyset$             &  $v_k.\mathit{RT}=w_{\mathit{RT}}(v_k, v_j)$ \\ \hline
  $v_j$: Aggre. & $v_j.\mathit{RT}=w_{\mathit{RT}}(v_k, v_j)$  & $v_j.\mathit{RT} = v_k.\mathit{RT}$ \\ \hline
\end{tabular}
\caption{Additional Condition for Merging $v_k$ and $v_j$}
\label{tab:mergeCondition}
\end{table}

\emph{Merging selection: }
%
%Assume we have obtained a set of adjacent vertices $B$ of $v_k$ that are qualified to be merged with $v_k$. $\mathit{SelectM}(v_k$, $B)$ returns vertices in $B$ that are chosen to be merged with $v_k$.
%
After filtering, vertex set $\mathit{VB}$ consists of $v_k$'s adjacent vertices that have passed the qualification check.
We call function $\mathit{SelectM}(v_k, \mathit{VB})$ that returns a subset $\mathit{VB}'\subseteq \mathit{VB}$ of vertices that should be merged with $v_k$.

If $v_k$ is an aggregate vertex, $v_k$ has a road type $v_k.\mathit{RT}$. All vertices in $\mathit{VB}$ are returned by $\mathit{SelectM}(v_k, \mathit{VB})$, i.e., $\mathit{VB}'=\mathit{VB}$. % to merge with $v_k$.
This is so because Table~\ref{tab:mergeCondition} enforces that $v_k.\mathit{RT}=w_{\mathit{RT}}(v_k, v_j)$, if $v_j$ is a simple vertex, and that $v_j.\mathit{RT} = v_k.\mathit{RT}$, if  $v_j$ is an aggregate vertex.

If $v_k$ is a simple vertex, the edges between $v_k$ and its adjacent vertices $v_1,  \ldots,  v_{m}\in \mathit{VB}$ may have different road types, i.e., $w_{\mathit{RT}}(v_k, v_{1})$,  $\ldots$, and $w_{\mathit{RT}}(v_k, v_{m})$ may be different.
Thus, $\mathit{SelectM}(v_k, \mathit{VB})$ returns the largest subset of vertices in $\mathit{VB}$ such that their edges that are incident to $v_k$ have the same road type. %selects majority vertices in $B$, the edges between which and $v_k$ have the same road type.
For example, let $\mathit{VB}=\{v_1, v_{2}, v_3\}$, $w_{\mathit{RT}}(v_k, v_{1})=1$,  $w_{\mathit{RT}}(v_k, v_{2})=1$, and $w_{\mathit{RT}}(v_k, v_{3})=2$, then $\mathit{SelectM}(v_k, \mathit{VB})$ returns $\mathit{VB}'=\{v_1, v_{2}\}$. %as $\mathit{VB}'$. % $v_k$. Later, the new aggregate vertex is given road type 1.

Given a trajectory graph ${\mathcal{G}}'= ({\mathbb{V}}', {\mathbb{E}}')$, where vertices in ${\mathbb{V}}'$ are those in the road network $\mathcal{G}$ traversed by trajectories, the pseudo code of the algorithm is presented in Algorithm~\ref{algo:cluster}.

\begin{algorithm}[!ht]
\caption{BottomUpClustering}
\label{algo:cluster}
\begin{small}
\KwIn{A trajectory graph: ${\mathcal{G}}' = ({\mathbb{V}}', {\mathbb{E}}')$;}
\KwOut{A Cluster Set $\mathit{VC}$;}
Initialize a priority queue $\mathit{PQ}$ based on vertex popularity; \\
Insert all vertices in $\mathbb{V}'$ into $\mathit{PQ}$; \\
\While{$\mathit{PQ}$ is not empty}
{
    Vertex $v_k \gets \mathit{PQ}.\mathit{extractMax}()$;\\
    Vertex set $\mathit{VA} \gets \mathit{obtainAdjacentVertices}(v_k)$; \\
    \If{$\mathit{VA}\neq \emptyset$}
    {
        Qualified vertex set $\mathit{VB} \gets \emptyset$;\\
        \For{each vertex $v_j \in \mathit{VA}$}
        {
            \If{$\mathit{CheckQ}(v_k$, $v_j)$}
            {
            	$\mathit{VB} \gets \mathit{VB} \cup \{v_j\}$
            }
%            \Else
%            {
%                Remove edge $(v_k, v_j)$ from $\mathbb{E}'$;
%            }
        }
        $\mathit{VB}' \gets \mathit{SelectM}(v_k, \mathit{VB})$;\\
 	 \For{each vertex $v_j\in \mathit{VA} \setminus \mathit{VB}'$}
        {
        	Remove edge $(v_k, v_j)$ from $\mathbb{E}'$;
        }
        \For{each vertex $v_j \in \mathit{VB}'$}
        {
        	Remove $v_j$ from $\mathit{PQ}$;\\
        	Call one of  $\mathit{MergeSS}(v_k, v_j)$, $\mathit{MergeAS}(v_k, v_j)$, and $\mathit{MergeAA}(v_k, v_j)$ depending on whether $v_k$ and $v_j$ are simple or aggregate vertices and generate a new aggregate vertex $v_{k}$.
        }
         Insert the new aggregate vertex $v_{k}$ into $\mathit{PQ}$;
    }
    \Else
     {
        $\mathit{VC}.add(v_k);$\\
     }
}
\Return $\mathit{VC}$;
\end{small}
\end{algorithm}
We utilize a priority queue to order simple and aggregate vertices according to their popularity values. The priority queue returns the vertex $v_k$ with the highest popularity to start a merge iteration (lines 1--4).
If $v_k$ has no adjacent vertices, it becomes a cluster (line 19). Otherwise, we consider whether $v_k$ should be merged with its adjacent vertices.
We first find those of $v_k$'s adjacent, qualified vertices $\mathit{VB}$ for merging with $v_k$ (lines 8--10).
Then we identify $\mathit{VB}'$ which includes the vertices that will actually be merged with $v_k$ and cut the graph between $v_k$ and the vertices in $\mathit{VA}\setminus\mathit{VB}'$ (lines 11--13).
Finally, we merge $v_k$ with its adjacent vertices in $\mathit{VB}'$ and add it back to the priority queue (lines 14--17).

\begin{figure}[!ht]
\centering
  \includegraphics[width=0.85\columnwidth]{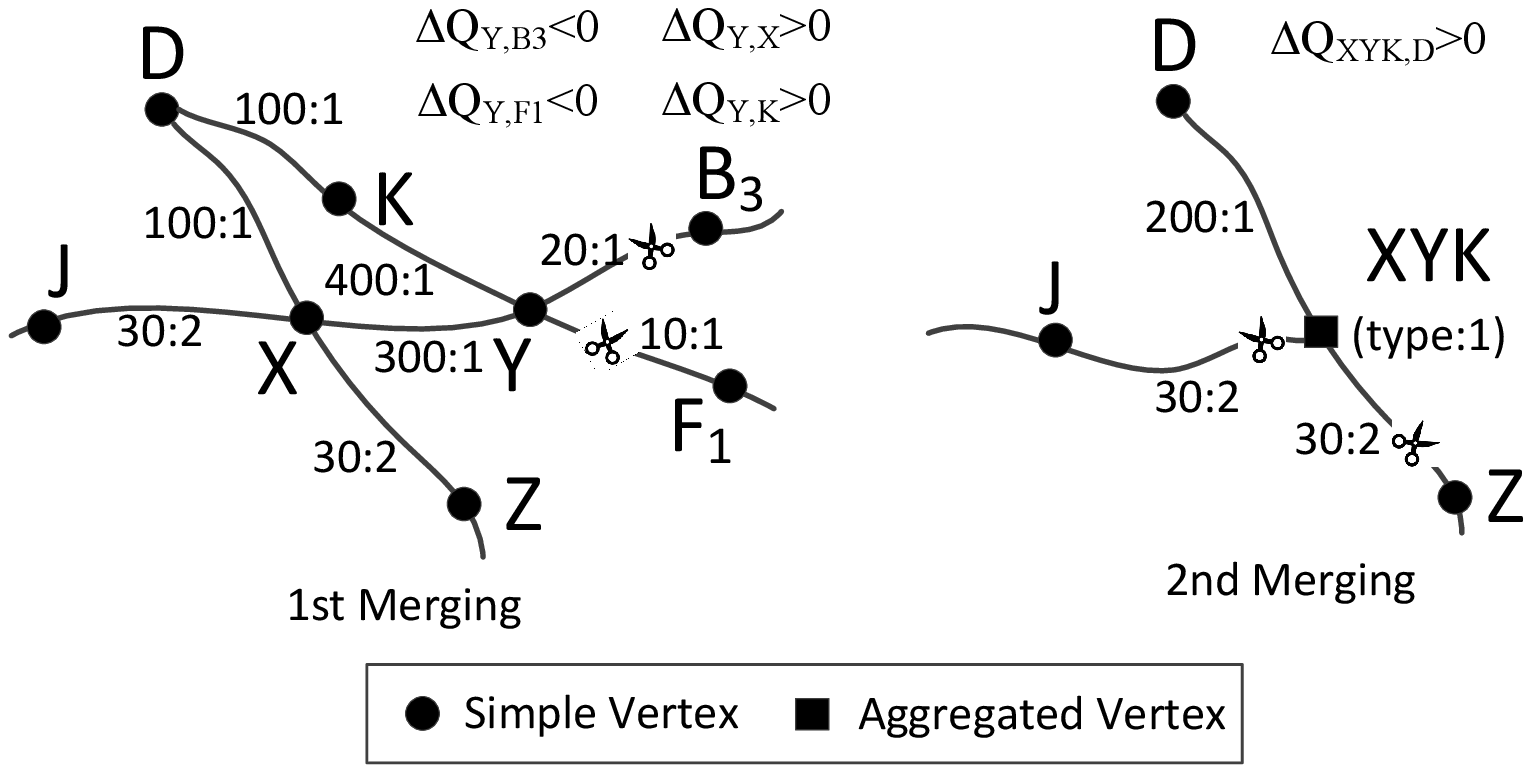}
\caption{An Example of the Clustering Process}
\label{fig:clustersteps}
%%\vspace{5pt}
\end{figure}

A simple example of the clustering algorithm is show in Figure~\ref{fig:clustersteps}.
In the beginning, simple vertex $Y$ has the largest popularity and is first popped from the priority queue.
We compute the modularity gains between $Y$ and $K$, $X$, $B_3$ and $F_1$, respectively. The modularity gains are also shown in Figure~\ref{fig:clustersteps}.
Vertices $B_3$ and $F_1$ are not merged with $Y$ because their modularity gains are negative. Edges $(Y, B_3)$ and $(Y, F_1)$ are removed from $\mathbb{E}'$.
Since vertices $K$ and $X$ have positive modularity gains and the road types of $(Y, K)$ and $(Y, X)$ are both 1, an aggregated vertex $\mathit{XYK}$ is formed with road type 1.

In the 2nd merging iteration, aggregate vertex $\mathit{XYK}$ has the largest popularity and thus is considered with its adjacent vertices. Vertices $J$ and $Z$ cannot be merged with $\mathit{XYK}$ since
the road types are inconsistent. Since the modularity gain between $D$ and $\mathit{XYK}$ is positive and the road type of edge $(D, \mathit{XYK})$ is consistent with the road type of $\mathit{XYK}$, a new aggregate vertex $\mathit{XYKD}$ with road type 1 is created.

% end of supplementary

During the clustering, we need not control manually the size of clusters, as a cluster ``ends'' automatically when merging it with the neighbors gives non-positive modularity gains or they have different road types. This prevents naturally clusters of extremely large sizes.
In addition, we maintain paths used by trajectories inside regions (see ``inner-region paths'' in Section~\ref{subsec:connecting regions}). This design is useful when the source and destination in a routing request is inside a region, which is common for large regions.

% which provide useful information when routing inside a region .

Based on the above, we are able to form regions in a trajectory graph where both properties (i) and (ii) are satisfied. %the numbers of trajectories in different regions are significantly different and where the vertices in a region are connected by edges with the same road types.
For example, the dashed circles in Figure~\ref{fig:clusters} indicate regions.
The {popularity} of edges in region $R_6$ is high, while the {popularity} of the edge between regions $R_2$ and $R_6$ is low; region $R_6$ has road type 1 edges, while the edge between regions $R_6$ and $R_1$ have road type 2.

\subsection{Region Graph}
\label{subsec:connecting regions}

We build a region graph $\mathcal{G_R}=(\mathbb{V}_R, \mathbb{E}_R)$ based on the obtained regions, which serves as a foundation for routing. The region graph can be regarded as a backbone of the road network graph.
To distinguish it from the road network graph, we call a vertex in the region graph \emph{region vertex} and an edge in the region graph \emph{region edge}.
In particular, a region vertex $R_i\in \mathbb{V}_R$ represents a region.
We proceed to show how to construct region edges by connecting region vertices, using the combination of two different strategies.

\begin{figure}[h]
\centering
  \includegraphics[width=0.95\columnwidth]{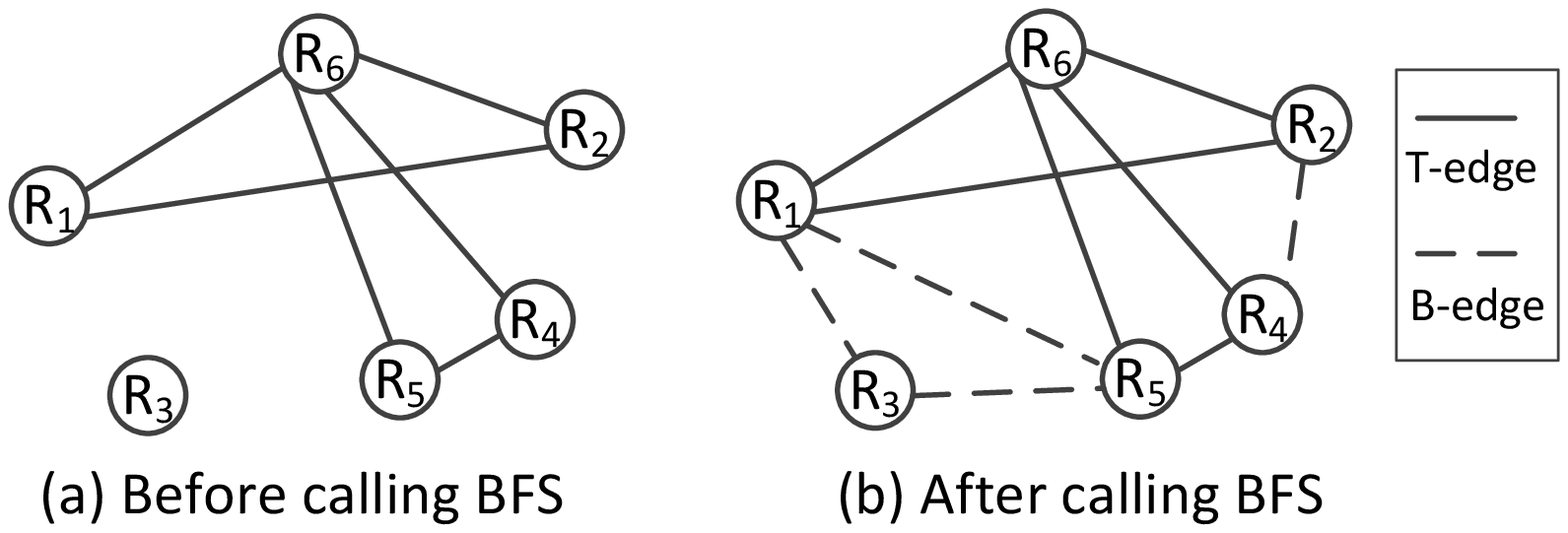}
\caption{Region Graph}
\label{fig:regiongraph}
%%\vspace{-10pt}
\end{figure}

\textbf{Constructing region edges from trajectories:} Having identified regions, trajectories that originally connected vertices in the road network are now utilized to connect regions.
If a trajectory exists that went through a vertex in region $R_i$ and a vertex in region $R_j$, we construct a region edge $(R_i, R_j)$.
Note that a trajectory may produce more than one region edge. In particular, if a trajectory went through vertices in $m$ regions, up to $\frac{m \cdot (m-1)}{2}$ region edges can be constructed.
For example, in Figure~\ref{fig:clusters}, trajectory $\mathcal{T}_1$ went through vertices in $R_1$, $R_6$, and $R_2$, and we are able to construct region edges $(R_1, R_6)$, $(R_1, R_2)$, and $(R_6, R_2)$, as shown in Figure~\ref{fig:regiongraph}(a).

%If a trajectory exists that went through a vertex in region $R_i$ at a vertex $v_a$ and entered region $R_j$ at a vertex $v_b$, we construct a region edge $(R_i, R_j)$.

%Formally, given regions
Each {region edge} $(R_i, R_j)$ is associated with a set $\mathbb{P}_{ij}$ of paths, where each path $P=\langle v_a,\cdots,v_b \rangle$ in $\mathbb{P}_{ij}$ was traversed by at least a trajectory that left $R_i$ at vertex $v_a$ and entered $R_j$ at vertex $v_b$.
A vertex at which a trajectory $\mathcal{T}_k$ enters or leaves a region is called a {\it transfer center}, e.g., $v_a$ and $v_b$.
%
%To ease the following discussion, we also denote a region edge $(R_i, R_j)$ as a path set $\mathbb{P}_{ij}$. They can be used interchangeably.

For example, region edge $(R_1, R_6)$ is associated with path $\langle J, X\rangle$ because trajectory $T_1$ left $R_1$ at vertex $J$ and entered $R_6$ at vertex $X$, and thus $J$ and $X$ are transfer centers.
Similarly, region edge $(R_1, R_2)$ is associated with path $\langle J, X, Y, B_3\rangle$, where $J\in R_1$ and $B_3\in R_2$ are transfer centers; and region edge $(R_6, R_2)$ is associated with path $\langle Y, B_3\rangle$, where $Y\in R_6$ and $B_3\in R_2$ are transfer centers.

For each region, we also maintain \emph{inner-region paths} based on trajectories.
Specifically, given a region $R_i$ and a trajectory $\mathcal{T}_k$, if $\mathcal{T}_k$ entered $R_i$ at $v_c$ and left $R_i$ at $v_d$, the path $P'=\langle v_c,\cdots,v_d \rangle$ %, where $\{v| v\in P' \wedge v\in R_i \}$,
that was traversed by $\mathcal{T}_k$ in $R_i$ is recorded as an {inner-region path} of $R_i$.
For example, regions $R_1$ and $R_3$ have inner-region paths $\langle A, J\rangle$ and $\langle G, H \rangle$, respectively.

However, when only using trajectories for constructing region edges, the resulting region graph may not be a connected graph. For example, in Figure~\ref{fig:clusters}, region $R_3$ is not connected with any other regions since no trajectory went through $R_3$ and other regions. Thus, we get the region graph in Figure~\ref{fig:regiongraph}(a). To enable the region graph to serve as a foundation for routing, we need to ensure that the region graph is connected.
To this end, we apply a breadth first search (BFS) based procedure to make the region graph connected.

To ease the following discussion, we call the region edges that are constructed from trajectories {\bf T-edges} and the region edges that are constructed from the BFS procedure {\bf B-edges}.

\textbf{BFS construction of region edges:} We consider the original road network graph $\mathcal{G}$. We conduct a BFS for each vertex $u_i$ in a region $R_i$. When the search reaches a vertex $u_j$ in a different region $R_j$, we stop further exploring $u_j$'s neighbors so that the search does not enter another region $R_k$ via $R_j$. If no T-edge or B-edge exists between regions $R_i$ and $R_j$, we build a B-edge as their region edge. We repeat the same procedure until all vertices in region $R_i$ are traversed.
%
%Here, we only connect isolated regions with their neighbors using B-edges.
The method of obtaining specific paths for B-edges will be discussed in detail in Section~\ref{sec:learntransfer}.

For instance, consider vertex $G$ in region $R_3$ in Figure~\ref{fig:clusters} and the original road network graph in Figure~\ref{fig:intro}. A BFS starting from $G$ visits vertices $A$ and $E$. Since vertex $A$ is in region $R_1$, a region edge $(R_3, R_1)$ is constructed as a B-edge. Similarly, since vertex $E$ is in region $R_5$, a region edge $(R_3, R_5)$ is constructed as a B-edge. The same procedure is applied to the other vertex in region $R_3$, i.e., vertex $H$, but it does not produce any new B-edges. After applying the same procedure to each region, we obtain the final region graph shown in Figure~\ref{fig:regiongraph}(b).

Different from T-edges that are composed by trajectory paths, B-edges have no path information because no trajectories went through the regions connected by the B-edges. To enable routing on top of the region graph, we need to know the paths when traveling between two regions that are connected by B-edges. To this end, in Section~\ref{sec:learntransfer}, we study how to learn and transfer appropriate paths for B-edges.

An alternative way to make the region graph connected is to connect every region pair, i.e., making the region graph fully connected.
However, the BFS based procedure has two benefits. First, it guarantees that there are no disconnected regions. Second, it tries to connect a disconnected region to its nearby regions, which makes the region graph simple. %and thus makes the region graph as simple as possible, which helps the efficiency when routing on the region graph in step 3. %---once a disconnected region is connected with a nearby region, the disconnected region is not connected also with regions that are further away.
%
%For instance, when region $R_3$ is connected with $R_5$ by a B-edge, $R_3$ is not connected with $R_4$ or $R_6$ which must go through $R_5$. %This is beneficial compared to connecting a region with all other regions, since learning appropriate paths to nearby regions is more preferable than learning paths to regions that are further away.
%%
%An extra B-edge, e.g., $(R_3, R_4)$, is redundant and unnecessary due to the following two reasons. First, while traveling from $R_3$ to $R_4$ via $R_5$ is possible, bringing in B-edge $(R_3, R_4)$ increases the complexity of routing based on the region graph. Second, when regions $R_5$ and $R_4$ have been connected by a T-edge, T-edge $(R_5, R_4)$ carries local drivers' routing intelligence better than B-edge $(R_3, R_4)$ does, because the former is associated with paths that occurred in trajectories while the paths associated with the latter are obtained by a learning process.

%%\vspace{-10pt}
\section{Identifying Paths for B-Edges}
%\section{Obtaining routing preferences}
\label{sec:learntransfer}

To enable routing using the region graph, we associate appropriate paths with all B-edges using a three-step method.
First, for each T-edge, we learn a routing preference from the set of paths that are associated with the T-edge, which explains why drivers choose specific paths.
%
%Second, we quantify the similarity between region pairs, which can then be employed as a similarity measure to quantify region edges since each region edge corresponds to a region pair, no matter the region edge is a T-edge or a B-edge.
%
Second, we quantify the similarity between T-edges and B-edges, and then we transfer routing preferences from T-edges to B-edges based on similarity.
Third, we apply the transferred routing preferences to identify appropriate paths for B-edges.
%%\vspace{-0.2cm}
\subsection{Step 1: Learning routing preferences for T-Edges}
\label{ssec:learnDP}

Each T-edge $(R_i, R_j)$ is with a set of paths $\mathbb{P}_{ij}$ (see Section~\ref{subsec:connecting regions}) that connects region $R_i$ to region $R_j$.
We learn a representative routing preference vector $V_{ij}$ for each T-edge $(R_i, R_j)$ that explains why drivers chose the paths in $\mathbb{P}_{ij}$.% in a given \emph{routing preference coordinate system}.
%

%In the paper, we learn a non-personalized preference for a T-edge, which means that the majority of drivers have a certain preference when traveling between two regions. %We demonstrate such preference is effective to model drivers' intelligence in the empirical studies.
%%
%Our framework of learning and transferring routing preference is generic and flexible, and it also supports to learn multiple preferences for a T-edge, such as: a preference vector for every driver or a preference vector for a period, e.g., peak hours. By slightly modifying Step 2 described in Section~\ref{ssec:transferDP}, different kinds of preference vectors can be transferred to B-edges.
%%
%However, how to learn different kinds of multiple preference vectors for T-edges and how to apply them accordingly to generate paths associated with B-edges are out of the scope of the paper. These may be considered as interesting future work.

%\textbf{routing preference Coordinate System: }

We consider two categories of features that may affect a driver's travel decisions---travel costs and road conditions.
Travel cost features describe the travel costs that drivers want to minimize.
Road condition features describe drivers' preferences or restrictions relating to road conditions.
%
%Again, we emphasize that the proposed techniques are capable of incorporating additional categories, e.g., numbers of traffic lights and specific turns such as left turns, right turns, and U-turns, into the preference vectors.
%
For example, we may consider three different travel cost features, \textsf{travel time (TT)}, \textsf{distance (DI)}, and \textsf{fuel consumption (FC)}; and three road condition features, e.g., \textsf{highways}, \textsf{residential roads}, and \textsf{highways and residential roads}. %, \textsf{avoid left turns}.

Based on the above, we use a 2-dimensional vector to represent a routing preference, where the so-called master dimension corresponds to travel cost features and the so-called slave dimension corresponds to road condition features. For example, vector $V=\langle$TT, Highway$\rangle$ indicates a preference for minimizing travel time and using highways.

%% Bin: this is not very consistently aligned with master/slave preferences. may need rewriting.
%For example, we may consider routing preference vectors in a 8-dimensional space, where the first 3 dimensions
%%For example, we may consider three different travel costs---
%correspond to \textsf{travel time (TT)}, \textsf{distance (DI)}, and \textsf{fuel consumption (FC)}, which are travel cost features; and the remaining five dimensions correspond to road condition features, e.g., %different road conditions---
%% which are travel cost features; and the remaining 5 dimensions correspond to road condition features, e.g.,
%\textsf{prefer highways}, \textsf{avoid highways}, \textsf{prefer residential roads}, \textsf{prefer both highways and residential roads}, \textsf{avoid left turns}. Then vector $\langle 1$, $0$, $0$, $1$, $0$, $0$, $0$, $0\rangle$ denotes a routing preference that favors fastest paths and has a preference for highways.
%
%To simplify the following discussion, we consider a
%
%we also transfer the vector as $\langle$TT, Highway$\rangle$, indicating that drivers prefer to minimize travel time and use highways.

Based on the routing preference model, we aim at identifying an appropriate preference vector for the T-edge $(R_i, R_j)$ based on its path set $\mathbb{P}_{ij}$.
Given the source and destination of a path $P_k \in \mathbb{P}_{ij}$ and a preference vector $V$, we are able to construct a path $P_k^V$ based on $V$ that connects the source and  destination of $P_k$. If $V$ captures the driver's preferences well, path $P_k^V$ should largely match the actual, or ground truth, path $P_k$.
Thus, we aim to identify a routing preference vector $V^*$ such that the constructed paths match the paths in $\mathbb{P}_{ij}$ as much as possible. Equivalently, we aim at solving the optimization problem %shown in Equation~\ref{eq:objectivefunction}.
%
%%\vspace{-5pt}
%\begin{equation}
%\label{eq:objectivefunction}
$V^*=\arg\max_{V\in \mathbb{V}} \sum_{P_k \in \mathbb{P}_{ij}} \mathit{pSim}(P_k, P_k^V)$,
%$V^*=\max_{V\in \mathbb{V}} \sum_{P_k \in \mathbb{P}_{ij}} \mathit{pSim}(P_k, P_k^V)$,
%%%\vspace{-5pt}
%\end{equation}
%
where $\mathbb{V}$ is a set of possible vectors and $\mathit{pSim}(\cdot, \cdot)$ is a path similarity function that evaluate the similarity between two paths.
%
%For example, assume that we identify a preference vector $V_k=\langle \mathit{TP1}, \mathit{TT} \rangle$ for path $P_k$. This means that the constructed path $P_k^{V_k}$, i.e., the fastest path (according to $\mathit{TT}$) with a preference for highways (according to $\mathit{TP1}$), matches the ground truth path $P_k$ better than the paths constructed using other preference vectors, such as preference $V_k^\prime=\langle \mathit{TP2}, \mathit{DI} \rangle$, indicating the shortest path with a preference for residential roads. However, path $P_k^{V_k}$ that is constructed by $V_k=\langle \mathit{TP1}, \mathit{TT} \rangle$ may still not match $P_k$ fully, although $V_k$ provides the best match among all other possible preference vectors in the given preference coordination system.

We use a popular path similarity function~\cite{DBLP:journals/vldb/0002GMJ15,erkut1998modeling}: % $\mathit{pSim}({P}_k, P_k^{V}) = \frac{\sum_{e\in P_k \cap P_k^{V}  } \mathit{length}(e)}{\sum_{e\in P_k}\mathit{length}(e)}$.
%
%%\vspace{-5pt}
%\begin{equation}
%\label{eq:psim}
%\small
\begin{equation}
\label{eq:psim}
\mathit{pSim}({P}_k, P_k^{V}) = \frac{\sum_{e\in P_k \cap P_k^{V}  } \mathit{len}(e)}{\sum_{e\in P_k}\mathit{len}(e)}.
%%\vspace{-5pt}
\end{equation}
%%
%where $\mathit{len}(e)$ returns the length of edge $e$.
%%
The intuition %Equation~\ref{eq:psim}
is two-fold: first, the more edges the constructed path $P_k^V$ shares with the ground-truth path $P_k$, the more similar the two paths are; second, the longer the shared edges are, the more similar the two paths are.

A naive way of solving the optimization problem is to search the whole space, i.e., all combinations of features in the master and slave dimensions. However, the search space can be very large, thus rendering the learning algorithm inefficient.
We instead
propose an efficient learning algorithm that is inspired by coordinate descent. In short, we first identify the best travel cost feature in the master dimension, and next, based on the chosen travel cost feature, we identify the best road condition features in the slave dimension. %asdas sad asd as asd
%we start from the travel cost features and then further identify the best feature from the road condition features.

Specifically, given the source and destination of each ground truth path $P_k$, we obtain a lowest-cost path using each cost type.
This yields three lowest-cost paths $\hat{P}_k^{\mathit{DI}}$, $\hat{P}_k^{\mathit{TT}}$, and $\hat{P}_k^{\mathit{FC}}$, for distance, travel time, and fuel consumption~\cite{DBLP:journals/geoinformatica/Guo0AJT15,DBLP:conf/gis/GuoM0JK12}, respectively.
We then measure the similarity between path $P_k$ and each of the three lowest-cost paths and choose the optimal cost type whose corresponding lowest-cost path has the highest similarity. %
Next, we identify the optimal road condition feature. For each road condition feature, we compute a new lowest-cost path based on the optimal cost type while making sure that the road condition feature is also satisfied. We check if the similarity between the new path and the ground truth path $P_k$ can be further improved.
The road condition feature that gives the largest improvement is chosen as the optimal road condition feature.

For example, if the optimal cost type is distance, we test if the shortest path with preferences for highways or residential roads can yield a higher similarity compared to shortest path without any road type preferences. If so, we choose the road type that gives the largest similarity improvements. Otherwise, all the road condition features are ignored.

\begin{figure}[h]
	\centering
\subfigure{
	\begin{minipage}[t]{0.22\textwidth}
		\includegraphics[width=1\textwidth]{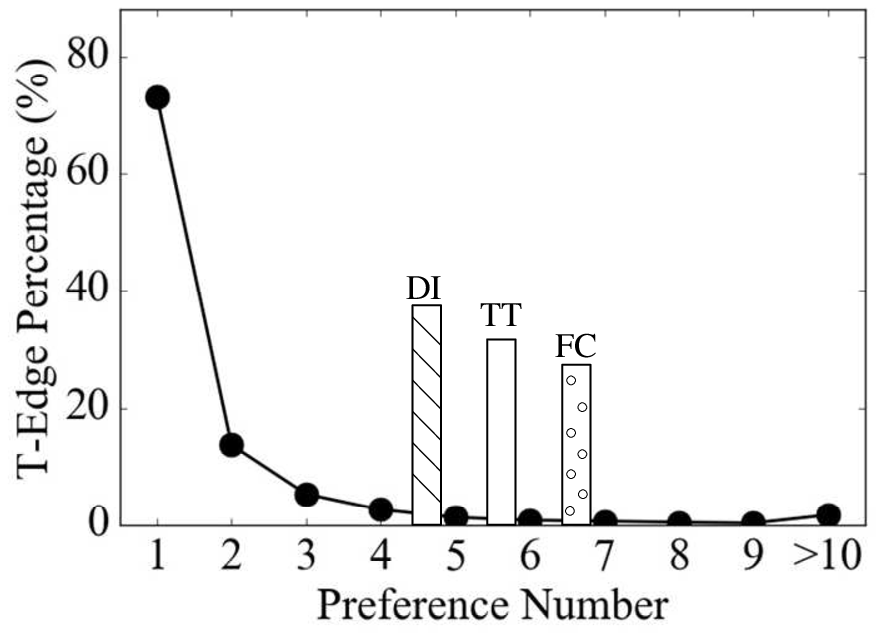}
       %\centering \footnotesize{(a) \# of Unique Preferences per T-edge}\label{fig:rp_vs_prefnum}
       \centering \footnotesize{(a) Distribution of Preferences}\label{fig:rp_vs_prefnum}
	\end{minipage}	
}
\subfigure{
	\begin{minipage}[t]{0.22\textwidth}
		\includegraphics[width=1\textwidth]{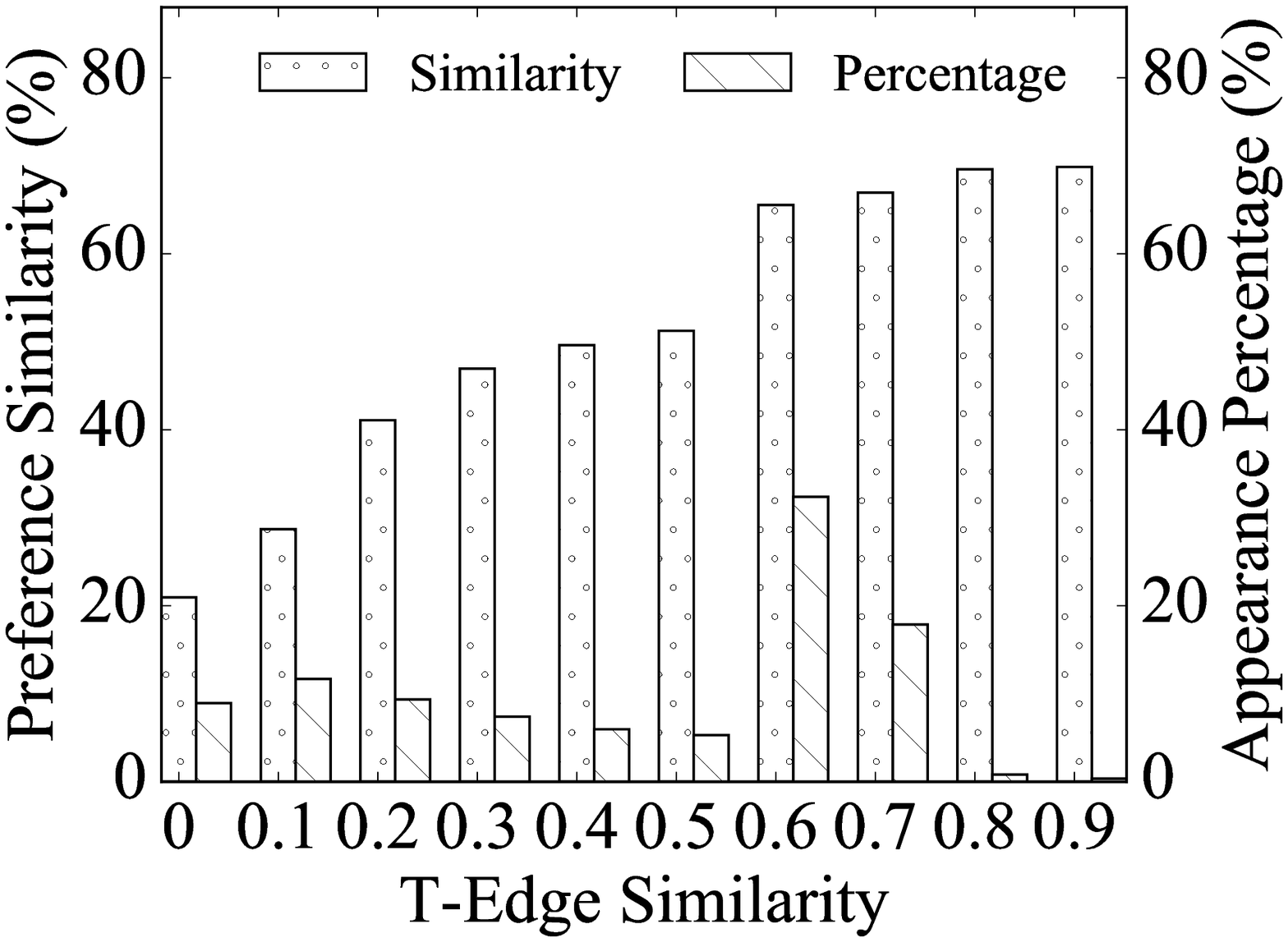}
       \centering \footnotesize{(b) vs. Preference Similarity}\label{fig:rp_vs_prefsim}
	\end{minipage}	
}
%%\vspace{-0.15cm}
	\caption{Statistical Evidences for Design Choices}
	\label{fig:rp_vs_pref}
%%\vspace{-0.15cm}
\end{figure}

Next, we provide statistical evidence to justify our design choice of choosing only a single representative preference for each T-edge.
Given a T-edge $(R_i, R_j)$, we learn a routing preference for each path in $\mathbb{P}_{ij}$, and we count the number of unique preferences.
The curve in Figure~\ref{fig:rp_vs_prefnum} shows that for more than 70\% of all T-edges, we obtain a single preference, although multiple paths often exist in $\mathbb{P}_{ij}$.
Thus, we chose to learn a single routing preference for each T-edges. On the other hand, Figure~\ref{fig:rp_vs_prefnum} also suggests that it is possible that a T-edge has more than one preference---we leave the modeling multiple preferences per T-edge as future research. % left as future work.
%
%The result (shown in supplementary document~\cite{supp}) shows that we obtain a single preference for more than 70\% of all T-edges, and two preferences for 14\% of those, although multiple paths often exist in $\mathbb{P}_{ij}$. Thus, we chose to learn a single routing preference for each T-edges. On the other hand, the result also suggests that it is possible that a T-edge has more than one preference---we leave the modeling multiple preferences per T-edge as future research. % left as future work.
%%\vspace{-0.1cm}

We also show the distribution of the learned routing preferences as bars in Figure~\ref{fig:rp_vs_prefnum}.
We aggregate more than 200 unique routing preferences based on their travel cost features, i.e., DI, TT, and FC.
The bars show that the
routing preferences are distributed almost uniformly, indicating that T-edges do have different routing preferences.
%
%\vspace{-0.2cm}
\subsection{Step 2: Transferring routing preferences}
\label{ssec:transferDP}

So far, we have identified preference vectors for T-edges. The next step is to associate preference vectors with B-edges, which can then be used to identify appropriate paths for B-edges.
To this end, we transfer the routing preferences of T-edges to similar B-edges, which follows the intuition that when two region edges are similar, they also have similar routing preferences. %
For example, if most local drivers choose the fastest paths with a preference for main roads to travel between a region in the city center and a northern suburb residential area, it is also likely that local drivers have this preference when traveling between another region in the city center and a southern suburb residential area.

%This is intuitive since road hierarchies in the same country often follow the same design standard.
% Bin: somewhere we need to say learn a single preference from all historical trajectories, and applying the single preference globally to identify paths for all B-edges is not a good idea. This is to justify region level preference transferring is a good choice.

% functions are appropriate. In particular,
%(1) region edges may have quite different preferences; and (2) similar region edges have similar preferences. %The

Based on the above intuition, we first introduce the similarity function that quantifies the similarity between two region edges and then provide an algorithm that transfers routing preferences between similar region edges.

\textbf{Similarity between two region edges: } Any region edge, a T-edge or a B-edge, connects two regions.
A region edge is described by the features of its two regions.
In particular, we use two elements $\mathit{dis}$ and $\mathbb{F}$ to describe a region edge $re$.
%In particular, we use a vector $re = \langle\mathit{dis}, \mathbb{F}\rangle$ to describe the feature of a region edge. Recall that the vector $V$ in Section \ref{ssec:learnDP} models the routing preference on the region edge. Thus, the assumption is formalized as that higher similarity of two region edge features $\mathit{re_i}$ and $\mathit{re_j}$ leads to higher similarity of two routing preferences $V_i$ and $V_j$ on the two region edges.

Element $\mathit{re.dis}$ is a real value, indicating the Euclidean distance between the centroids of the two regions connected by the region edge. %, which suggests how far the two regions are apart from each other.
The distance information is an influential factor when drivers choose their paths. For example, drivers may prefer the fastest paths if they travel long distances, but they may prefer the shortest paths when traveling at shorter distances. % if the difference of travel time is not obvious among alternative paths.

Next, element $re.\mathbb{F}$ describes the functionalities of the two regions. Element $re.\mathbb{F}$ is also essential because, for example, when traveling between two business districts and between a residential area and a city center, drivers may have different preferences.
In particular, we use a set of road types to describe the functionality of a region~\cite{forbes1999urban}. For each region, we consider all edges that are incident to the vertices in the region and select top-$k$ road types of the edges as the region's road type set. For example, regions $R_i$ and $R_j$ have top-2 road type sets $\{$\textsf{TP1}, \textsf{TP2}$\}$ and $\{$\textsf{TP3}, \textsf{TP4}$\}$, respectively. Then, region edge $(R_i$, $R_j)$ has element $\mathit{re.}\mathbb{F}$ that is the Cartesian product of the road type sets from both regions: $\mathit{re.}\mathbb{F}=\{$$\langle$\textsf{TP1, TP3}$\rangle$, $\langle$\textsf{TP1, TP4}$\rangle$, $\langle$\textsf{TP2, TP3}$\rangle$, $\langle$\textsf{TP2, TP4}$\rangle$$\}$.
%
%The function features can represent many different aspects. For example, we can use the categories of points-of-interests (POIs) in the regions to describe the function features. If one region's top-2 POIs are \textsf{cinema} and \textsf{bar}, and the other region's top-2 POIs are \textsf{office} and \textsf{fastfood} then $\mathbb{F}=\{$\textsf{cinema-office}, \textsf{cinema-fastfood}, \textsf{bar-office}, \textsf{bar-fastfood}$\}$.
%
%Alternatively,
%For example, a residential area should have predominantly residential roads and no highways.
%
%For example, if one region has TP1 and TP2 edges and the other region has TP2 and TP3 edges, we have $\mathbb{F}=\{$\textsf{TP1-TP2}, \textsf{TP1-TP3}, \textsf{TP2-TP2}, \textsf{TP2-TP3}$\}$.
%
% Bin: feature set F is redefined. - sign is avoided.

Based on the above, the similarity between two region edges $re_i$ and $re_j$, quantified by the similarity of their feature vectors, is defined as follows. %in Equation~\ref{eq:resim}.
%\begin{equation}
%\vspace{-5pt}
%\begin{small}
\[
\small
\label{eq:resim}
\mathit{reSim}(re_i, re_j)= \frac{\min(re_i.\mathit{dis}, re_j.\mathit{dis})}{\max(re_i.\mathit{dis}, re_j.\mathit{dis})} + \mathit{J}(re_i.\mathbb{F}, re_j.\mathbb{F}).
%\vspace{-5pt}
\]
%\end{small}
%end{equation}
%
%The similarly function is a linear combination of distance similarity and region function similarity. %, controlled by a relative importance parameter $\alpha$.
The similarity function is the sum of distance similarity and region function similarity. %, controlled by a relative importance parameter $\alpha$.
For distance similarity, the more similar the two distances are, the larger the similarity is. This captures the intuition that travels between equally far apart regions may tend to have similar routing preferences. % with traveling between another further apart regions but may have different routing preference with traveling between closeby regions.
For region function similarity, we use Jaccard similarity to evaluate the similarity between the region functions. If the two region edges share more function features, meaning that they connect similar region pairs, travels on the two region edges are expected to have similar routing preferences.

To justify design choices, that (i) similar region edges have similar routing preferences and that (ii) the proposed region edge similarity function $\mathit{reSim}(\cdot, \cdot)$ is effective, we show the results of an experiment using preferences learned from T-edges in Figure~\ref{fig:rp_vs_prefsim}.
%
%Figure~\ref{fig:rp_vs_prefnum} shows that none of preferences is learned dominantly across T-edges (see supplementary document~\cite{supp} for details) and
First, the ``Similarity'' bars show that similar T-edges have similar routing preferences, while dissimilar T-edges have dissimilar routing preferences. Second, the ``Percentage'' bars show the percentages of T-edge pairs that fall in a different T-edge similarity ranges. There are many similar (e.g., similarity above 0.5) T-edges, although there are few highly similar (e.g., similarity above 0.9) T-edges. This makes it possible to transfer routing preferences among region edges, and these observations indicate that the design choices are purposeful.% They together justify the design choices.

%evidences also prevent another design choice where we learn a single preference from historical trajectories and then assigning the single preference globally to all B-edges.

\noindent
\textbf{Transferring preferences among similar region edges: }We adopt the idea of graph-based transduction learning~\cite{DBLP:conf/pkdd/TalukdarC09,DBLP:conf/cikm/WangYQSW11} to transfer routing preferences from T-edges to similar B-edges.
First, we build a undirected, weighted graph, where a vertex represents a region edge, which can be a T-edge or a B-edge. Given a total of $n$ region edges, we use an adjacency matrix $\mathbf{M}\in \mathbb{R}^{n\times n}$ to record the edge weights of the graph. Specifically, $\mathbf{M}[i,j]$ equals to the similarity $\mathit{reSim}(re_i, re_j)$ between region edges $re_i$ and $re_j$, where $1 \leqslant i, j \leqslant n$. % and $i\neq j$.

Next, we introduce an adjacency matrix reduction threshold $\mathit{amr}$. In the adjacency matrix, we only keep the values that exceed $\mathit{amr}$; otherwise, we set the values to 0. This way, the adjacency matrix only captures ``sufficiently'' similar region edge pairs, which enables to control the accuracy of the transferred preferences. The less dense resulting matrix also improves efficiency (see Figure~\ref{fig:exp_lp_sim} in experiments). %Meanwhile, it also makes the adjacency matrix sparser which also improves efficiency.

\begin{figure}[h]
\centering
  \includegraphics[width=0.85\columnwidth]{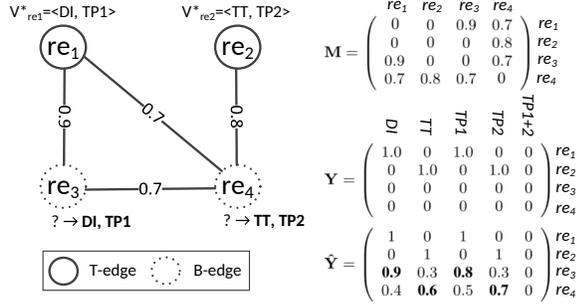}
\caption{Transferring routing preferences}
\label{fig:lpgraph}
\end{figure}

Figure~\ref{fig:lpgraph} shows a graph with four vertices (i.e., $n=4$) representing two T-edges and two B-edges.
The corresponding matrix $\mathbf{M}$ is also shown. For example, $\mathbf{M}[1,3]=0.9$ indicates that the similarity between $re_1$ and $re_3$ is 0.9, and $\mathbf{M}[2,3]=0$ indicates that the similarity between $re_2$ and $re_3$ is smaller than threshold $\mathit{amr}$.

In the next step, we use a matrix $\mathbf{Y}\in \mathbb{R}^{n \times p}$ to denote
the initial routing preferences of different region edges. Here, $n$ is the total number of region edges, including T-edges and B-edges, and %$p$ is the total number of valid routing preferences from T-edges.
$p$ is the total number of travel cost and road condition features that are used for modeling routing preferences in Section~\ref{ssec:learnDP}. % preferences from T-edges.

To illustrate, we consider two travel cost features DI and TT and three road condition features indicating preferences on road type TP1, TP2, and both, i.e., TP1+2.
In this setup, matrix $\mathbf{Y}$ has $p=5$ columns that represent features DI, TT, TP1, TP2, and TP1+2.

Each row in $\mathbf{Y}$ corresponds to a region edge's routing preference.
For a T-edge, the features corresponding to its learned routing preference $V^*$ are set to 1.
For example, assuming that T-edge $re_1$ has preference vector $V_{re_1}^*=\langle$DI, TP1$\rangle$, the first row of $\mathbf{Y}$ is set to $(1, 0, 1, 0, 0)$.
Similarly, if T-edge $re_2$ has preference vector $V_{re_2}^*=\langle$TT, TP2$\rangle$, the second row is set to $(0, 1, 0, 1, 0)$, as shown in Figure~\ref{fig:lpgraph}.
Next, since the routing preferences of the B-edges are unknown, the rows that represent B-edges are set to $(0, 0, 0, 0, 0)$. %, as shown in Figure~\ref{fig:lpgraph}.

The transduction learning yields matrix $\hat{\mathbf{Y}}\in
\mathbb{R}^{n \times p}$ that records the transferred routing preferences for the B-edges.
Specifically, $\hat{\mathbf{Y}}[i, j]$ indicates the
probability of region edge $re_i$ having the $j$-th routing preference feature.
For B-edge $re_i$, the travel cost feature with the largest
probability, i.e., $\arg\max_{x\in\{1, 2\}}\hat{\mathbf{Y}}[i, x]$, is used
as the final travel cost feature.
In the example $\hat{\mathbf{Y}}$ in Figure~\ref{fig:lpgraph}, this is DI for $re_3$ and TT for $re_4$. % according to the example $\hat{\mathbf{Y}}$ shown in Figure~\ref{fig:lpgraph}.
The road type feature with the largest probability, i.e., $\arg\max_{x\in\{3, 4, 5\}}\hat{\mathbf{Y}}[i, x]$, is used as the final road type feature.
In the example $\hat{\mathbf{Y}}$ in Figure~\ref{fig:lpgraph}, this is TP1 for $re_3$ and TP2 for $re_4$. %, according to $\hat{\mathbf{Y}}$ in Figure~\ref{fig:lpgraph}.
Finally, B-edges $re_3$ and $re_4$ obtain the transferred routing preferences $V_{re_3}=\langle$DI, TP1$\rangle$
and $V_{re_4}=\langle$TT, TP2$\rangle$, respectively.

Now the remaining question is how to obtain $\hat{\mathbf{Y}}$, which is the core of the transduction learning.

\noindent
\textbf{Obtain matrix $\hat{\mathbf{Y}}$: }
We obtain $\hat{\mathbf{Y}}$ by minimizing the following objective
function
%\vspace{-5pt}
\[
\small
O(\hat{\mathbf{Y}})=\sum_{x=1}^{p}\Big[\underbrace{(\mathbf{Y}_{\cdot x}-\hat{\mathbf{Y}}_{\cdot x})^{\mathrm{T}}\mathbf{S}(\mathbf{Y}_{\cdot x}-\hat{\mathbf{Y}}_{\cdot x})}_{\text{Keeping T-edges' preferences}}
%\vspace{-5pt}
\]
\begin{equation}
\label{eq:objective}
\small
    ~~~~~~~~~~~~~~~~~~~~~~~~~~~~~~~~~+ \underbrace{\mu_1 \hat{\mathbf{Y}}_{\cdot x}^{\mathrm{T}} \mathbf{L} \hat{\mathbf{Y}}_{\cdot x}}_{\text{Transferring preferences to B-edges}} + \underbrace{\mu_2 ||\hat{\mathbf{Y}}_{\cdot x}||_2^2}_{\text{Regularization}}\Big],
    %\vspace{-5pt}
\end{equation}
%
%\[
%\small
%O(\hat{\mathbf{Y}})=\sum_{x=1}^{p}\Big[\underbrace{(\mathbf{Y}_{\cdot x}-\hat{\mathbf{Y}}_{\cdot x})^{\mathrm{T}}\mathbf{S}(\mathbf{Y}_{\cdot x}-\hat{\mathbf{Y}}_{\cdot x})}_{\text{Keeping T-edges' preferences}} + \underbrace{\mu_1 \hat{\mathbf{Y}}_{\cdot x}^{\mathrm{T}} \mathbf{L} \hat{\mathbf{Y}}_{\cdot x}}_{\text{Transferring preferences to B-edges}} + \underbrace{\mu_2 ||\hat{\mathbf{Y}}_{\cdot x}||_2^2}_{\text{Regularization}}\Big],
%\]
%
%\[
%    + \underbrace{\mu_1 \hat{\mathbf{Y}}_{\cdot x}^{\mathrm{T}} \mathbf{L} \hat{\mathbf{Y}}_{\cdot x}}_{\text{Transferring preferences to B-edges}} + \underbrace{\mu_2 ||\hat{\mathbf{Y}}_{\cdot x}-\mathbf{R}_{\cdot x}||_2^2}_{\text{Regularization}}\Big],
%\]
%
where $\mathbf{Y}_{\cdot x}$ and $\hat{\mathbf{Y}}_{\cdot x}$ indicate the $x$-th column of matrices $\mathbf{Y}$ and $\hat{\mathbf{Y}}$, respectively. Hyper-parameters
$\mu_1$ and $\mu_2$ control the relative influences of the second and third terms in the objective function, respectively.
%
%where $\mathbf{Y}_{\cdot x}$, $\hat{\mathbf{Y}}_{\cdot x}$,
%and $\mathbf{R}_{\cdot x}$ indicate the $x$-th column of the
%corresponding matrices.

The intuition of each term of the objective function is as follows. %three-fold, as it has three terms.
First, the T-edges should keep the routing preferences that are learned in step 1.
The T-edges' learned routing preferences serve as training data in the transduction learning process.

Matrix $\mathbf{S}\in \mathbb{R}^{n \times n}$ is an auxiliary matrix that indicates which region edges are T-edges. In particular, we organize the region edges such that the first $x$ edges are T-edges and the remaining $n-x$ edges are B-edges. Then, $\mathbf{S}$ is a diagonal matrix, where the first $x$ diagonal entries are set to 1 and the remaining diagonal entries are set to 0.
Specifically, we have
$\begin{tiny}\mathbf{S}=\left(
  \begin{array}{cccc}
    1 & 0 & 0 & 0 \\
    0 & 1 & 0 & 0 \\
    0 & 0 & 0 & 0\\
    0 & 0 & 0 & 0 \\
  \end{array}
\right)
\end{tiny}$ in our example because $re_1$ and $re_2$ are T-edges.

Based on $\mathbf{S}$, the first term actually computes the sum of the squared differences between $\mathbf{Y}_{\cdot x}$ and $\hat{\mathbf{Y}}_{\cdot x}$ of the rows that represents T-edges. By minimizing the first term, we try to identify a $\hat{\mathbf{Y}}$ that minimizes the difference. This means that the T-edges should try to keep their learned preferences from step 1. On the other hand, the first term does not pose any constraints between $\mathbf{Y}_{\cdot x}$ and $\hat{\mathbf{Y}}_{\cdot x}$ of the rows that represents B-edges.

Second, the T-edges' routing preferences are transferred to B-edges. The transfer process ensures that the more similar the two region edges are, the more similar their routing preferences are. This is realized by the use of the unnormalized graph Laplacian matrix $\mathbf{L}$ in the second term of Equation~\ref{eq:objective}.
In particular, $\mathbf{L}=\mathbf{D}-\mathbf{M}$, where $\mathbf{M}$ is the adjacency matrix and $\mathbf{D}$ is a diagonal matrix where $\mathbf{D}[i,i]=\sum_{k\in\{1\ldots n\}} \mathbf{M}[i,k]$ and $\mathbf{D}[i,j]=0$ if $i\neq j$.
%
%Following the running example shown in Figure~\ref{fig:lpgraph},
In our example, we have \\
$\begin{tiny}
\mathbf{D}=\left(
  \begin{array}{cccc}
    1.6 & 0 & 0 & 0 \\
    0 & 0.8 & 0 & 0 \\
    0 & 0 & 1.6 & 0\\
    0 & 0 & 0 & 2.2 \\
  \end{array}
\right)
\end{tiny}$
$\begin{tiny}
\mathbf{L}=\left(
  \begin{array}{cccc}
    1.6 & 0 & -0.9 & -0.7 \\
    0 & 0.8 & 0 & -0.8 \\
    -0.9 & 0 & 1.6 & -0.7\\
    -0.7 & -0.8 & -0.7 & 2.2 \\
  \end{array}
\right)
\end{tiny}$

With the help of $\mathbf{L}$, the second term actually computes the sum of the products of the similarities of two region edges and the differences of the two region edges' corresponding routing preferences. When the similarity of the two region edges is high, a small difference between their routing preferences make the product significant.
Minimizing the second term in the objective function  has the effect of smoothly spreading the routing preferences from T-edges to B-edges such that (1) two region edges with high similarities have highly similar routing preferences; (2) two region edges with low similarities may have dissimilar routing preferences.

%The intuition Matrix $\mathbf{L}$ is a graph Laplacian matrix
%derived from the graph's adjacency matrix $\mathbf{L}$.

Third, we conduct L2 regularization~\cite{DBLP:conf/pkdd/TalukdarC09,DBLP:conf/cikm/WangYQSW11}  to avoid over-fitting.

Next, we need to minimize the objective function. By differentiating Equation~\ref{eq:objective} by $\hat{\mathbf{Y}}_{\cdot x}$ and then setting it to 0, we get
%\vspace{-5pt}
\begin{equation}
\label{eq:solve}
%\small
(\mathbf{S}+\mu_1 \mathbf{L} +\mu_2 \mathbf{I}) \hat{\mathbf{Y}}_{\cdot x} = \mathbf{S} \mathbf{Y}_{\cdot x}
%\vspace{-5pt}
\end{equation}
Using basic linear algebra practice, Equation~\ref{eq:solve} can be solved by iterative approximation algorithms~\cite{golub2012matrix}, e.g., the Jacobi method~\cite{DBLP:conf/pkdd/TalukdarC09} or the conjugate gradient method~\cite{DBLP:journals/tkde/YangKJ14}.
We need to solve Equation~\ref{eq:solve} $p$ times; and each time, we obtain a $\hat{\mathbf{Y}}_{\cdot x}$ where $x\in\{1, 2, \ldots, p\}$.
Finally, we obtain $\hat{\mathbf{Y}}$.

%intuition is to keep the labels for seed trajectories;
%to spread labels over the graph while ensuring that similar trajectories
%(evaluated by Equation~\ref{eq:sim}) obtain similar labels; and to do
%regularization to avoid over-fitting.
%
%Specifically,
%%
%How to construct the auxiliary matrices $\mathbf{S}$ and $\mathbf{R}$,
%and how to choose appropriate values for hyper-parameters $\mu_1$ and
%$\mu_2$ are beyond the scope of the paper and is covered
%elsewhere~\cite{DBLP:conf/cikm/WangYQSW11}.

%\newpage
%
%
%$\mathbf{M}=\left(
%  \begin{array}{cccc}
%    1.0 & 0 & 0.9 & 0.7 \\
%    0   & 1.0  & 0 & 0.8 \\
%    0.9 & 0 & 1.0 & 0.7 \\
%    0.7 & 0.8 & 0.7 & 1.0 \\
%  \end{array}
%\right)$
%
%
%%\vspace{30pt}
%
%$\mathbf{Y}=\left(
%  \begin{array}{ccccc}
%    {1.0} & 0 & {1.0} & 0 & 0 \\
%    0 & {1.0} & 0 & {1.0} & 0\\
%    0 & 0 & 0 & 0 & 0\\
%    0 & 0 & 0 & 0 & 0\\
%  \end{array}
%\right)$
%
%%\vspace{5pt}
%
%$\mathbf{\hat{Y}}=\left(
%  \begin{array}{ccccc}
%    {1} & 0 & {1} & 0 & 0 \\
%    0 & {1} & 0 & {1} & 0\\
%    \textbf{0.9} & 0.3 & \textbf{0.8} & 0.3 & 0\\
%    0.4 & \textbf{0.6} & 0.5 & \textbf{0.7} & 0\\
%  \end{array}
%\right)$

\subsection{Step 3: Applying Transferred Preferences}
\label{ssec:applyDP}

After step 2, each B-edge has a transferred preference vector. For each B-edge, we now identify a few appropriate paths according to its transferred preference vector.
Consider B-edge $(R_i, R_j)$. Recall that a region has transfer centers where trajectories enter and leave the region (see Section~\ref{subsec:connecting regions}). For each pair of a transfer center from $R_i$ and a transfer center from $R_j$, we identify a path according to the preference vector. Finally, the identified paths are associated with B-edge $(R_i, R_j)$.

We proceed to modify Dijkstra's algorithm to accommodate the preference, as shown in Algorithm~\ref{algo:dij}.
To ease the discussion, we assume that a B-edge is associated with a transferred routing preference vector $\langle$DI, TP1$\rangle$, meaning that minimizing travel distance and using road type TP1 are preferred. Recall that the first dimension is master dimension and the second dimension is the slave dimension. %} (or {\it preference}).

The overall procedure is similar to the classical Dijkstra's algorithm.
In the algorithm, each vertex is associated with two attri\-butes---a cost attribute that records the cost of travel from the source to the vertex and a parent attribute that records the parent vertex of the vertex. And we use a priority queue $Q$ to control the order of visiting different vertices (lines 1--4).

Here, the cost value maintained in a vertex corresponds to the specific cost type feature for the master dimension of a given preference vector. For example, when considering preference vector $\langle$DI, TP1$\rangle$, each vertex is associated with a cost that equals to the distance (according to DI) from the source vertex to the vertex.

%Thus, when exploring different paths, the weight of an edge is also according to the distance of an edge. represented by $w_{\mathit{master}}(u, x)$ in lines 10--11.

The algorithm always chooses the vertex with the lowest cost, say vertex $u$, to continue exploring (line 6). When exploring from $u$, we differentiate two cases (lines 7--14): (i) at least one edge $(u, x)$ satisfies the slave preference, and (ii) no edge $(u, x)$ exists that satisfies the slave preference.
For case (i), only edges that satisfy slave preference are explored.
For case (ii), all $u$'s adjacent vertices are explored.
This way, we make sure that the preferences on both the master and slave dimensions are accommodated by the algorithm.

%For example, according to $p=\langle$DI, TP1$\rangle$, if edge $(u, x)$ is a TP1 road, edge $(u, x)$ should be explored. And according to the preference on the master dimension, i.e., DI, weight function $w_{\mathit{master}}(u, x)$ returns the length of edge $(u, x)$.

%%\vspace{-10pt}
\begin{algorithm}[!htb]
\caption{ApplyingPreferencesModifiedDijkstra}
\label{algo:dij}
\begin{small}
\KwIn{Preference Vector: $V=\langle \mathit{master}, \mathit{slave}\rangle$; Source and destination vertices: $v_s$, $v_d$; Road Network: $\mathcal{G}$; }
\KwOut{Path $P$ that connects $v_s$ and $v_d$}

\For{each vertex $v\in \mathcal{G}.\mathbb{V}$}
{
    $v.\mathit{cost} \leftarrow +\infty$;
    $v.\mathit{parent}\leftarrow \mathit{null}$;\\
}
$v_s.\mathit{cost} \leftarrow 0$;\\

Initialize a priority queue $Q$ and add all vertices to $Q$; \\

\While{$v_d$ is still in $Q$}
{
   vertex $u \leftarrow Q.\mathit{extractMin}()$; \\
   Boolean $\mathit{noneSat} \gets$ \emph{false};  \\
    \If{there does not exist a vertex $x$ such that x is $u$'s adjacent vertex and edge $(u, x)$'s road type satisfies $V.\mathit{slave}$}
   {
      $\mathit{noneSat} \gets$ \emph{true};
   }
   \For{each vertex $x$ that is adjacent $u$}
   {
        \If{edge $(u, x)$'s road type satisfies $V.\mathit{slave}$ $\vee \mathit{noneSat}$}
        {
            \If{$u.\mathit{cost} + w_{\mathit{V.master}}(u, x)< x.\mathit{cost}$}
            {
                $x.\mathit{cost} \leftarrow u.\mathit{cost} + w_{\mathit{V.master}}(u, x)$;\\
                $x.\mathit{parent} \leftarrow u$;
            }
        }
   }
}
Construct $P$ from $v_d$ using the parent attributes and return $P$;\\
%\Return $V$;% \langle m^\prime, s_1^\prime, s_2^\prime, \ldots, s_X^\prime \rangle$;
\end{small}
%%\vspace{-10pt}
\end{algorithm}
%%\vspace{-10pt}

The three steps yield a region graph where each region edge has a set of paths, meaning that the region graph can serve as a foundation for routing.

\section{Routing on Region Graphs}

Given an arbitrary pair of a source $v_s$ and a destination $v_d$ in the original road network graph $\mathcal{G}$, we present a routing algorithm that is able to recommend a path connecting them, using the region graph.
We distinguish two cases.

\textbf{Case 1:} Vertex $v_s$ is in a region, say $R_s$, and vertex $v_d$ is also in a region, say $R_d$.
If both vertices are in the same region, i.e., $R_s = R_d$, since we maintain inner-region paths inside regions, we check if trajectories exist that traverse from $v_s$ to $v_d$. If yes, we return a path with the largest number of trajectory traversals; if no, we return the fastest path.

If the vertices are not in the same region, i.e., $R_s \neq R_d$, we first identify a region path based on the region graph and then map the region path to a path in the original road network.
%
%We first present a routing algorithm on the region graph ${G}_R$ that identifies a region path that connect $R_s$ and $R_d$. %in Algorithm~\ref{algo:regiongraph}.
%

\textbf{Routing on the region graph:} The intuition is to find a region path that follows fewer region edges to reach the destination region $R_d$.
This is because if a region path consists of many region edges, it involves the stitching of many paths from different trajectories, which may not represent coherent routing preferences.
Thus, in the routing procedure, we always prefer to follow a region edge that enables us to go to a region that is geometrically close to the destination region.
When a region edge exists that directly connects $R_s$ and $R_d$, we always use that region edge.
Otherwise, we give higher priorities to the region edges that lead to regions that are closer to the destination region $R_d$.

To illustrate, consider the region graph shown in Figure~\ref{fig:regiongraph}(b) and assume the physical locations of the regions are also represented in Figure~\ref{fig:regiongraph}(b).
Assume that regions $R_1$ and $R_4$ are given as the source and destination regions.
When exploring from $R_1$, region $R_5$ is preferred over regions $R_2$, $R_3$ and $R_6$ because $R_5$ is much closer to destination region $R_4$.
%Compared to regions $R_3$ and $R_6$, region $R_5$ is preferred to be explored from source region $R_1$ since $R_5$ is much closer to destination region $R_4$.
%
Finally, the region path $\langle (R_1$, $R_5)$, $(R_5$, $R_4) \rangle$ is returned.

Recall that each region edge corresponds to some paths in the original road network graph.
Based on this, a region path can be mapped back to a path in the road network graph, which is then returned as the result.

\textbf{Case 2:} At least one of $v_s$ and $v_d$  is not in a region.
%This includes the cases when $v_s$ is not in a region, vertex $v_d$ is not in a region, and both $v_s$ and $v_d$ are not in regions.
In this case, we find appropriate region vertices for $v_s$ or/and $v_d$. Then, we apply the procedure from case 1.

\begin{figure}[h]
\centering
  \includegraphics[width=0.9\columnwidth]{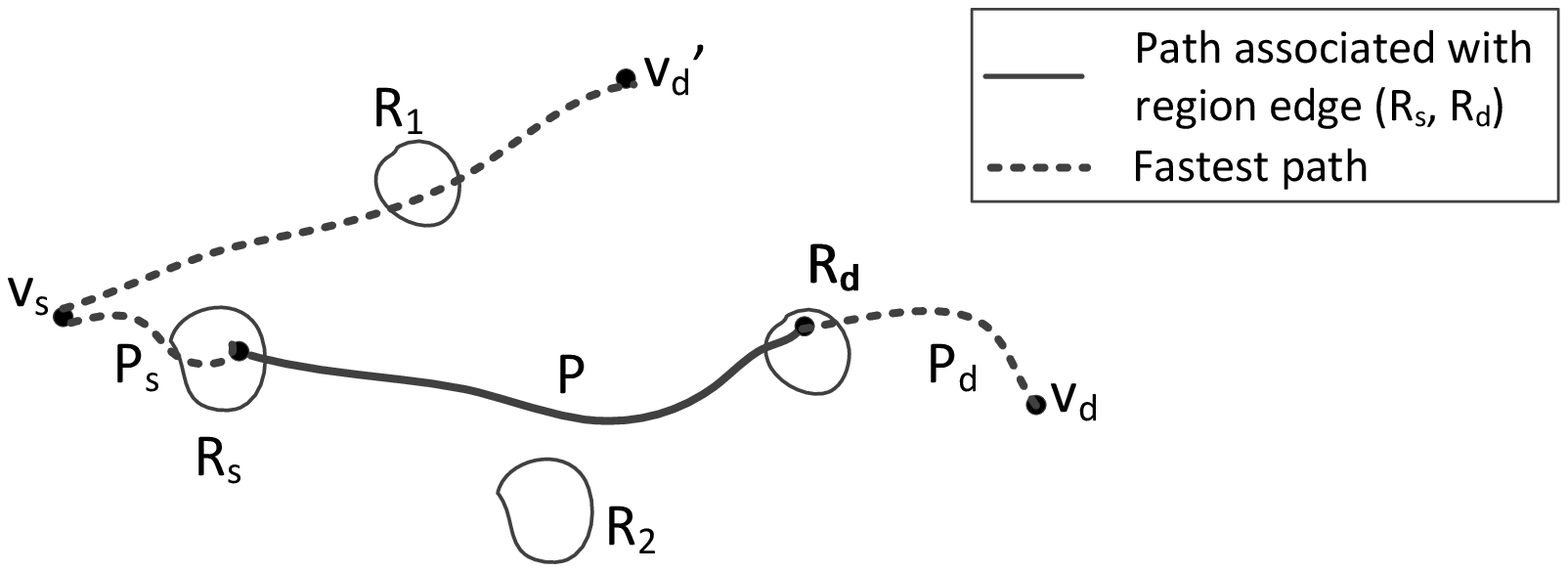}
\caption{Routing, Case 2}
\label{fig:routing}
%%\vspace{-10pt}
\end{figure}

To this end, we issue a fastest path finding algorithm from $v_s$ to $v_d$ based on road network graph $\mathcal{G}$. If a region is visited by the algorithm, we consider it as a candidate region $R_s$. Similarly, we can identify a candidate region $R_d$.
%
%The candidate region that is closest to $v_s$ (or $v_d$) is considered as the corresponding region $R_s$ (or $R_d$).
Then we apply the  procedure from case 1 with source region $R_s$ and destination region $R_d$ to identify a path $P$.
%
%Finally, we return a path that consists of three sub-paths---the sub-path of the fastest path from $v_s$ to $R_s$, the path $P$ that connects $R_s$ and $R_d$, and the sub-path of the fastest path from $R_d$ to $v_d$.
%
Finally, we return a path that consists of three sub-paths---the fastest path from $v_s$ to $R_s$, denoted as $P_s$, the path $P$ that connects $R_s$ and $R_d$, and the fastest path from $R_d$ to $v_d$, denoted as $P_d$, as shown in Figure~\ref{fig:routing}.
In case there is only one or no candidate region, we simply return the fastest path, e.g., in the case of $v_s$ and $v_d^\prime$ in Figure~\ref{fig:routing}.
%\vspace{-10pt}

\section{Empirical Study}

We conduct a comprehensive empirical study on two substantial GPS trajectory data sets. % to elicit design properties of the proposed methods, including the effectiveness and efficiency of the \emph{L2R} routing algorithm.

%%\vspace{-8pt}
\subsection{Experimental Setup}
\label{ssec:setup}

\noindent
\textbf{Road Network and GPS Trajectories:}
We use two road networks, $N_1$ and $N_2$, both obtained from OpenStreetMap (\url{openstreetmap.org}).
$N_1$ represents the road network of Denmark and includes 667,950 vertices and 1,636,040 edges, which are contained in a 320 km $\times$ 370 km rectangular region.
$N_2$ represents the road network of Chengdu, China and includes 27,671 vertices and 77,444 edges, which are contained in a 33 km $\times$ 25 km rectangular region.

We use two GPS data sets $D_1$ and $D_2$ from $N_1$ and $N_2$, respectively.
$D_1$ consists of more than 180 million high-frequency GPS records that were
collected by 183 vehicles at 1~Hz (i.e., one GPS record per second)
%during week days
in 2007 and 2008.
$D_2$ consists of 100 million low-frequency GPS records that were collected by 10,864 taxis from August 3rd to 30th 2014. The sampling rate varies from 0.03 Hz to 0.1 Hz. We only use parts of trajectories where taxi have passengers on.
%
%$D_2$ consists of 50 billion GPS records that occurred from September 2012 to November 2012 by [to add] drivers (\url{}). The sampling rate varies but is at least 0.2 Hz (i.e, at least one GPS record every 5 seconds).
%
We map match~\cite{DBLP:conf/gis/NewsonK09} the GPS records in $D_1$ and $D_2$ onto $N_1$ and $N_2$, respectively, obtaining 466,305 trajectories and 185,284 trajectories, where the trajectories in $D_2$ represent trips with passengers.
The travel distance distributions of the trajectories are shown in Table~\ref{tbl:tra_stat}.
%

%We map match~\cite{DBLP:conf/gis/NewsonK09} the GPS records in $D_1$ and $D_2$ onto the road network $N_1$ and $N_1$, respectively. $D_1$ yields 466,305 trajectories and $D_2$ generates 28,069 trajectories where taxis have passengers on. The travel distance distribution of the trajectories is shown in Table~\ref{tbl:tra_stat}.

%%\vspace{-0.2cm}
\begin{table}[ht!]
\centering
\scriptsize
\begin{tabular}{|c|c|c|c|c|} \hline
Distance ($\mathit{km}$) & (0,10] & (10,50] & (50,100] & (100, 500] \\ \hline
\# Trajectories of $D_1$ & 427,430 & 35,271 & 2,263 & 1,341 \\ \hline
Percentage (\%) & 91.6 & 7.6 & 0.5 & 0.3 \\ \hline\hline
%Distance ($\mathit{km}$) & (0,2] & (2,5] & (5,10] & (10, 35] \\ \hline
%\# Trajectories of $D_2$ & 10,959 & 14,927 & 2,117 & 66 \\ \hline
Distance ($\mathit{km}$) & (0,2] & (2,5] & (5,10] & (10, 35] \\ \hline
\# Trajectories of $D_2$ & 29,256 & 105,503 & 43,473 & 7,052 \\ \hline
Percentage (\%) & 15.8 & 56.9 & 23.5 & 3.8 \\ \hline
\end{tabular}
\caption{Statistics of Trajectories}\label{tbl:tra_stat}
%%\vspace{-10pt}
\end{table}
%%\vspace{-0.4cm}

\noindent
\textbf{Training and Testing Data:}
%
%We partition the trajectories into training data and testing data in $D_1$ and $D_2$:
%
%
%For $D_2$, trajectories that occurred during the first 21 days are used as training data and that occurred in the last 7 days are used as testing data.
%
We partition the trajectories in $D_1$ and $D_2$ into training data and testing data.
Specifically, %Different training data sets are used.
trajectories that occurred during the first 18 months in $D_1$ and the first 21 days in $D_2$ are used as training data; and trajectories that occurred in the last 6 months in $D_1$ and the last 7 days in $D_2$ are used as testing data.

\noindent
\textbf{Evaluation Criteria:}
For each trajectory in the testing data {\it Test}, we record its source and destination, departure time, and the actual path used by the trajectory.
Since the aim of the paper is to reuse local drivers' routing intelligence to recommend paths, the paths used by the local drivers are considered as the ground truth (GT) paths.
%

% Bin: emphasize that departure time is used in fastest path and google path.
In the experiments, we run learn-to-route (\textit{L2R}) on each pair of source and destination in {\it Test},
and we compare the returned path with the GT path, using the path similarity function in Section~\ref{ssec:learnDP}.
In addition, we also identify the shortest path, the fastest path, the paths returned by two personalized routing algorithms and by Google Maps, and compare them with the GT path.
The departure time is used when identifying the \textit{L2R} paths, the fastest paths and the Google Maps paths.

We also report results w.r.t. the accuracy using a different but also popular path similarity function~\cite{DBLP:journals/vldb/0002GMJ15,erkut1998modeling} to evaluate the similarity between the routing path and the ground truth path. % $\mathit{pSim}({P}_k, P_k^{V}) = \frac{\sum_{e\in P_k \cap P_k^{V}  } \mathit{length}(e)}{\sum_{e\in P_k}\mathit{length}(e)}$.
%
%%\vspace{-5pt}
\begin{equation}
\label{eq:psim2}
%\small
%\mathit{pSim}(P_a, {P}_b) = \frac{\sum_{e\in P_a \cap P_b} \mathit{len}(e)}{\sum_{e\in P_a \cup P_b}\mathit{len}(e)}
\mathit{pSim}({P}_k, P_k^{V}) = \frac{\sum_{e\in P_k \cap P_k^{V}  } \mathit{len}(e)}{\sum_{e\in P_k \cup P_k^{V}}\mathit{len}(e)}
%%\vspace{-5pt}
\end{equation}
It follows the intuition of the similarity function in Section~\ref{ssec:learnDP}, but utilizes the union of segments in the constructed path $P_k^V$ and the ground-truth path $P_k$ as the denominator.

%We report results using a different similarity function in Appendix~\ref{sec:app_exp}.
%
%+++Meanwhile, other alternative routing algorithms.
%%
%Assume that a L2R path $L=\langle e_1, e_2, \ldots, e_{m}\rangle$ and a GT path $G=\langle e_1, e_2, \ldots, e_{n}\rangle$ connect the same pair of source and destination. The accuracy of $L$ is defined as:
%%Sets $$ $x$ ($x\leqslant \mathit{min}(n, m)$) segments of $L$ appear in the GT route.
% \[
%  \mathit{Accuracy} = \frac{|L\cap G|}{|G|} \cdot 100\%
% \]
%+++why not use pSim defined in section 5.1? we must use consistent similarity equation.

Results are categorized according to the lengths of the GT paths and according to whether the source or destination of a GT path belongs to a region in the obtained region graph. If both the source and destination of a GT path are in regions, the path is in category {\it InRegion}. If either the source or the destination is in a region, the path is in category {\it InOutRegion}. If neither source nor destination belongs to a region, the path is in category {\it OutRegion}.

\noindent
\textbf{Implementation Details:}
All algorithms are implemented in Java using JDK 1.8.
We conduct experiments on a server with a 64-core AMD Opteron(tm) 2.24 GHZ CPU, 528 GB main memory under Ubuntu Linux.
%A computer with Windows 7 Enterprise, Intel (R) Core (TM) i7-2620M CPU @ 2.70GHz, 8G RAM, is used for the experiments.
%
We use distance, travel time, and fuel consumption as the travel cost features, where the fuel consumption is computed based on speed limits using vehicular environmental impact models~\cite{DBLP:journals/geoinformatica/Guo0AJT15}.
We use six commonly used road types from OpenStreetMap as road condition features: motorway, trunk, primary, secondary, tertiary, and residential.
%We use six commonly used road types in OpenStreetMap (i.e., motorway, trunk, primary, secondary, tertiary, and residential) as road condition features. % which considers the road types recorded in OpenStreetMap.
%
The transduction learning algorithm for transferring preferences (cf.\ Section~\ref{ssec:transferDP}) is implemented using the Junto library (\url{github.com/parthatalukdar/junto}). %\footnote{https://github.com/parthatalukdar/junto}.
%%\vspace{-6pt}
%\subsection{Feature Studies}
\subsection{Evaluation of Design Choices}
\label{ssec:para}

We evaluate the design choices chosen for \emph{L2R}. In particular, we show the effect of important parameters by varying them according to Table~\ref{tbl:para} where default values are shown as bold.
%
%show features of learning and transferring routing preferences and features of the \textit{L2R} solution via a few parameters.
%
%It is unnecessary to tune \textit{L2R} every time before run it. \textit{L2R} with the default values, shown as bold in Table~\ref{tbl:para}, can be utilized directly.
%
When we vary one parameter, we keep the remaining parameters at their default values. %experiments on one of the parameters, we use default values for other parameters.
We show results for both $D_1$ and $D_2$ in most empirical studies, but omit some results for $D_2$ when they show little difference to those of $D_1$.
%%\vspace{-0.2cm}
\begin{table}[h]
\centering
\scriptsize
\begin{tabular}{|c|c|} \hline
Parameters & Values \\ \hline
%Cluster Range ($\mathit{km}$) & {\bf 2}, 3, 5, 10 \\ \hline
\# T-edges  & 1X, 2X, 3X, 4X, {\bf 5X} \\ \hline
Threshold $\mathit{amr}$  & 0.5, 0.6, {\bf 0.7}, 0.8, 0.9 \\ \hline
%Training Data Size ($D_1$) & 1M, 6M, 12M, {\bf 18M} \\ \hline
%Training Data Size (\emph{Driver}) & 2\%, 25\%, 50\%, {\bf 75\% } \\ \hline
\end{tabular}
\caption{Parameters of L2R}\label{tbl:para}
%%\vspace{-10pt}
\end{table}

\noindent
{\bf Region Sizes:}
We report the sizes of the obtained regions by computing their convex hulls and then reporting their areas (in km$^2$) and maximum diameters (in km).
%
%%\vspace{-0.15cm}
\begin{table}[hp!]
\centering
\scriptsize
\begin{tabular}{|c|c|c|c|c|} \hline
% 4270
Size (km$^2$) & (0,2] & (2,10] & (10,100] & $>$100 \\ \hline
$D_1$ & 3,357 (78.6\%) & 539 (12.6\%)  &304 (7.12\%)  & 70 (1.63\%)  \\
      &  / 9.5 & / 15.8 & / 29.9 & / 304.1 \\ \hline \hline
% 538
Size (km$^2$) & (0,2] & (2,5] & (5,10] & $>$10 \\ \hline
$D_2$ & 388 (72.1\%)  & 127 (23.6\%)  & 19 (3.53\%) & 4 (0.74\%) \\
      & / 4.24 & / 8.17& / 8.59 & / 6.22 \\ \hline
\end{tabular}
\caption{Region Sizes}\label{tbl:cluster_stat}
%%\vspace{-10pt}
\end{table}
%%\vspace{-0.25cm}
%
Table~\ref{tbl:cluster_stat} reports the numbers of regions whose area falls in given ranges and the maximum diameter of the regions in each range. There are a few large regions, but most regions have sizes less than 2 km$^2$. This indicates that the proposed modularity-based clustering is able to control the region size and avoids very large regions.
$D_1$ has a few large regions, which represent backbone highways. Since we maintain inner-region paths for regions, large regions do not affect the final routing quality.

\noindent
\textbf{Transferring Preferences:}
We study the accuracy of transferring preferences from T-edges to B-edges in Step 2.
As we have no ground-truth preferences for B-edges, we cannot evaluate the accuracy of the transferred preferences in a straightforward manner.
To evaluate the accuracy, we randomly partition the preferences of T-edges into 5 partitions.
We reserve one partition as a ground truth. Next, we use partition 1; partitions 1 and 2; partitions 1, 2, and 3; and partitions 1, 2, 3, and 4, to conduct the preference transfer. For each T-edge in the reserved partition, we obtain a transferred preference, which we compare it with the ground truth preference.
We report the accuracy of the transferred preference against the ground truth preference using Jaccard similarity.

\begin{figure}[hp!]
\centering%
\begin{minipage}{.5\textwidth}
	\centering
	\subfigure[Varying \# T-Edges]{
		\label{fig:exp_lp_seedNum}
		\begin{minipage}[b]{0.46\textwidth}
			\includegraphics[width=1\textwidth]{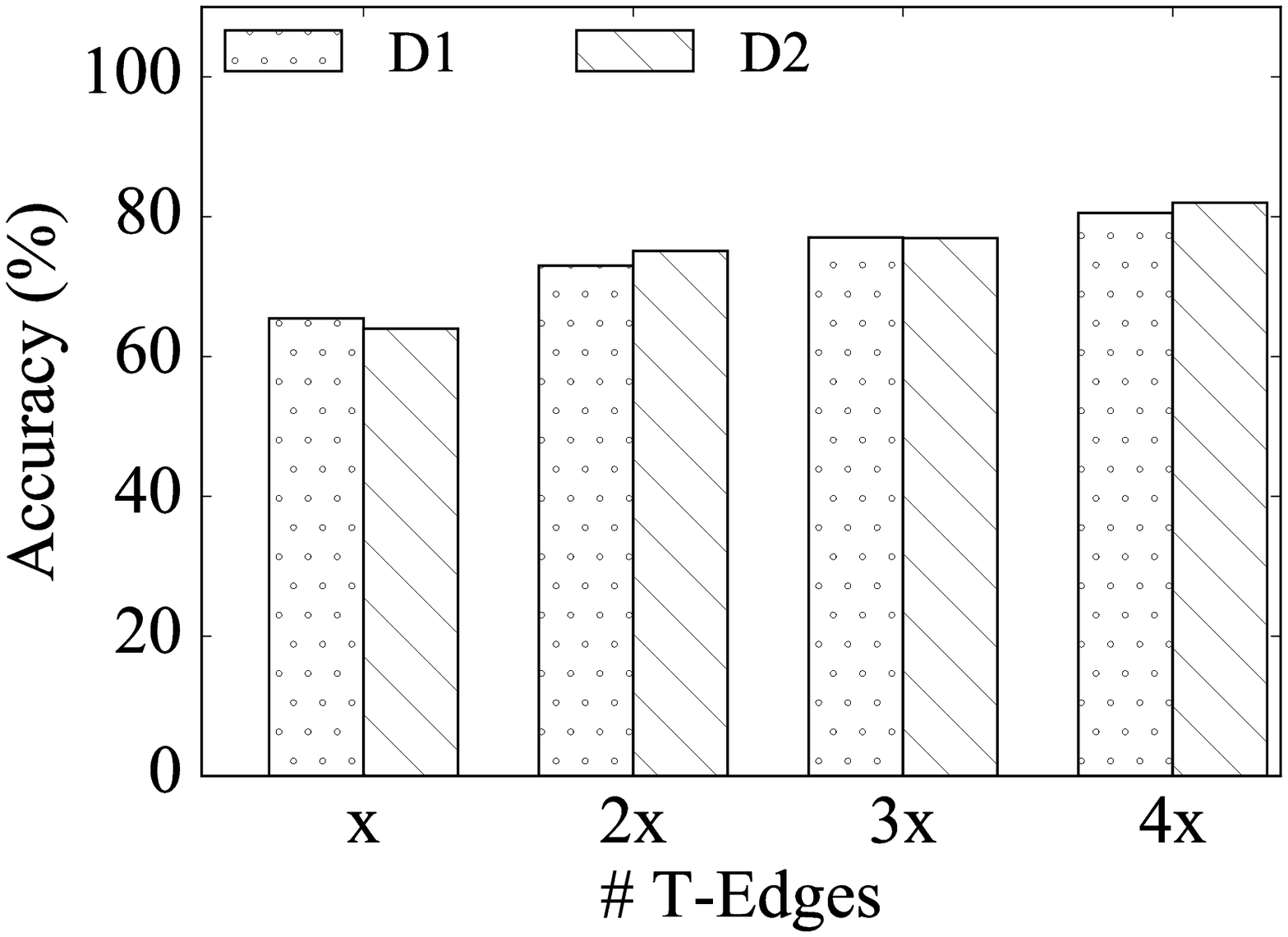}
		\end{minipage}
	}
	\subfigure[Varying $\mathit{amr}$, $D_1$]{
		\label{fig:exp_lp_sim}
		\begin{minipage}[b]{0.46\textwidth}
			\includegraphics[width=1\textwidth]{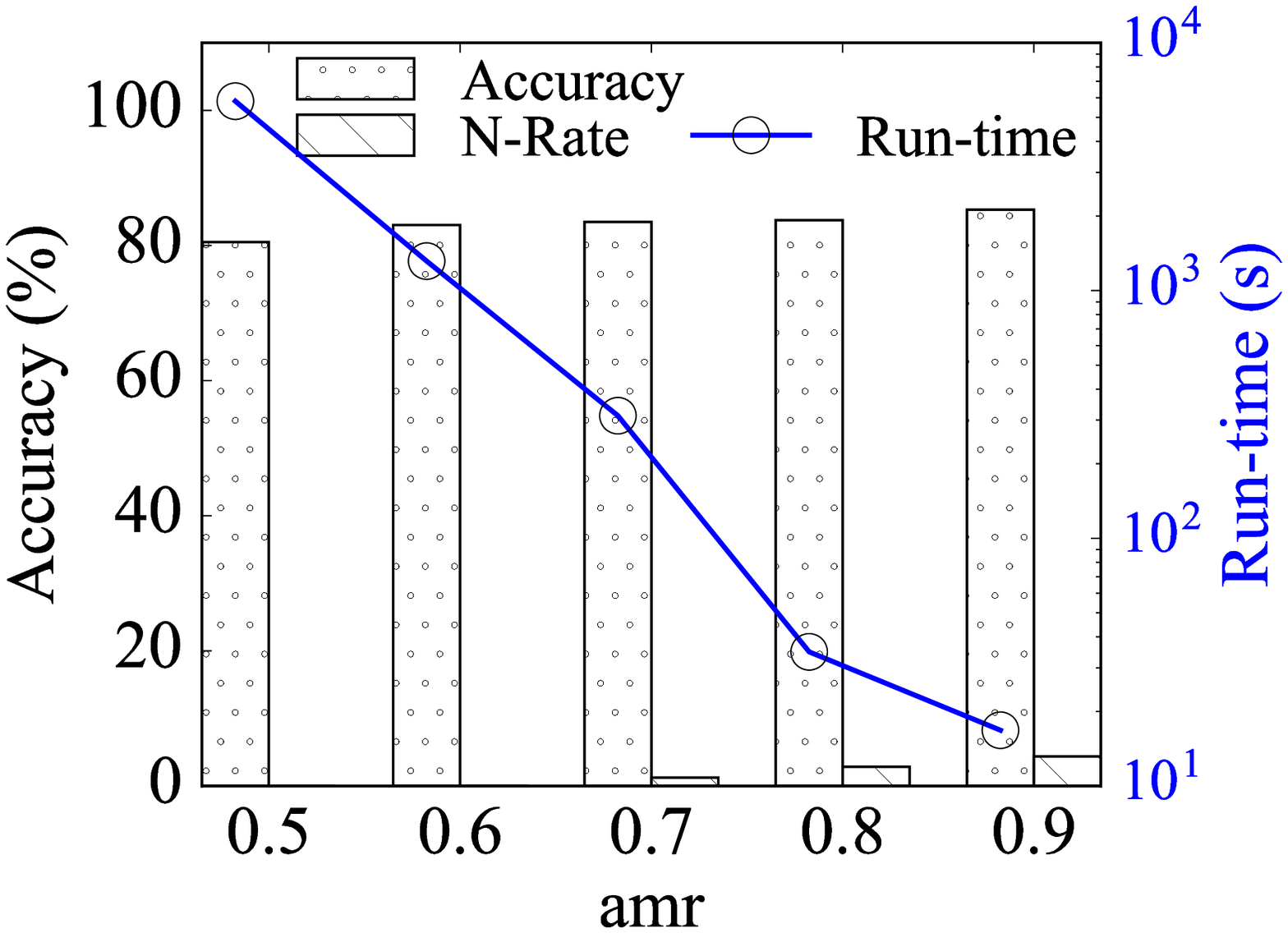}
		\end{minipage}
	}
	\caption{Parameters of Preference Transfer}
\end{minipage}%
\end{figure}

Figure~\ref{fig:exp_lp_seedNum} shows the accuracy when using 1, 2, 3, and 4 partitions, labeled as X, 2X, 3X, and 4X.
The results indicate that the more preferences of T-edges are used, the better the accuracy we get. Therefore, we use all the preferences of T-edges (i.e., 5X) in the remaining experiments.

Next, we consider the effect of the adjacency matrix reduction parameter threshold $\mathit{amr}$ on the transfer process. Since Figure~\ref{fig:rp_vs_prefsim} already suggests that when the similarity between two region edges is low, their preference vectors are dissimilar, we vary $\mathit{amr}$ from 0.5 to 0.9 and ignore small values.

We use 4 partitions of T-edge preferences to build the adjacency matrix and use the last partition as the ground truth preferences.
%The region edge similarities recorded in the adjacency matrix $\mathbf{M}$ influence both the accuracy and efficiency of the preference transfer.
We report the accuracy of the transferred preferences against the ground truth measured using Jaccard similarity, the null rate (N-rate), i.e., the percentage of transferred preferences that get null values, and the run-time in Figure~\ref{fig:exp_lp_sim}.

The accuracy of the transfer process increases slightly as $\mathit{amr}$ increases and is not sensitive to the change of $\mathit{amr}$ values when $\mathit{amr}$ exceeds 0.5.
This is intuitive because a large $\mathit{amr}$ enables transfer of routing preferences from T-edges only to highly similar B-edges.
%When a B-edge is similar to multiple T-edges with different similarities, the process naturally chooses to transfer the preference of a B-edge from the most similar B-edge.
%
However, as the $\mathit{amr}$ value increases, the graph used in the graph-based transduction learning may become disconnected. Thus, some B-edges cannot be associated with transferred preferences and thus get a null preference vector.
A smaller $\mathit{amr}$ has the effect that the graph used in the graph-based transduction obtains many edges and thus takes longer run-time.
The setting $\mathit{amr}=0.7$ gives the best trade-off, i.e., relatively high accuracy and efficiency and low null rate. We thus use this value in the remaining experiments. We simply associate fastest paths with B-edges with null preference vectors.

\begin{figure*}
	\centering
	\subfigure[By Distance, $D_1$]{
		\begin{minipage}[b]{0.22\textwidth}
			\includegraphics[width=1\textwidth]{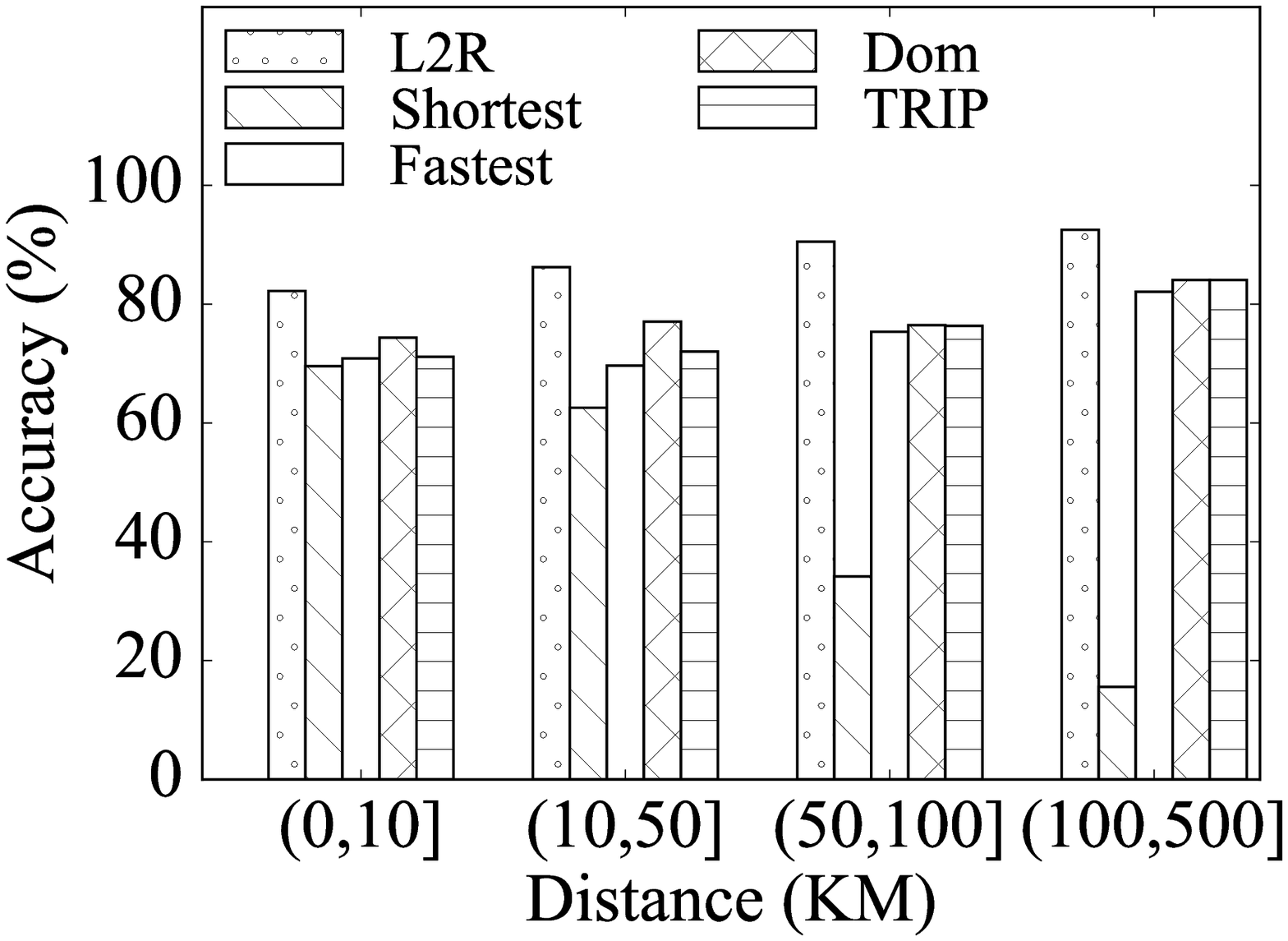}
		\end{minipage}
	}
	\subfigure[By Region, $D_1$]{
		\begin{minipage}[b]{0.22\textwidth}
			\includegraphics[width=1\textwidth]{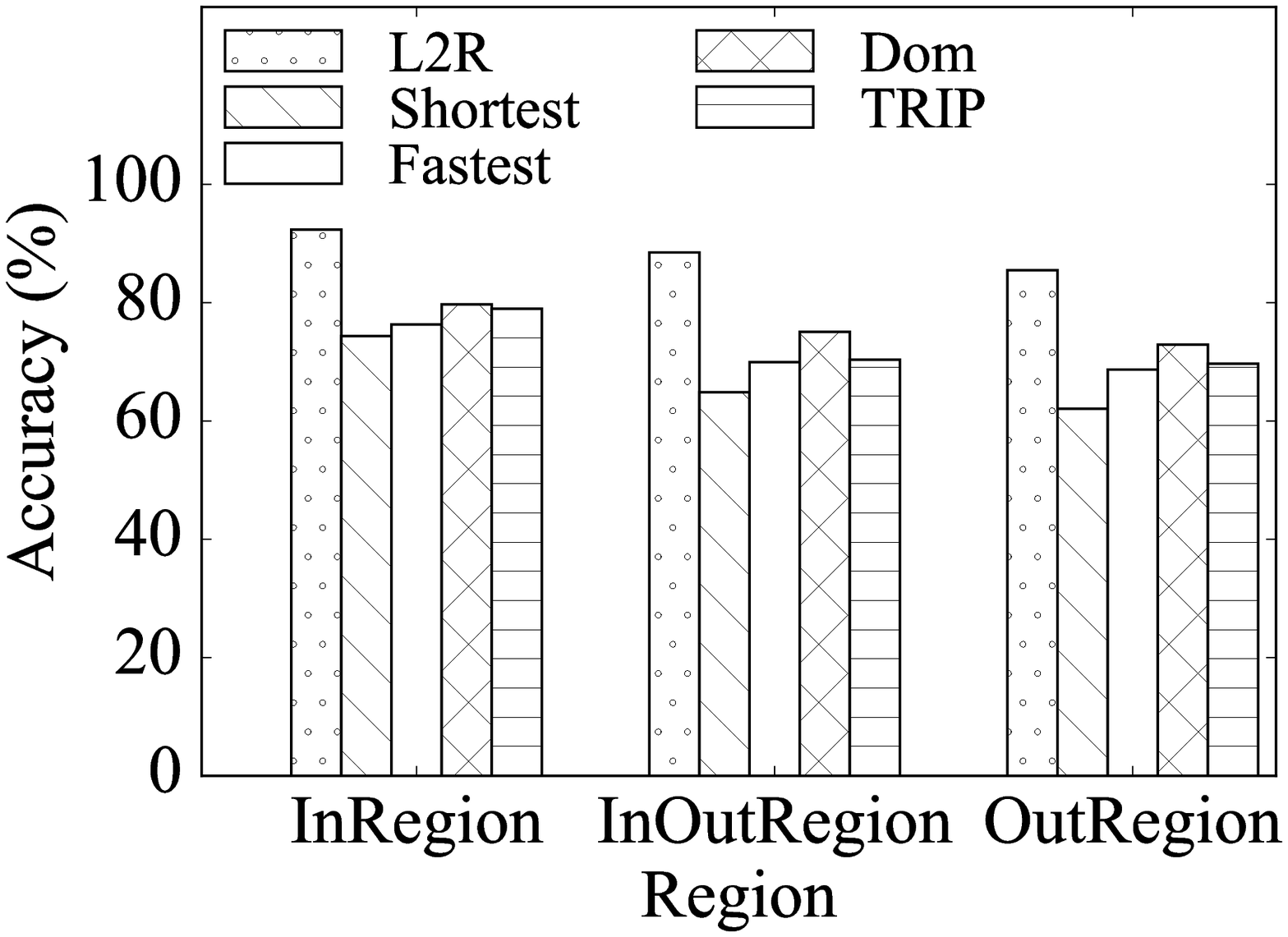}
		\end{minipage}
	}
\subfigure[By Distance, $D_2$]{
		\begin{minipage}[b]{0.22\textwidth}
			\includegraphics[width=1\textwidth]{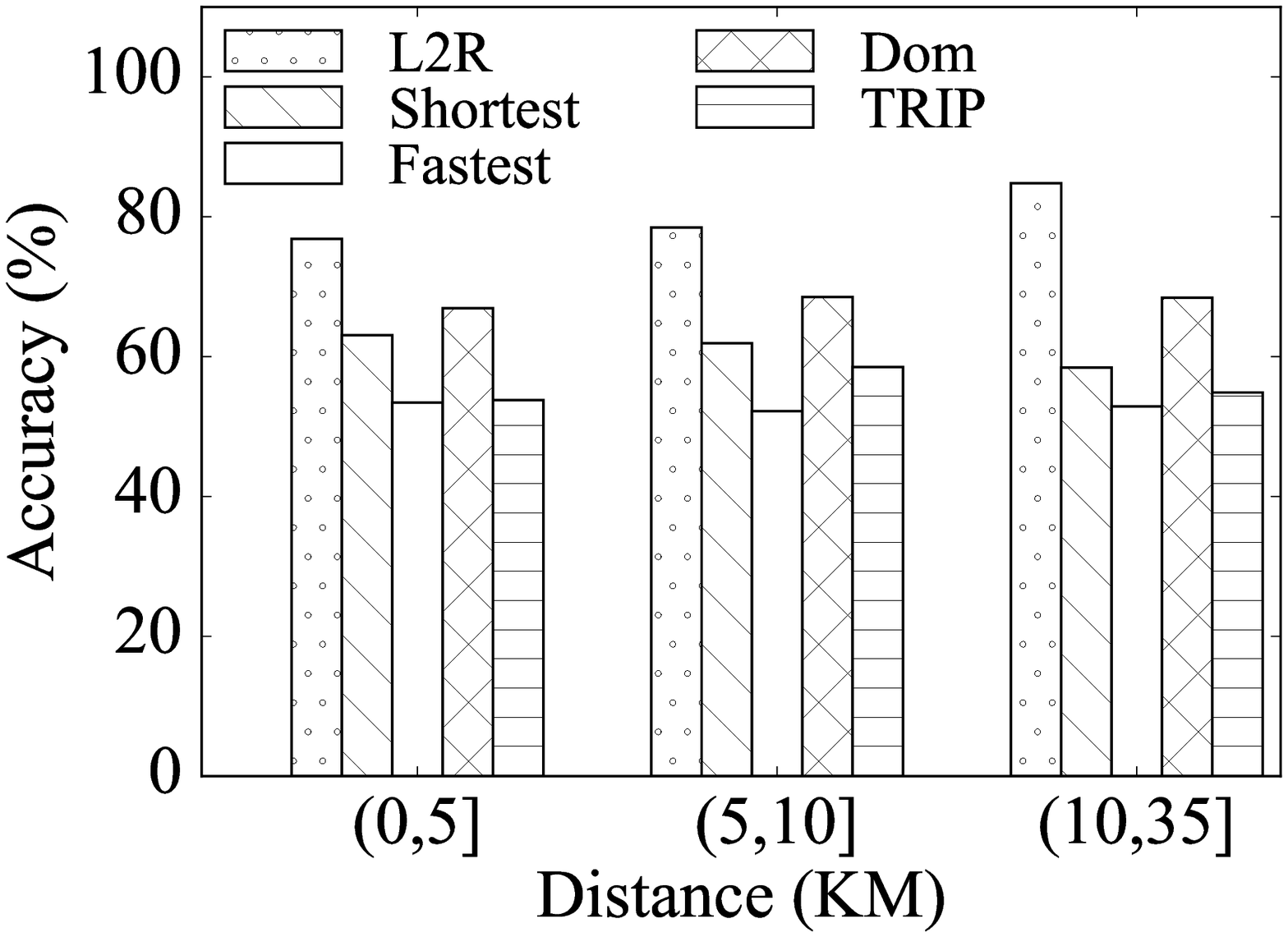}
		\end{minipage}
	}
	\subfigure[By Region, $D_2$]{
		\begin{minipage}[b]{0.22\textwidth}
			\includegraphics[width=1\textwidth]{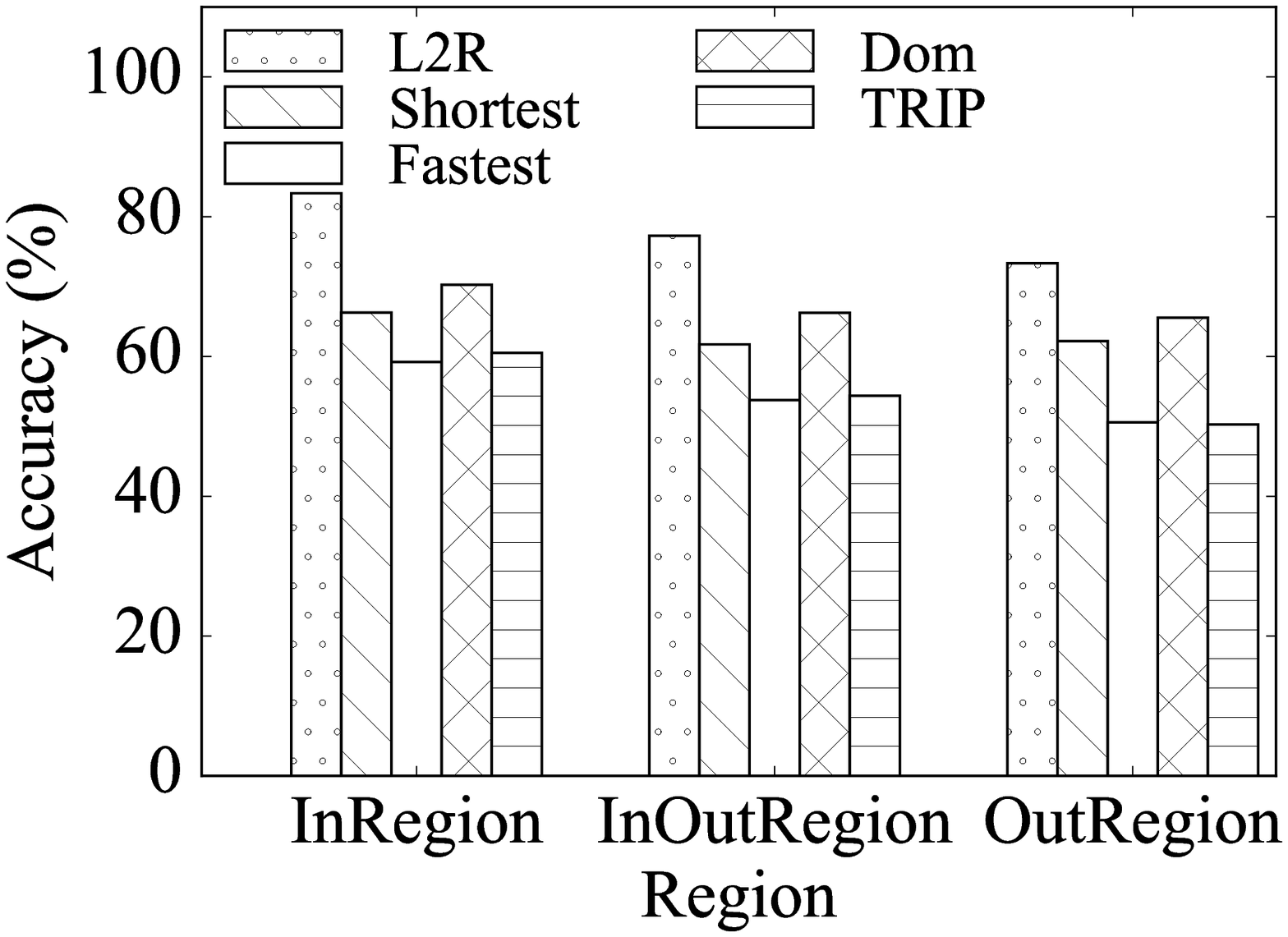}
		\end{minipage}
	}
%%\vspace{-4pt}
	\caption{Accuracy using Equation~\ref{eq:psim}}
	\label{fig:exp_bytime_accu}
%\vspace{-10pt}
\end{figure*}

\begin{figure*}[!ht]
	\centering
	\subfigure[By Distance, $D_1$]{
		\begin{minipage}[b]{0.22\textwidth}
			\includegraphics[width=1\textwidth]{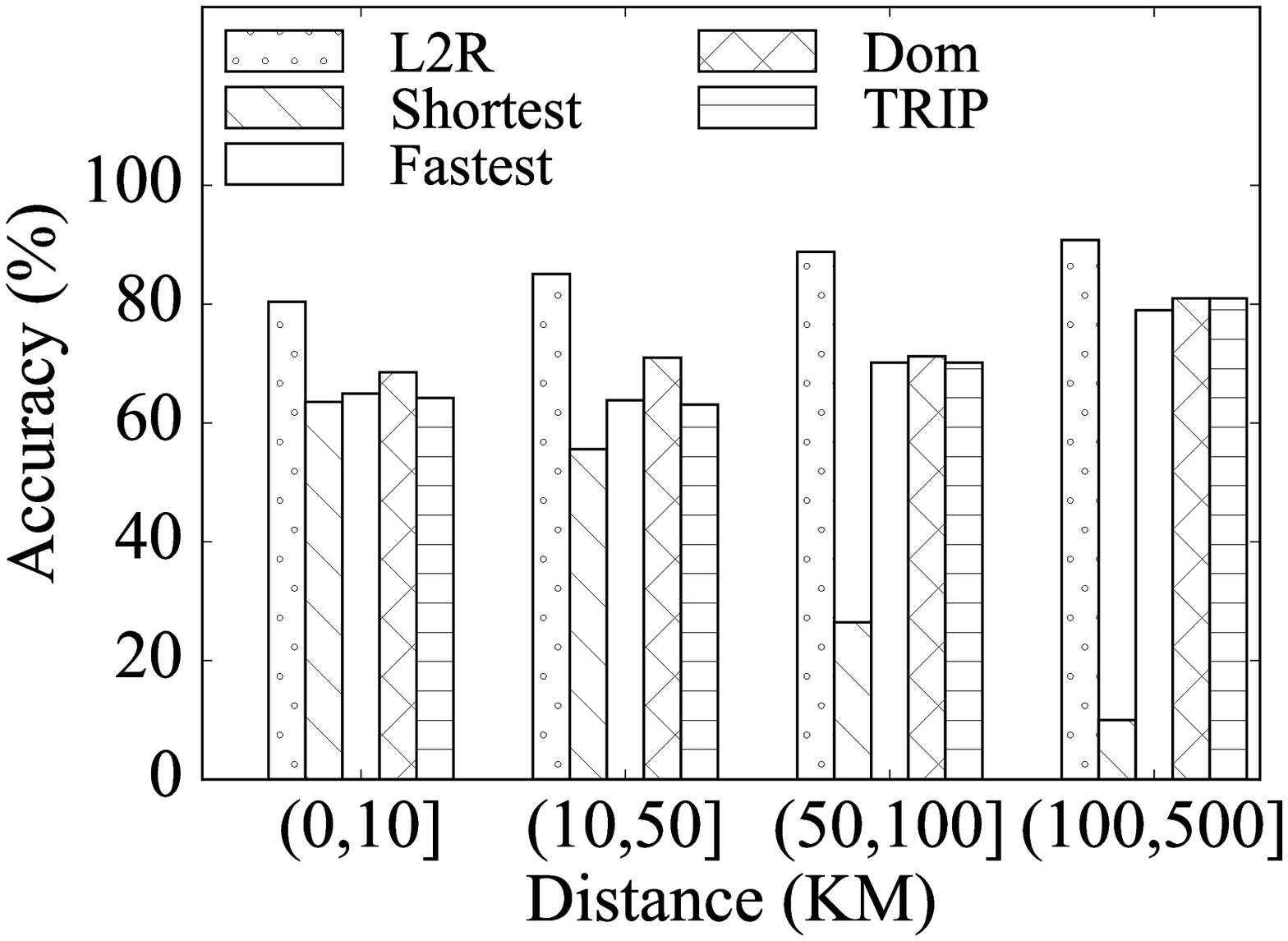}
		\end{minipage}
	}
	\subfigure[By Region, $D_1$]{
		\begin{minipage}[b]{0.22\textwidth}
			\includegraphics[width=1\textwidth]{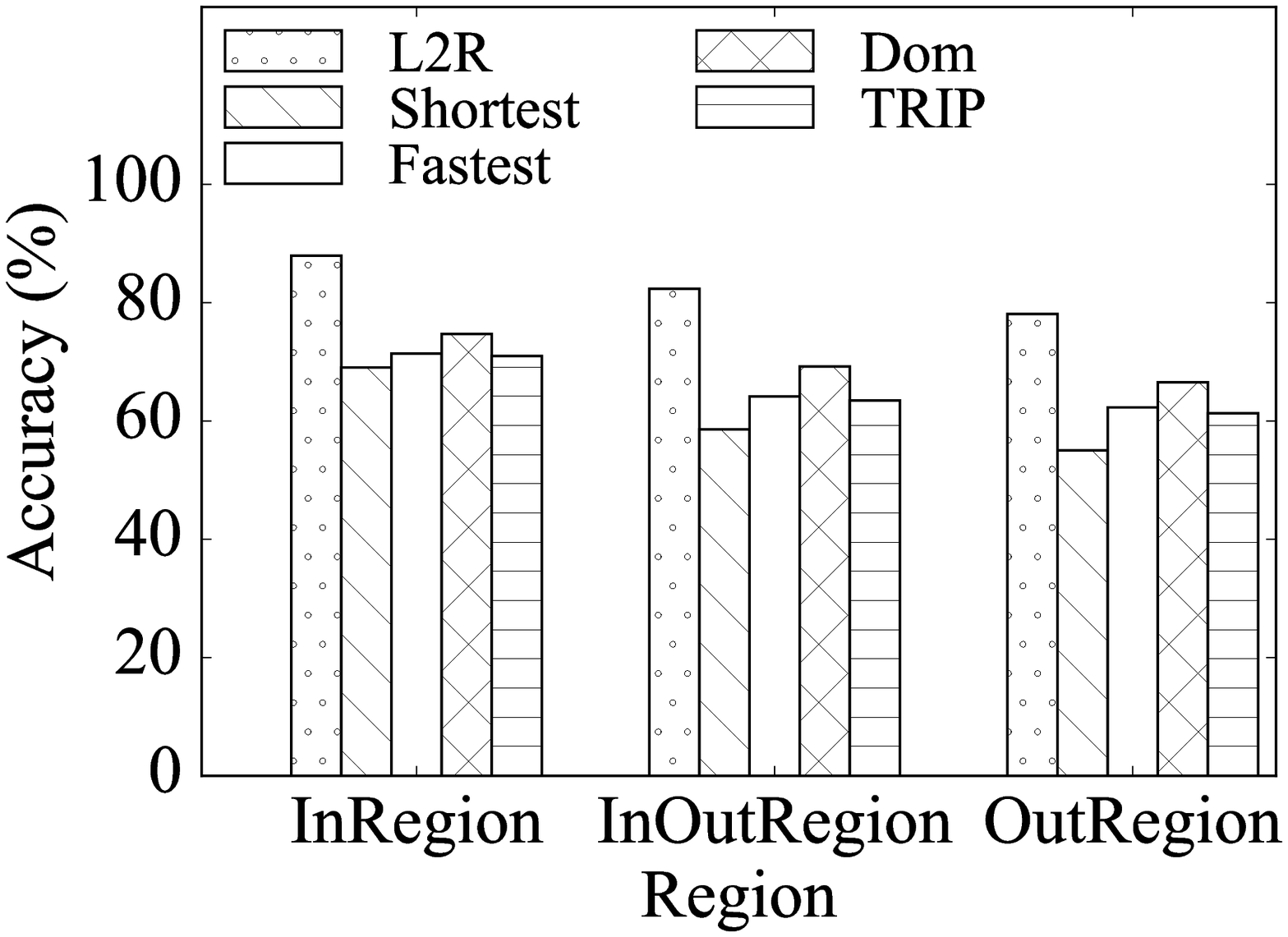}
		\end{minipage}
	}
\subfigure[By Distance, $D_2$]{
		\begin{minipage}[b]{0.22\textwidth}
			\includegraphics[width=1\textwidth]{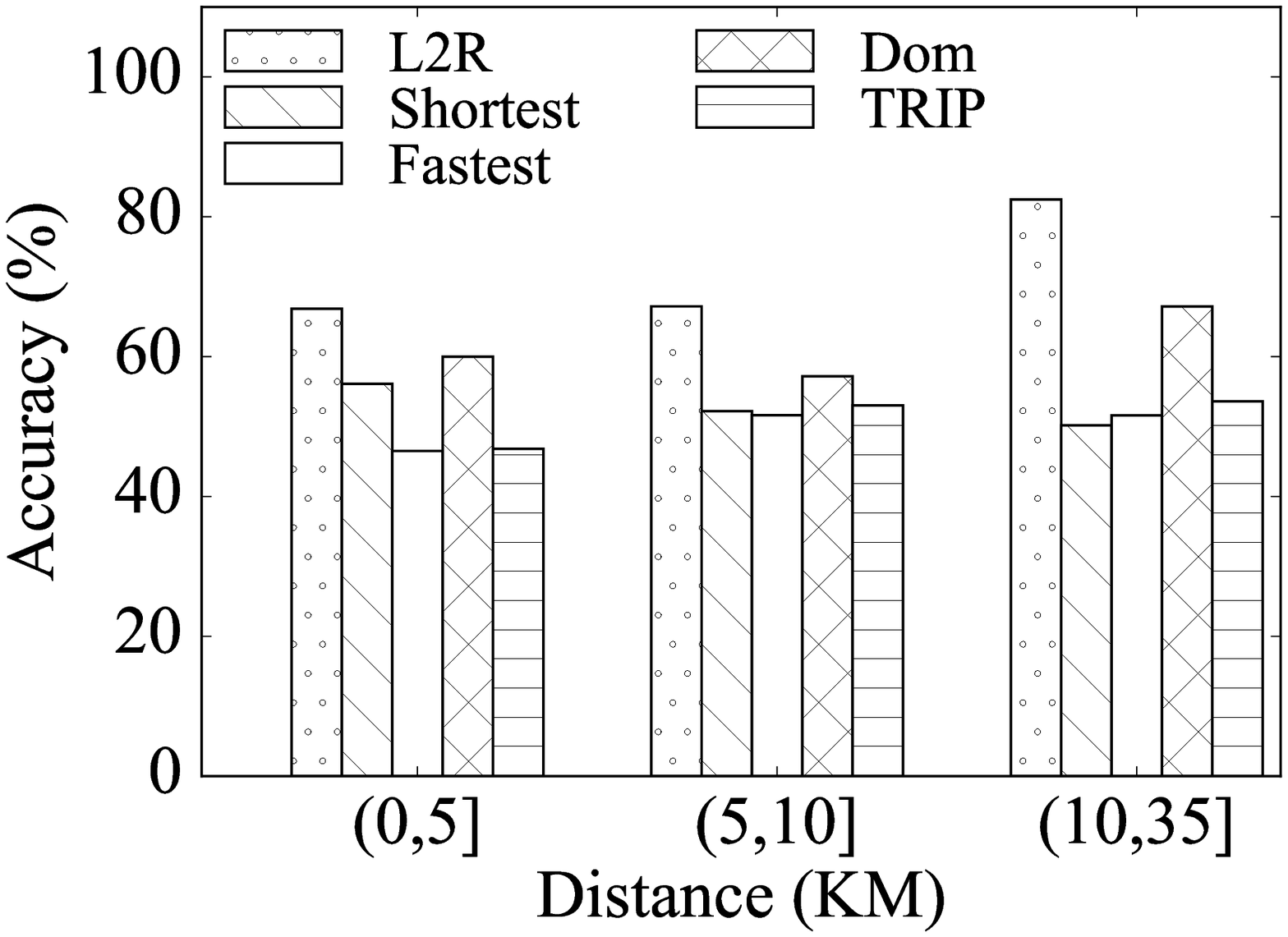}
		\end{minipage}
	}
	\subfigure[By Region, $D_2$]{
		\begin{minipage}[b]{0.22\textwidth}
			\includegraphics[width=1\textwidth]{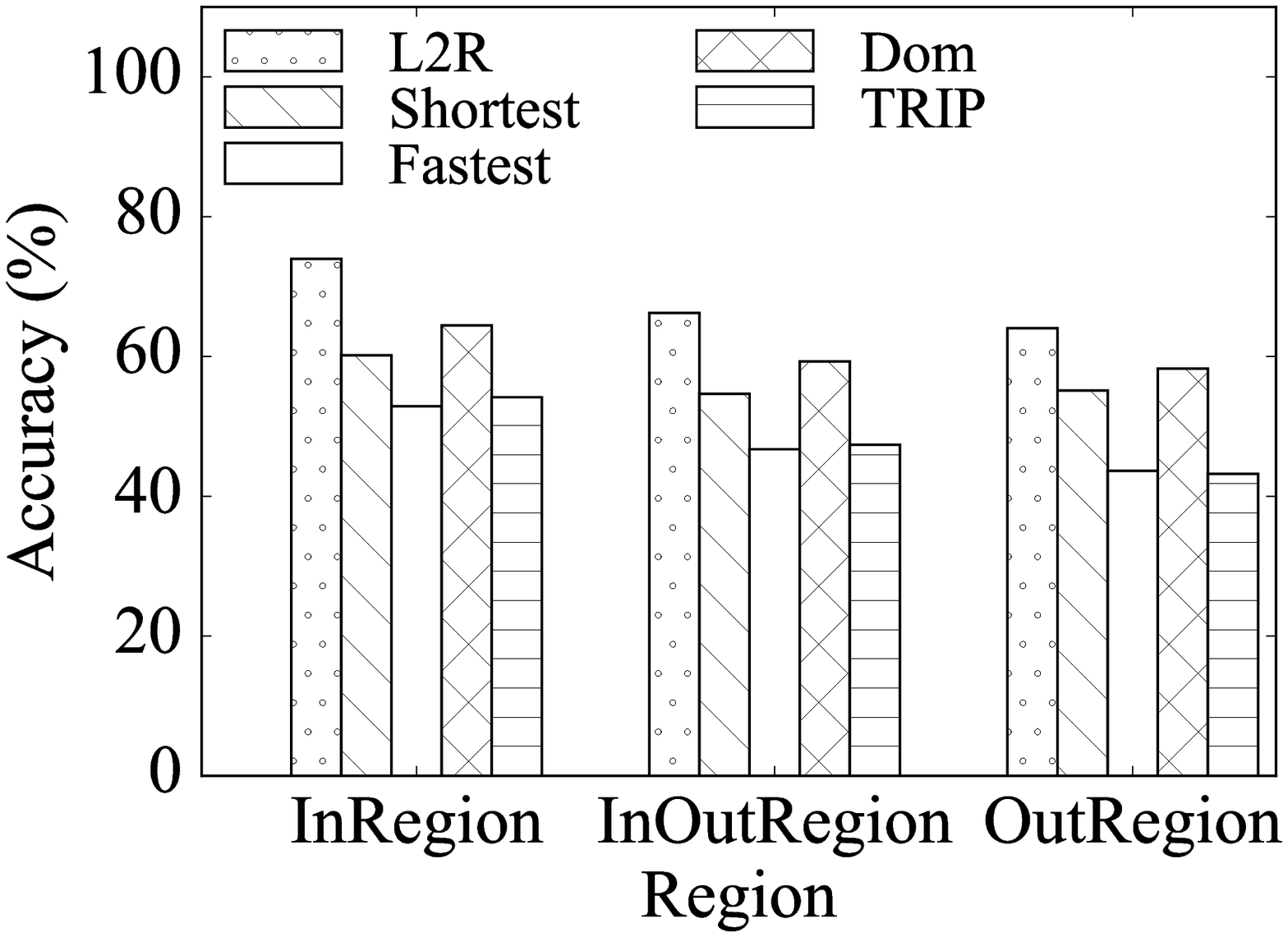}
		\end{minipage}
	}
	\caption{Accuracy using Equation~\ref{eq:psim2}}
	\label{fig:exp_bytime_accu_union}
%\vspace{-10pt}
\end{figure*}

\begin{figure*}
	\centering
	\subfigure[By Distance, $D_1$]{
		\begin{minipage}[b]{0.22\textwidth}
			\includegraphics[width=1\textwidth]{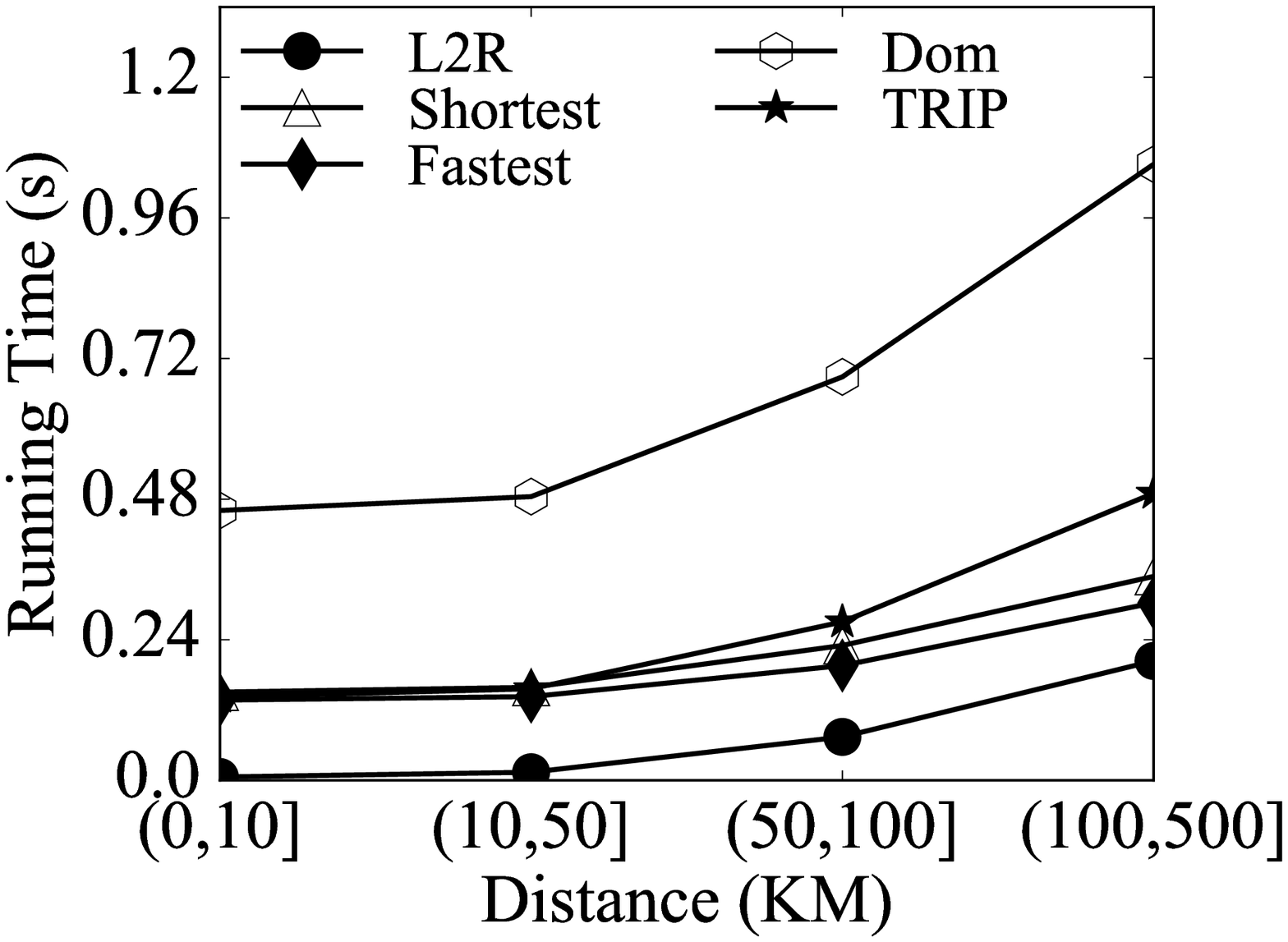}
		\end{minipage}
	}
	\subfigure[By Regions, $D_1$]{
		\begin{minipage}[b]{0.22\textwidth}
			\includegraphics[width=1\textwidth]{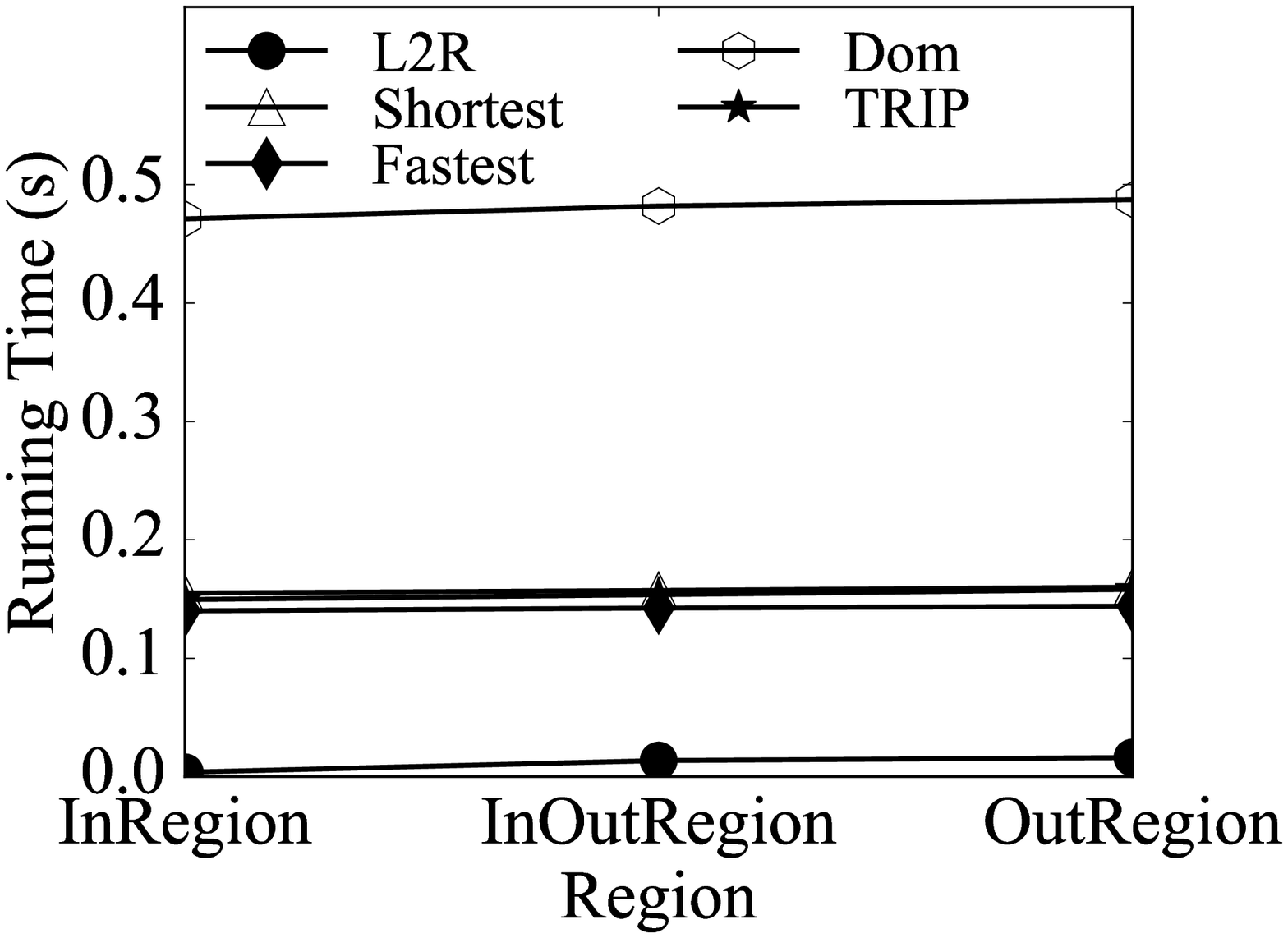}
		\end{minipage}
	}
    \subfigure[By Distance, $D_2$]{
		\begin{minipage}[b]{0.22\textwidth}
			\includegraphics[width=1\textwidth]{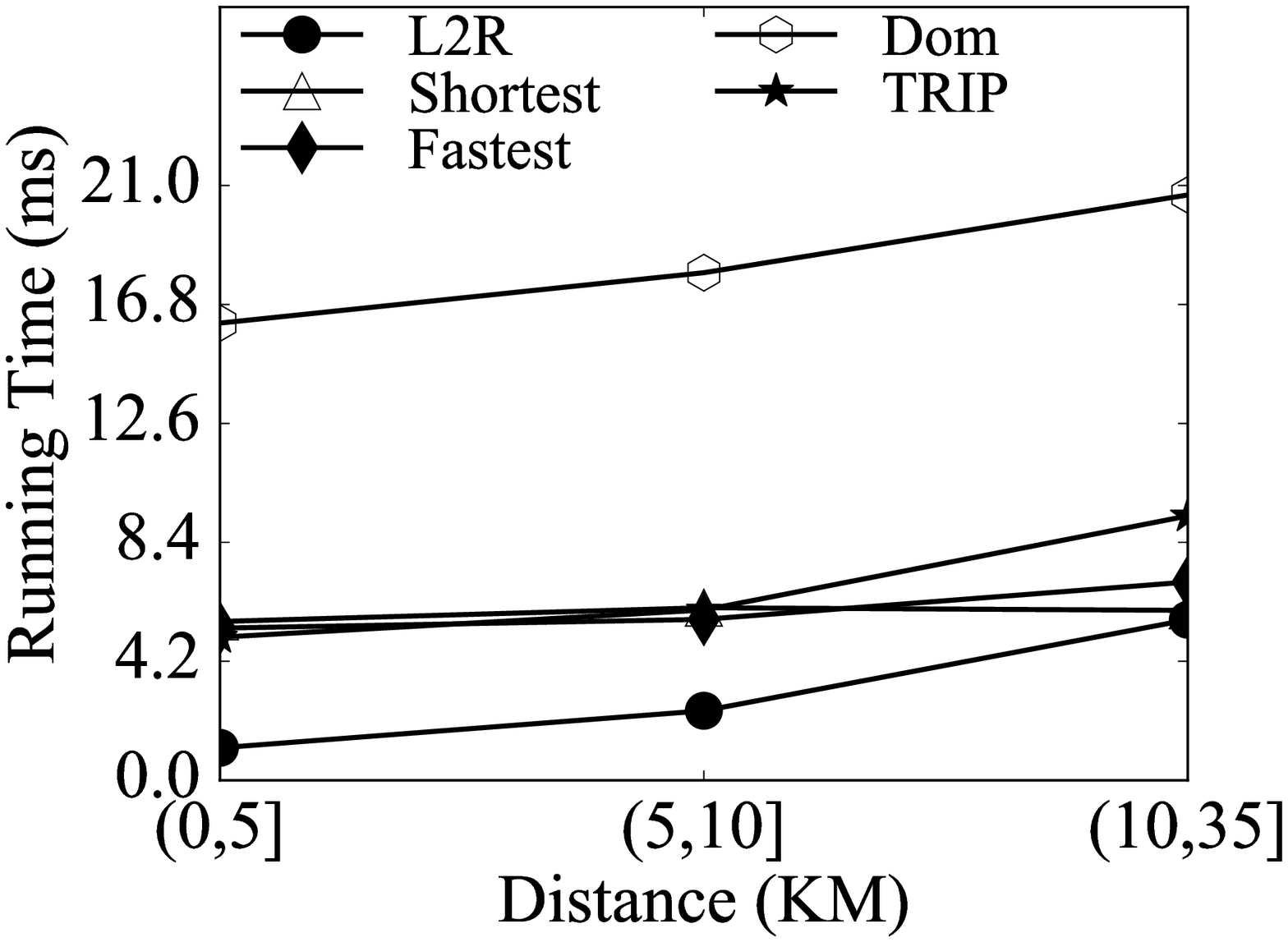}
		\end{minipage}
	}
	\subfigure[By Regions, $D_2$]{
		\begin{minipage}[b]{0.22\textwidth}
			\includegraphics[width=1\textwidth]{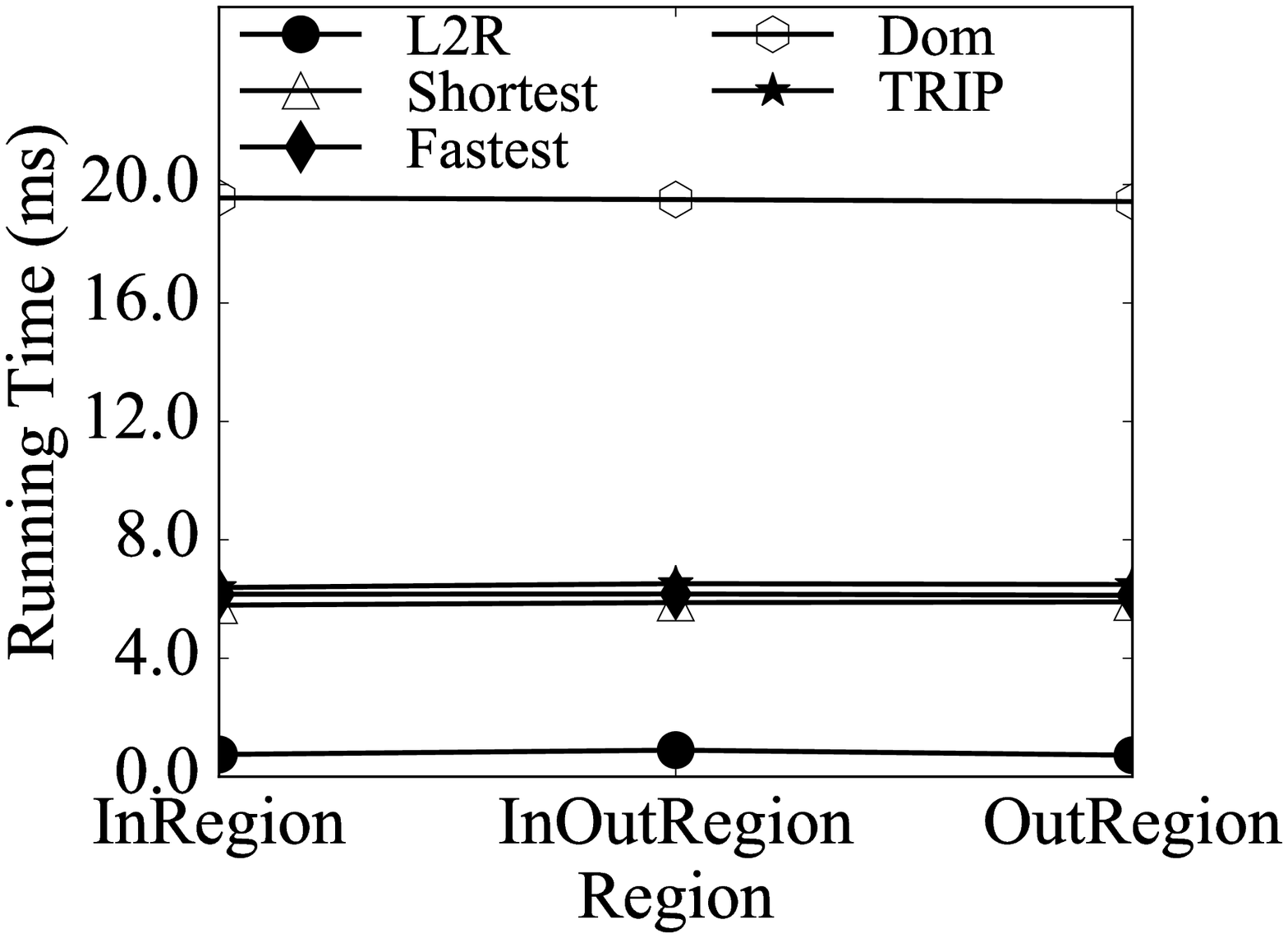}
		\end{minipage}
	}
%%\vspace{-4pt}
	\caption{Efficiency}
	\label{fig:exp_bytime_perf}
%\vspace{-10pt}
\end{figure*}

\begin{figure*}
	\centering
	\subfigure[By distance, $D_1$]{
		\begin{minipage}[b]{0.22\textwidth}
			\includegraphics[width=1\textwidth]{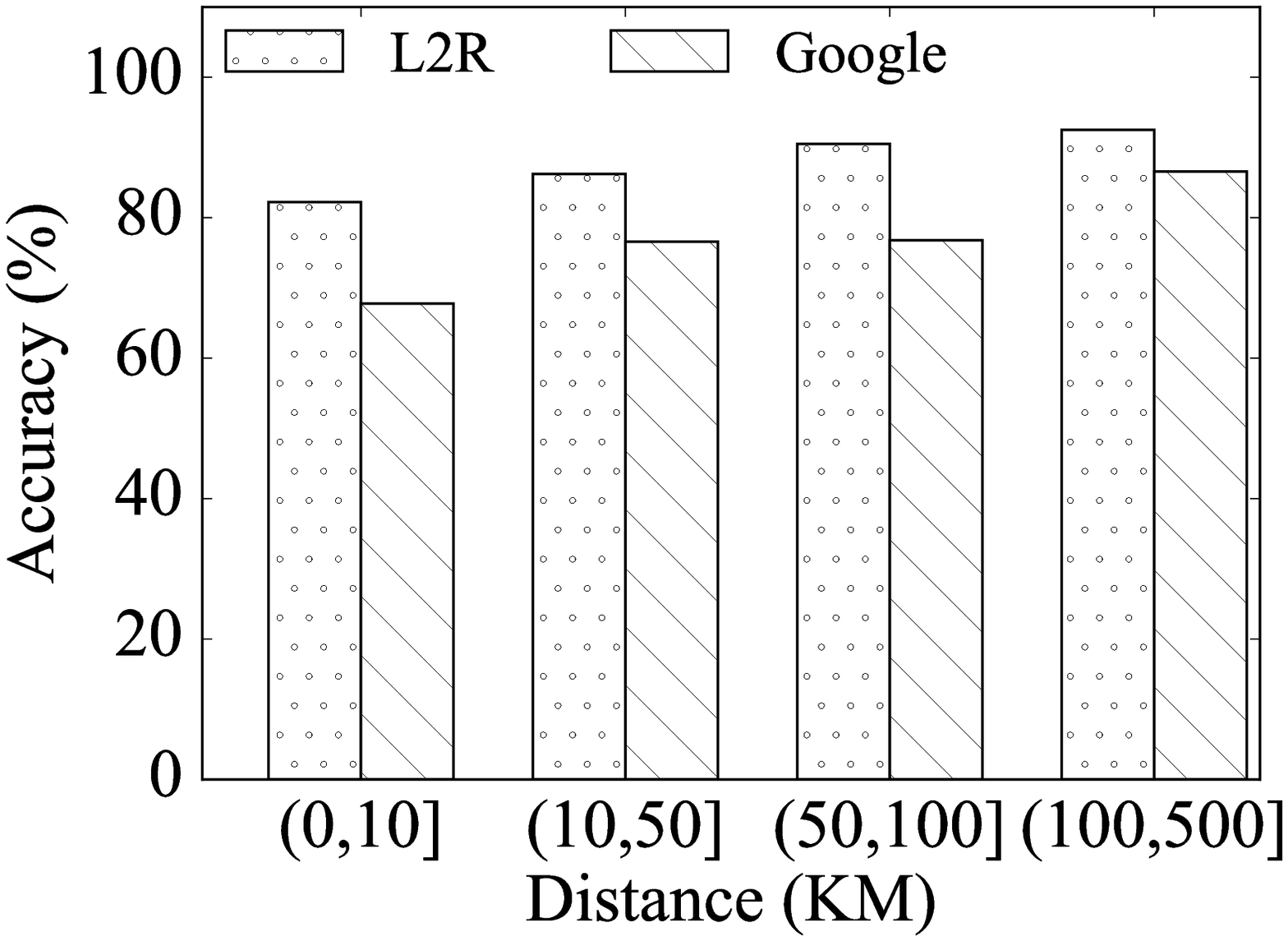}
		\end{minipage}
	}
	\subfigure[By regions, $D_1$]{
		\begin{minipage}[b]{0.22\textwidth}
			\includegraphics[width=1\textwidth]{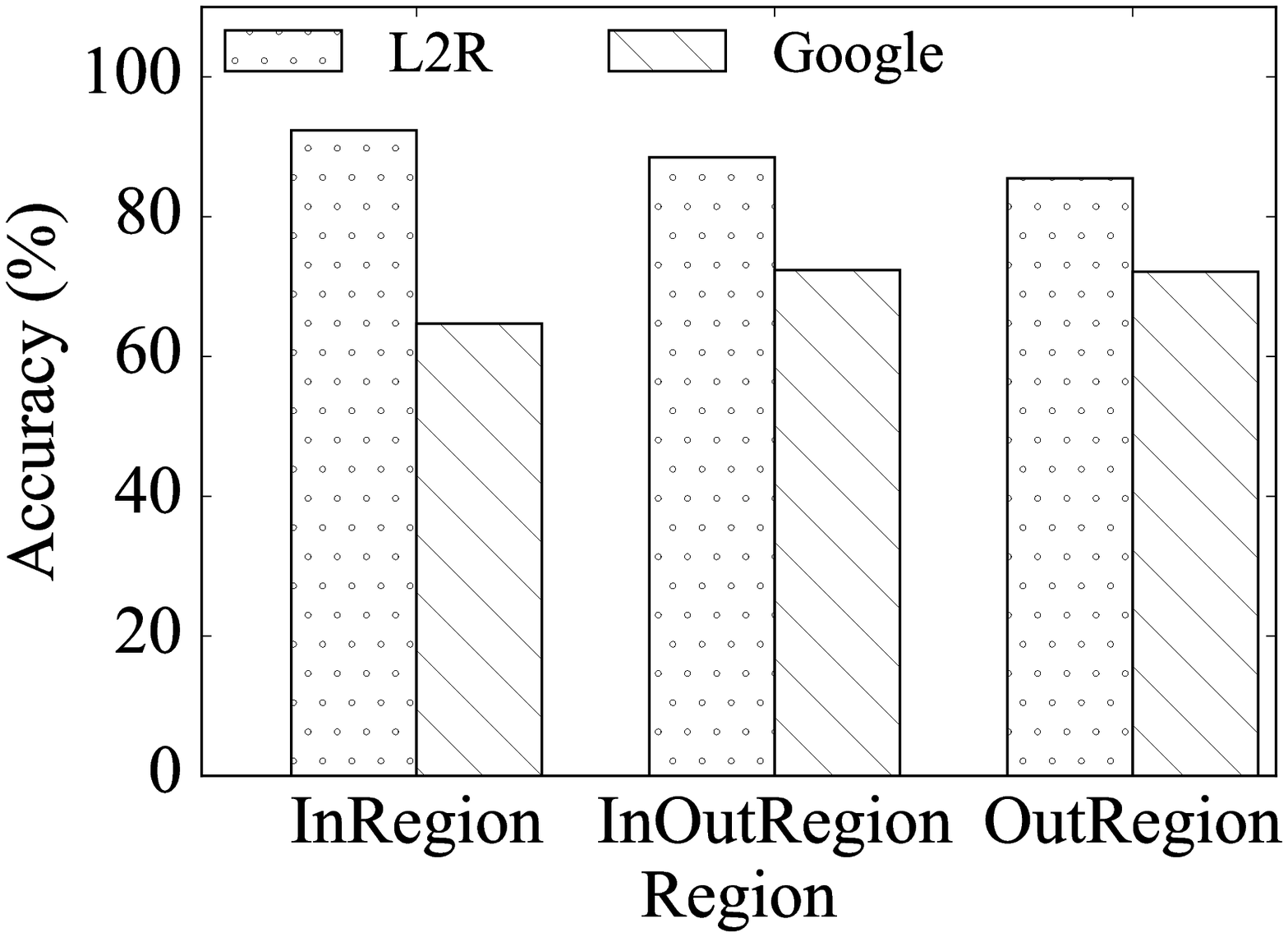}
		\end{minipage}
	}
\subfigure[By distance, $D_2$]{
		\begin{minipage}[b]{0.22\textwidth}
			\includegraphics[width=1\textwidth]{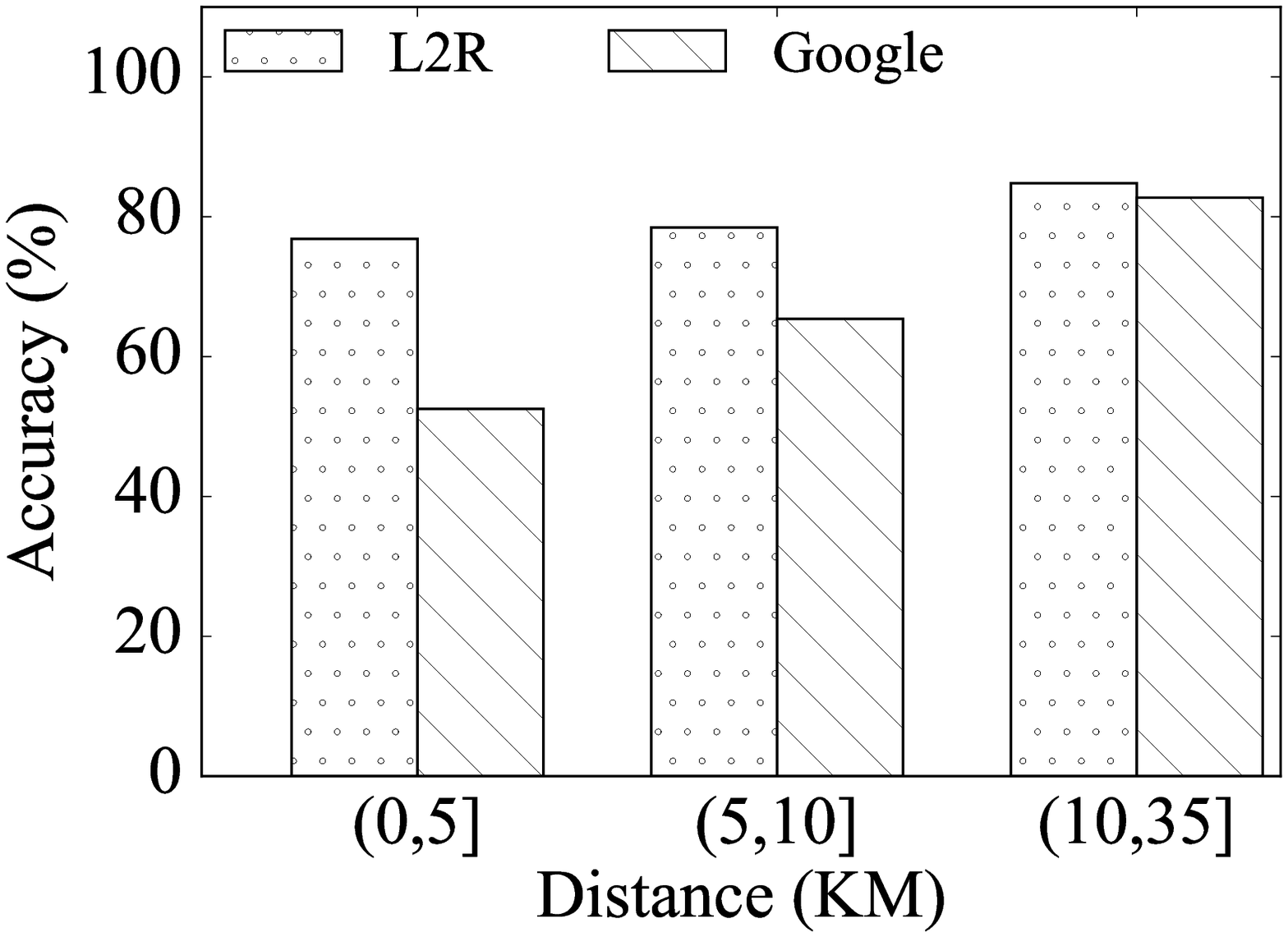}
		\end{minipage}
	}
	\subfigure[By regions, $D_2$]{
		\begin{minipage}[b]{0.22\textwidth}
			\includegraphics[width=1\textwidth]{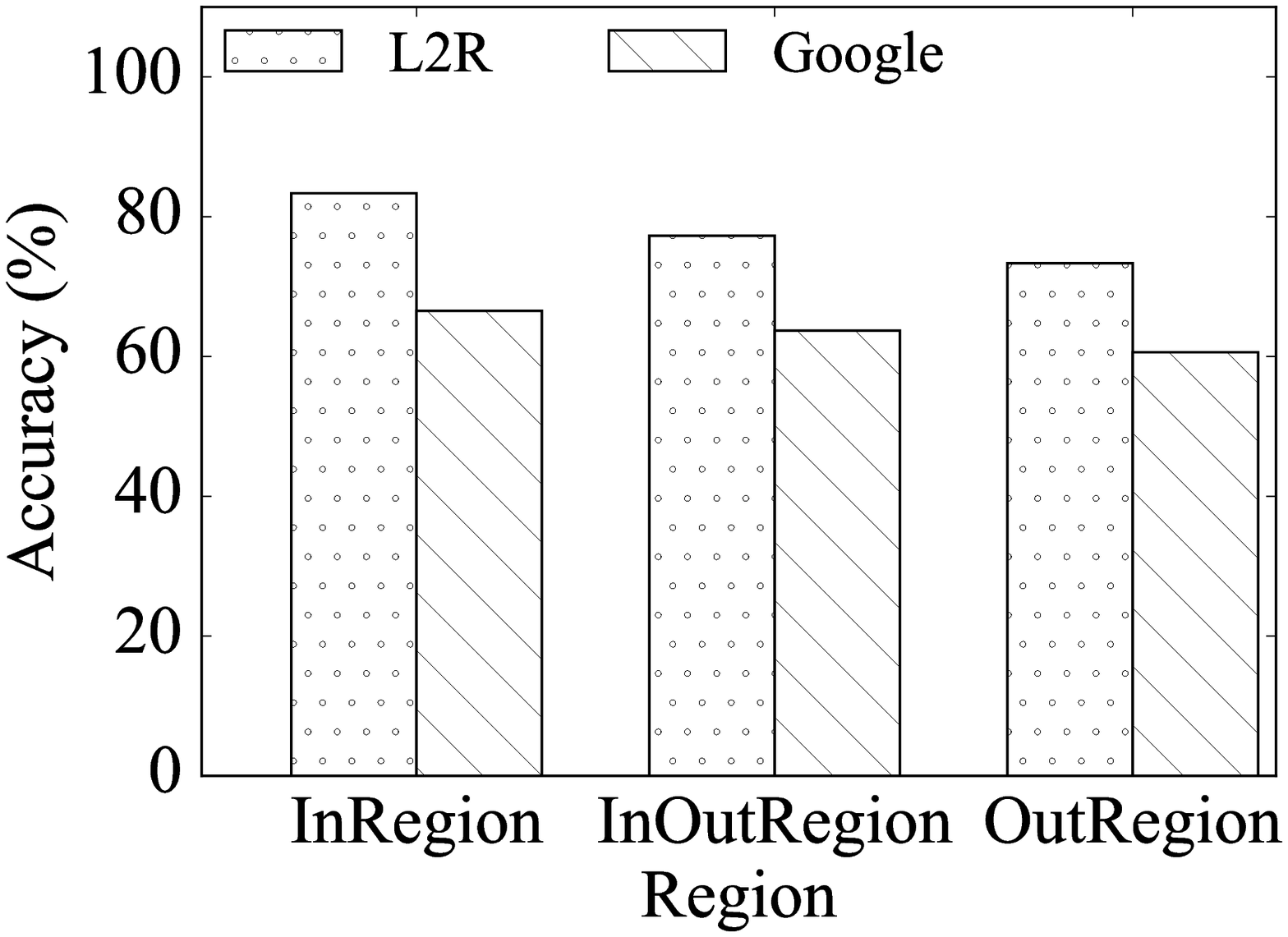}
		\end{minipage}
	}
	\caption{Comparison with Google Maps}
	\label{fig:exp_google}
%\vspace{-10pt}
\end{figure*}

%%\vspace{-0.1cm}
\subsection{Comparisons with Other Routing Algorithms}
\label{sec_exp:compare}
We proceed to compare \emph{L2R} with the shortest and fastest routing algorithms and with two personalized routing algorithms.
We apply Dijkstra's algorithm to identify the shortest (\emph{Shortest}) and fastest paths (\emph{Fastest}). %When identifying shortest paths, edge weights represents edge lengths.
%
%When identifying fastest paths, edge weights represent travel times that are derived from the trajectories that occurred on the edges, thus accounting for time-varying traffic conditions. In particular, each edge is associated with a peak travel time and an off-peak travel time.
%
We do not apply advanced speeding up techniques for routing, e.g., contraction hierarchy~\cite{DBLP:conf/wea/GeisbergerSSD08}, since they have no improvement over the accuracy  but only over the query efficiency. When applying such speed-up techniques, the efficiency of computing all paths, including the \emph{L2R} paths, can be improved consistently. We leave such performance improvements as an interesting future research direction.

We also consider two personalized routing algorithms, {\it Dom}~\cite{DBLP:journals/vldb/0002GMJ15} and {\it TRIP}~\cite{DBLP:conf/aaai/LetchnerKH06}, that are able to find personalized ``shortest'' paths between arbitrary source and destination for individual drivers.
The algorithms first learn a global routing preference (rather than a routing preference for each region pair in this paper) for each driver from the driver's historical trajectories, then use the learned preference to obtain new, personalized weights for all edges, and finally apply shortest-path finding using the new edge weights. Specifically, {\it Dom} utilizes a routing preference that considers distance, travel time, and fuel consumption, whereas {\it TRIP} uses a routing preference that considers only travel time.
In the experiment, we apply each algorithm to learn a routing preference according to a driver's trajectories in the training data. For each trajectory in the testing data, we obtain the source, the destination, and the driver id. Then we apply {\it Dom} and {\it TRIP} to compute the personalized, shortest path connecting the source and the destination according to the driver id.
%
%we use the source and destination, driver identifier  for whom we have learned a routing preference, we run the algorithm using the routing preference to obtain a path. We learn 3 contexts for {\it Dom}.
%
Other routing algorithms that use historical trajectories, e.g., \cite{DBLP:conf/icde/ChenSZ11, DBLP:conf/icde/DaiYGD15, DBLP:conf/mdm/CeikuteJ15}, do not support routing between arbitrary source and destination, and thus are not comparable to \emph{L2R}.

\noindent
\textbf{Accuracy:}
The accuracies of \emph{L2R}, \emph{Shortest}, \emph{Fastest}, {\it Dom}, and {\it TRIP} are calculated using the path similarity functions (see Equations~\ref{eq:psim} and~\ref{eq:psim2}),
and are reported in Figures~\ref{fig:exp_bytime_accu} and~\ref{fig:exp_bytime_accu_union}.
%
%Appendix Appendix~\ref{sec:app_exp} reports on the same evaluation using a different, but also widely applied, path similarity function.% are reported in Appendix~\ref{sec:app_exp}.

\emph{Shortest}'s accuracy drops as the travel distance increases.
This is because \emph{Shortest} tends to find a path that approximates the straight line segment from a source to a destination.
Such paths are often not preferred by drivers. %---none of experienced drivers would choose to drive through streets where they often need to pay extra attentions to pedestrians and bicycles.
In $D_1$, when traveling longer distances, highways are usually preferred. However, given the fact that using highways often yield longer travel distances, \emph{Shortest} does not return such paths. Therefore, the accuracy of \emph{Shortest} is poor for longer distances.

The accuracy of \emph{Fastest} is comparable to that of \emph{Shortest} for small travel distances. However, \emph{Fastest} achieves much higher accuracy when travel distance is longer. %Similar to the shortest path, the fastest route in short distance traveling is not always picked up by local drivers. This is because the local drivers have knowledge about the real traffic in a city, and they, sometimes, avoid the fastest route if they know the traffic is complicated there.
When travelling longer distances, highways usually offer the lowest travel times and are therefore returned by \emph{Fastest}. Thus, \emph{Fastest} achieves much better accuracy than does \emph{Shortest}.

\emph{Dom} achieves higher accuracy than the other routing methods, except \emph{L2R}, because it learns routing preferences that consider the trade-off among distance, travel time, and fuel consumption for individual drivers. However, as it conducts an expensive multi-objective skyline routing process, it requires significantly more running time than other methods (see Figure~\ref{fig:exp_bytime_perf}).
\emph{TRIP} is slightly more accurate than \emph{Fastest} due to the personal ratio learned for each driver, and it needs similar running time to \emph{Shortest} and \emph{Fastest}.

\emph{L2R} achieves the highest accuracy in all settings.
The accuracy increases as the travel distance becomes longer---this is achieved by capturing the preference for  different travel costs and road types in the region graph.

The accuracy of \emph{L2R} decreases when sources and (or) destinations are not in regions. This is intuitive because when no historical trajectories are available for path finding, an \emph{L2R} path simply coincides with the fastest path. However, when historical trajectories can be utilized, \emph{L2R} improves the accuracy of the fastest path (see InOutRegion and OutRegion in Figure~\ref{fig:exp_bytime_accu}(b)).

\noindent
\textbf{Online Running Time:}
Run-times are reported in Figure~\ref{fig:exp_bytime_perf}.
In all settings, \emph{L2R} is most efficient. This is because the path finding process is conducted on the region graph, which is much smaller than the original road network graph.
When sources and (or) destinations are not in regions, the run-time of \emph{L2R} increases because it needs extra time to identify the fastest paths from the source (destination) to a region.

The personalized routing \emph{Dom} requires significantly more running time as it conducts an expensive multi-objective skyline routing process. Next, \emph{Trip} has a running time similar to those of \emph{Shortest} and \emph{Fastest} as all three perform single-objective routing. \emph{Trip} just uses personalized  weights.

\noindent
\textbf{Offline Processing Time for \emph{L2R}:} When using all training data and default parameters,
the offline processing time for constructing the region graph (Section~\ref{sec:networkPartitioning})
and for executing steps 1--3 to learn and transfer routing preferences (Section~\ref{sec:learntransfer}) for %$D_1$ are 0.35, 4.08, 1.77, and 0.11 hours, respectively, and for $D_2$ are 0.35, 4.08, 1.77, %
$D_1$ are 21, 245, 106, and 7 minutes, respectively,
and for $D_2$ are 9, 10, 29, and 0.06 minutes, respectively.
Note that such offline processing is parallelizable, e.g., by MapReduce~\cite{DBLP:conf/dasfaa/YangMQZ09,DBLP:conf/waim/YuanSWYZY10}.

%%\vspace{-0.1cm}
\subsection{Comparison with Google Maps}
%\textbf{Performance:}
We also compare \emph{L2R} with Google Maps.
%We run Google Direction API\footnote{URL to the API} on the testing sets and obtain the paths returned by the API.
%
We query the Google Directions API using a source, a destination, and the departure time from the testing set as arguments to obtain a \emph{Google path}, which consists of a sequence of \emph{waypoints}, represented by longitude-latitude coordinates.

\begin{figure}[!ht]
\centering
  \includegraphics[width=0.8\columnwidth]{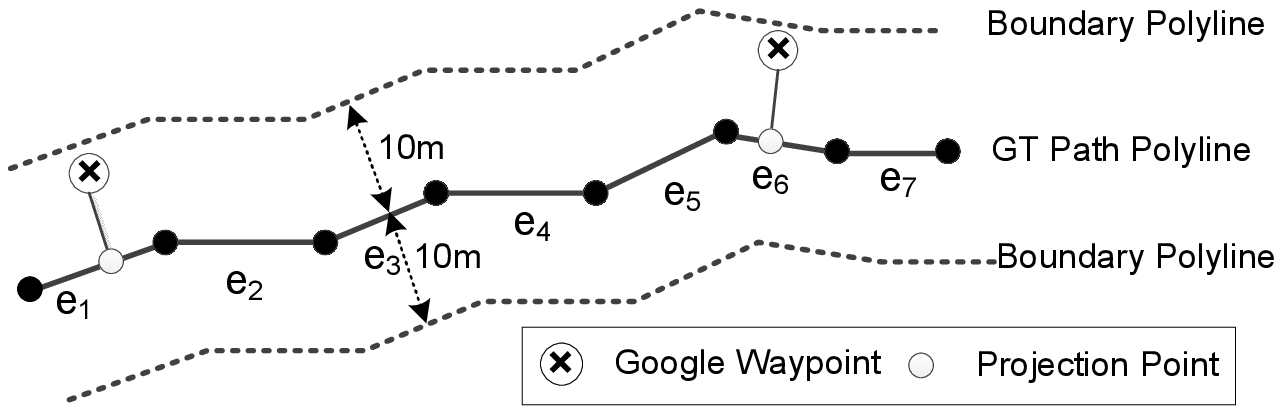}
\caption{Google Accuracy}
\label{fig:google}
%%\vspace{-10pt}
\end{figure}

We follow an existing methodology~\cite{DBLP:conf/mdm/CeikuteJ15} to compute the similarity between a Google path and a GT path.
We first represent a GT path as a polyline in the longitude-latitude coordinate system. We call this polyline a \emph{GT path polyline}.
Next, we introduce two polylines that are parallel to the GT path polyline and are 10 meters away on each side. We thus obtain a \emph{band} around the GT path polyline.
The solid line in Figure~\ref{fig:google} shows the GT path polyline, and the two dashed lines indicate the band.
When a Google waypoint is within the band, it is a matched waypoint. We project each matched waypoint onto the GT path polyline to obtain a \emph{projection point}.
If two consecutive waypoints are matched waypoints, we regard the edges between their projection points as the edges on which the Google path is matched to the GT path, e.g., edges $e_1$, $e_2$, $e_3$, $e_4$, $e_5$, and $e_6$ in Figure~\ref{fig:google}. The above enables us to use the path similarity function in the similarity function in Section V-A of the paper submission to measure the similarity between  Google and GT paths.

%
%Using the intermediate points, we build a band, with deviation of 10m, around a GT route.

We report the accuracy of Google vs. \emph{L2R} paths in Figure~\ref{fig:exp_google}. %Google paths show similar results for \emph{Time} and \emph{Driver}.
%%
%%%\vspace{-0.4cm}
%\begin{figure}[hp!]
%	\centering
%	\subfigure[By distance, $D_1$]{
%		\begin{minipage}[b]{0.22\textwidth}
%			\includegraphics[width=1\textwidth]{fig/eps/GByTimebyDist_accu2.eps}
%		\end{minipage}
%	}
%	\subfigure[By regions, $D_1$]{
%		\begin{minipage}[b]{0.22\textwidth}
%			\includegraphics[width=1\textwidth]{fig/eps/GByTimebyRegion_accu2.eps}
%		\end{minipage}
%	}
%%\subfigure[By distance, $D_2$]{
%%		\begin{minipage}[b]{0.22\textwidth}
%%			\includegraphics[width=1\textwidth]{fig/eps/GByTimebyDist_accu2_CD.eps}
%%		\end{minipage}
%%	}
%%	\subfigure[By regions, $D_2$]{
%%		\begin{minipage}[b]{0.22\textwidth}
%%			\includegraphics[width=1\textwidth]{fig/eps/GByTimebyRegion_accu2_CD.eps}
%%		\end{minipage}
%%	}
%	\caption{Comparison with Google Maps}
%	\label{fig:exp_google}
%%%\vspace{-0.4cm}
%\end{figure}
%
In particular, the accuracy of Google paths lies between 60\% and 85\%, and the accuracy increases with the travel distance. However, Google paths show no pattern when we categorize according to whether the source and destination belong to regions.
%
%Due to the space limitation, we report accuracy of \emph{L2R} and Google when testing data is categorized by travel distance for data set {\it Time} and report the accuracy when the source and destination are categorized by whether they belong to regions for data set {\it Driver}.
%
In all settings, \emph{L2R} achieves higher accuracy, indicating that \emph{L2R} has the potential to improve the quality of state-of-the-art routing services. % with local drivers' historical trajectories. % help of local driver

\section{Conclusion and Outlook}

We propose a learn-to-route solution that enables comprehensive trajectory-based routing. %which incorporates the routing preferences identified from historical local drivers' trajectories to enhance routing service qualities.
The solution encompasses an algorithm that clusters road intersections into regions, yielding a derived region graph. It learns routing preferences for region pairs with sufficient trajectories and transfers these preferences to region pairs with insufficiently many trajectories. It then utilizes the learned and transferred preferences to enable routing. Empirical studies offer evidence that the solution is practical and is able to compute high-quality routes.
In future work, it is of interest to consider finer granularity modeling of time-dependency, e.g., using a time-varying region graph, real-time region graph updates when receiving new trajectories, and the modeling of more than one preference for each T-edge.

%\begin{small}
%\bibliographystyle{IEEEtran}
%\bibliography{erc}
%\end{small}


\end{document}